\documentclass[acmsmall,screen,nonacm]{acmart}
\usepackage{amsmath,amsfonts}
\usepackage{algorithmic}
\usepackage{algorithm}
\usepackage{array}
\usepackage{textcomp}
\usepackage{url}
\usepackage{verbatim}
\usepackage{graphicx}
\usepackage{threeparttable}
\usepackage{multirow}
\usepackage{needspace}
\usepackage{tabularx}
\usepackage{longtable}
\usepackage{makecell}
\usepackage{booktabs}
\usepackage[table]{xcolor}
\usepackage{enumitem}

\definecolor{darkgreen}{rgb}{0.0, 0.5, 0.0}
\definecolor{darkred}{rgb}{0.8, 0.0, 0.0}

\setcopyright{none}
\begin{document}

\title{Plug-and-Play Dramaturge: A Divide-and-Conquer Approach for Iterative Narrative Script Refinement via Collaborative LLM Agents}

\author{Wenda Xie}
\authornote{Both authors contributed equally to this research.}
\email{xiewenda24@mails.ucas.edu.cn}
\affiliation{%
  \institution{Institute of Automation, Chinese Academy of Sciences}
  \city{Beijing}
  \country{China}
}
\author{Chao Guo}
\authornotemark[1]
\email{chao.guo@ia.ac.cn}
\affiliation{%
  \institution{Institute of Automation, Chinese Academy of Sciences}
  \city{Beijing}
  \country{China}
}
\author{Yanqing Jing}
\email{frogjing@tencent.com}
\affiliation{%
  \institution{Tencent}
  \city{Shenzhen}
  \country{China}
}
\author{Junle Wang}
\email{wangjunle@gmail.com}
\affiliation{%
  \institution{Tencent}
  \city{Shenzhen}
  \country{China}
}
\author{Yisheng Lv}
\email{yisheng.lv@ia.ac.cn}
\affiliation{%
  \institution{Institute of Automation, Chinese Academy of Sciences}
  \city{Beijing}
  \country{China}
}
\author{Fei-Yue Wang}
\email{feiyue.wang@ia.ac.cn}
\affiliation{%
  \institution{Institute of Automation, Chinese Academy of Sciences}
  \city{Beijing}
  \country{China}
}

\renewcommand{\shortauthors}{Xie et al.}

\begin{abstract}
Although LLMs have been widely adopted for creative content generation, a single-pass process often struggles to produce high-quality long narratives. How to effectively revise and improve long narrative scripts like scriptwriters remains a significant challenge, as it demands a comprehensive understanding of the entire context to identify global structural issues and local detailed flaws, as well as coordinating revisions at multiple granularities and locations. Direct modifications by LLMs typically introduce inconsistencies between local edits and the overall narrative requirements. To address these issues, we propose Dramaturge, a task and feature oriented divide-and-conquer approach powered by hierarchical multiple LLM agents. It consists of a Global Review stage to grasp the overall storyline and structural issues, a Scene-level Review stage to pinpoint detailed scene and sentence flaws, and a Hierarchical Coordinated Revision stage that coordinates and integrates structural and detailed improvements throughout the script. The top-down task flow ensures that high-level strategies guide local modifications, maintaining contextual consistency. The review and revision workflow follows a coarse-to-fine iterative process, continuing through multiple rounds until no further substantive improvements can be made. Comprehensive experiments show that Dramaturge significantly outperforms all baselines in terms of script-level overall quality and scene-level details. Our approach is plug-and-play and can be easily integrated into existing methods to improve the generated scripts.
\end{abstract}


\maketitle

\begin{figure}[!h]
\centering
\includegraphics[width=0.6\linewidth]{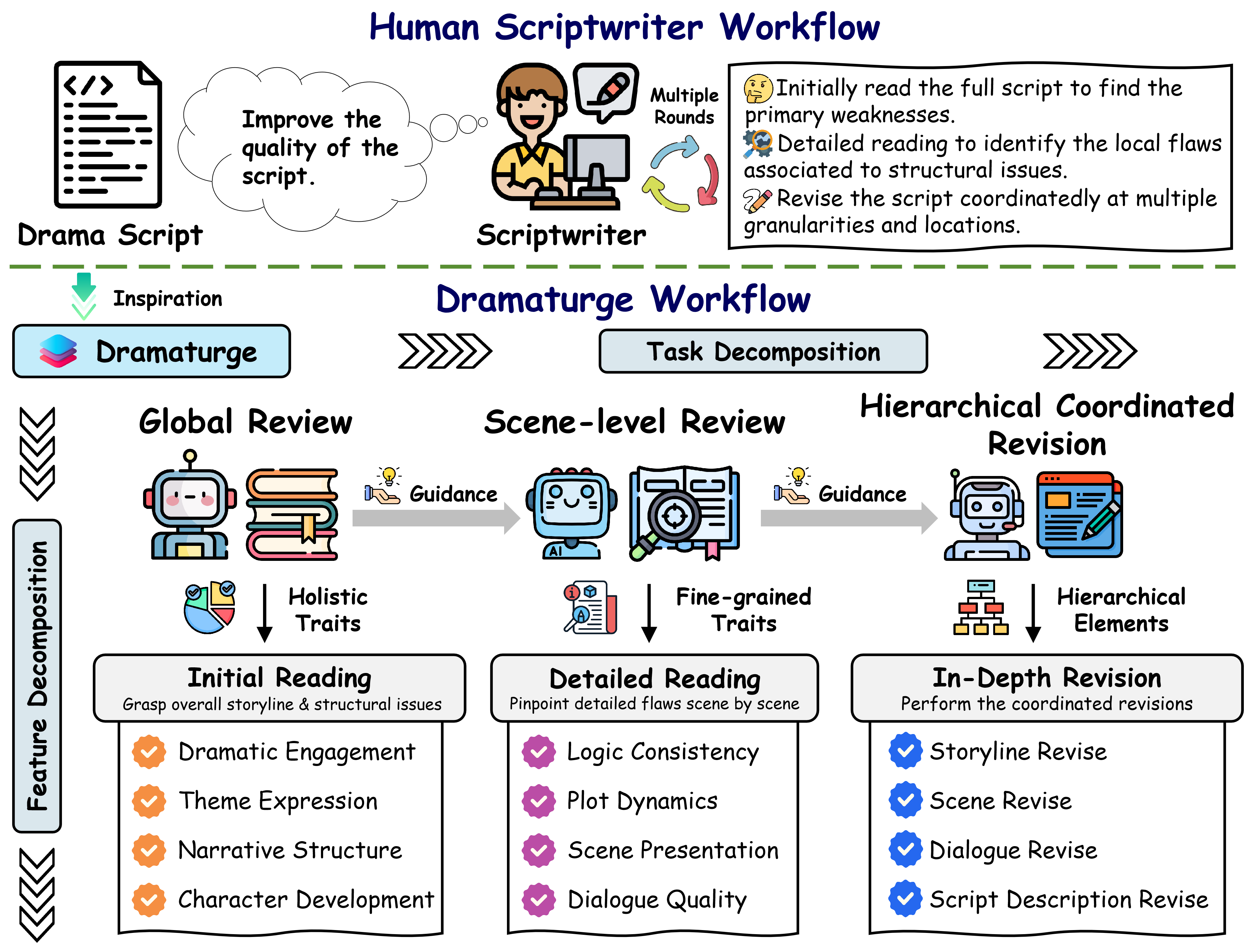}
\caption{Our Dramaturge is inspired by the human scriptwriting process and performs Global Review, Scene-level Review, and Hierarchical Coordinated Revision to iteratively refine narrative scripts via a task and feature oriented divide-and-conquer strategy.}
\label{fig:intro}
\end{figure}

\section{Introduction}

Large Language Models (LLMs) have demonstrated remarkable capabilities in generating diverse and engaging stories \cite{mirowski2023dramatron, yang2023doc, li2024value, zhao2024narrativeplay, feng2025ss}. 

However, existing LLM-based methods generate narratives from scratch using a single-pass workflow, even when multiple attempts are applied during generation. This overlooks a fundamental characteristic of creative practice: human scriptwriters craft scripts through multiple cycles of review and revision \cite{nelmes2008developing, batty2021room} to progressively improve script quality. As a result, current methods often struggle to produce high-quality scripts with coherent and nuanced storytelling \cite{tian2024large, marco2024pron}. This discrepancy highlights the importance of script refinement—a critical yet underexplored area in LLM-based creative writing, especially for long-form narrative generation \cite{gervas2004story, mulholland2015storyscope, porteous2010applying, guo2021conditional, lisena2023ontology, zhang2025lifebench}.

How to effectively revise and improve long narrative scripts using LLMs remains a significant challenge because \cite{wu2025shifting, wu2025longeval, chen-etal-2025-longleader}:

(1) Comprehensively understanding long scripts and accurately identifying fundamental structural weaknesses along with their associated localized flaws is difficult. Without such insight, revisions tend to be superficial text editing and fail to resolve deeper narrative issues.

(2) It is challenging to coordinate revisions across multiple granularities and locations to maintain consistency between the global narrative and local modifications, while newly introduced contradictions may degrade script quality.

To address these challenges, we propose \textsc{Dramaturge}, a divide-and-conquer approach for iterative narrative script refinement via collaborative LLM agents. This approach emulates scriptwriters' workflow of initial reading, detailed reading, and in-depth revision, as illustrated in Fig.~\ref{fig:intro}. 
The strategy is applied across task and feature dimensions. First, the complex task is decomposed into three hierarchical stages: (1) a Global Review stage that performs a holistic analysis to grasp the storyline and identify structural issues; (2) a Scene-level Review stage that pinpoints local flaws and generates concrete, actionable suggestions for each scene; (3) a Hierarchical Coordinated Revision stage that revises the script coordinately across multiple locations. 
Second, through feature decomposition, each stage deploys a team of specialized agents, each focusing on a distinct dimension. The individual suggestions are consolidated to resolve conflicts and are then routed to the relevant agents in subsequent stages. 
The top-down task flow ensures that high-level strategies from the Global Review guide the detailed suggestions during the Scene-level Review, both of which jointly direct the local edits in the revision stage. Such coordination ensures that local modifications are aligned with the global narrative objectives.
Finally, the entire workflow follows coarse-to-fine iterations by multiple rounds until the script quality stabilizes. This prevents superficial revisions by first adjusting the fundamental storyline before refining details, thereby achieving deep and comprehensive script enhancements.

In summary, the main contributions of this paper include:

\begin{itemize}[topsep=0pt,itemsep=2pt,parsep=0pt]
    \item We introduce Dramaturge, a plug-and-play framework for iterative coarse-to-fine narrative script refinement, which can be easily integrated into existing methods to facilitate in-depth reviews and revisions.
    \item We propose a task and feature oriented divide-and-conquer strategy to achieve coordinated review and revision across multiple granularities and locations in scripts. This approach enables high-level strategies to guide local modifications while preserving contextual consistency.
    \item Comprehensive experiments show that Dramaturge significantly outperforms all strong baselines, improving upon the original scripts by 53.2\% in script-level overall quality and 65.1\% in scene-level details, and surpassing the strongest baseline by 8.2\% and 18.4\%, respectively.
\end{itemize}

\section{Related Work}

\subsection{LLM-Based Narrative Generation}

Narrative generation methods have gradually evolved from early generative neural networks \cite{fan2018hierarchical, yao2019plan, fan2019strategies, yu2020draft, ammanabrolu2021automated} to LLM-based approaches that leverage prompt engineering, hierarchical generation frameworks, and multi-agent role-playing. Dramatron \cite{mirowski2023dramatron} proposes a hierarchical generation framework that decomposes screenwriting into structured stages and generates high-quality long-form scripts. DOC \cite{yang2023doc} introduces detailed outline control to solve long-story coherence degradation problems. HoLLMwood \cite{chen2024hollmwood} adopts a role-playing mechanism in which specialized LLMs act as writers, editors, and actors, collaborating through iterative feedback cycles. IBSEN \cite{han2024ibsen} considers real-time user participation and proposes a director–actor collaboration framework to enable controllable script generation in human–AI interaction. FilmAgent \cite{xu2024filmagent} expands beyond traditional scriptwriting to encompass film production automation through multi-agent coordination. Ex3 \cite{lei2024ex3} improves long-form novel generation by combining structure-aware instruction tuning with a tree-based expansion method, enhancing narrative coherence and character development. Neeko \cite{yu2024neeko} adopts character-specific LoRA adapters and a gating mechanism to enable scalable multi-character role-playing. CRITICS \cite{bae2024collective} enhances long-form story generation by incorporating a collaborative multi-agent critique process, addressing limitations in creativity and expressiveness of traditional plan-based methods. TaleForge \cite{nguyen2024taleforge} introduces an interactive multimodal system that enables personalized story creation through user-driven character and environment customization, addressing the growing demand for adaptive storytelling experiences. A knowledge-guided storytelling approach \cite{pan2024guiding} leverages structured graphs to enhance narrative consistency and factual accuracy. R2 \cite{lin2025r} proposes a Reader–Rewriter framework for adapting novels into screenplays, addressing hallucination and causality issues via refinement and plot graph construction. Domain-specific approaches \cite{cao2025multi} address cultural requirements by generating traditional opera scripts that preserve authenticity and narrative conventions. BookWorld \cite{bookworld2025} constructs multi-agent societies grounded in established fictional universes, addressing the challenge of simulating dynamic characters and consistent worldviews in book-based settings. 

Despite multiple attempts during generation, existing narrative generation methods still adhere to a single-pass workflow and lack iterative revisions employed by scriptwriters to progressively improve script quality. This limitation constrains the quality of the generated scripts. In this paper, we propose a pluggable framework that iteratively refines scripts in a coarse-to-fine manner, enabling systematic improvements while preserving contextual consistency.

\subsection{LLM-driven Multi-agent Systems}
Recent work shows that collaborative LLM agents exhibit superior cognitive and reasoning capabilities over individual LLMs in complex tasks, and multi-agent collaboration has been widely applied to content creation \cite{zhang2024survey, durante2024agent, guo2024large, xi2025rise, qian2025scaling}. CAMEL \cite{li2023camel} proposes a communicative agent framework to study collaborative behaviors and multi-agent interactions. ChatDev \cite{qian2024chatdev} creates virtual companies where agents play professional roles to solve complex tasks through collaboration. AutoAgents \cite{chen2024autoagents} develops an automatic agent generation framework to dynamically create and coordinate specialized agents based on task requirements. STORM \cite{shao2024assisting} introduces a multi-agent writing system that synthesizes topic outlines through retrieval and multi-perspective question asking to generate long-form articles. Agents' Room \cite{huot2024agents} consider story generation as a multi-step collaboration problem, proposing planning and writing agents to achieve enhanced story coherence. Chain of Agents \cite{zhang2024chain} addresses long-context processing challenges by distributing tasks across multiple LLM agents in a sequential chain for improved efficiency and accuracy. A multi-agent framework \cite{du2024improving} enhances factuality and reasoning in language models via collaborative deliberation. Debate-to-Write \cite{hu2024debate} introduces persona-driven multi-agent frameworks for diverse argument generation, enabling more comprehensive perspective coverage. AgentCoord \cite{pan2024agentcoord} provides visual exploration tools for understanding and optimizing coordination strategies in LLM-based multi-agent systems.

LLM-driven multi-agent systems are widely used to decompose and tackle complex tasks. In our task, coordinating local modifications with global narrative coherence remains challenging. We leverage the logical reasoning capabilities of LLM agentic systems and propose divide-and-conquer strategies along both task and feature dimensions to revise long-context scripts, enabling coordinated revisions across multiple granularities and locations under global guidance.


\subsection{Parallel Art}

Beyond narrative generation, broader frameworks for AI-assisted creation have been proposed. Inspired by parallel systems and ACP methods\cite{wang2004parallel, wang2004artificial, wang2007toward, wang2016steps}, Parallel Art \cite{guo2019parallel, guo2023chatgpt, guo2023artverse} establishes a framework for human-machine collaborative artistic creation through computational experiments and iterative optimization between virtual and actual systems. Building on this foundation, Parallel Theaters \cite{ni2023parallel, ni2025theaters} further introduces a drama-creation approach that coordinates human, digital, and robotic artists. Through narrative reasoning and progression, it transforms decision-making processes into immersive dramatic interactions, thereby enabling what is known as a decision theater. 

Following the notion of computational experiments, we build a closed-loop pipeline for generation, evaluation, and optimization, enabling iterative revision and progressive improvement of long-form narratives.

\section{Methodology}

\subsection{Overall Architecture}
Dramaturge takes a narrative script as input and progressively refines it by emulating the scriptwriters' iterative process of initial reading, detailed reading, and in-depth revision. It enables seamless integration with existing script generation methods as a post-processing enhancement module. 

\begin{figure}[!t]
\centering
\includegraphics[width=0.95\linewidth]{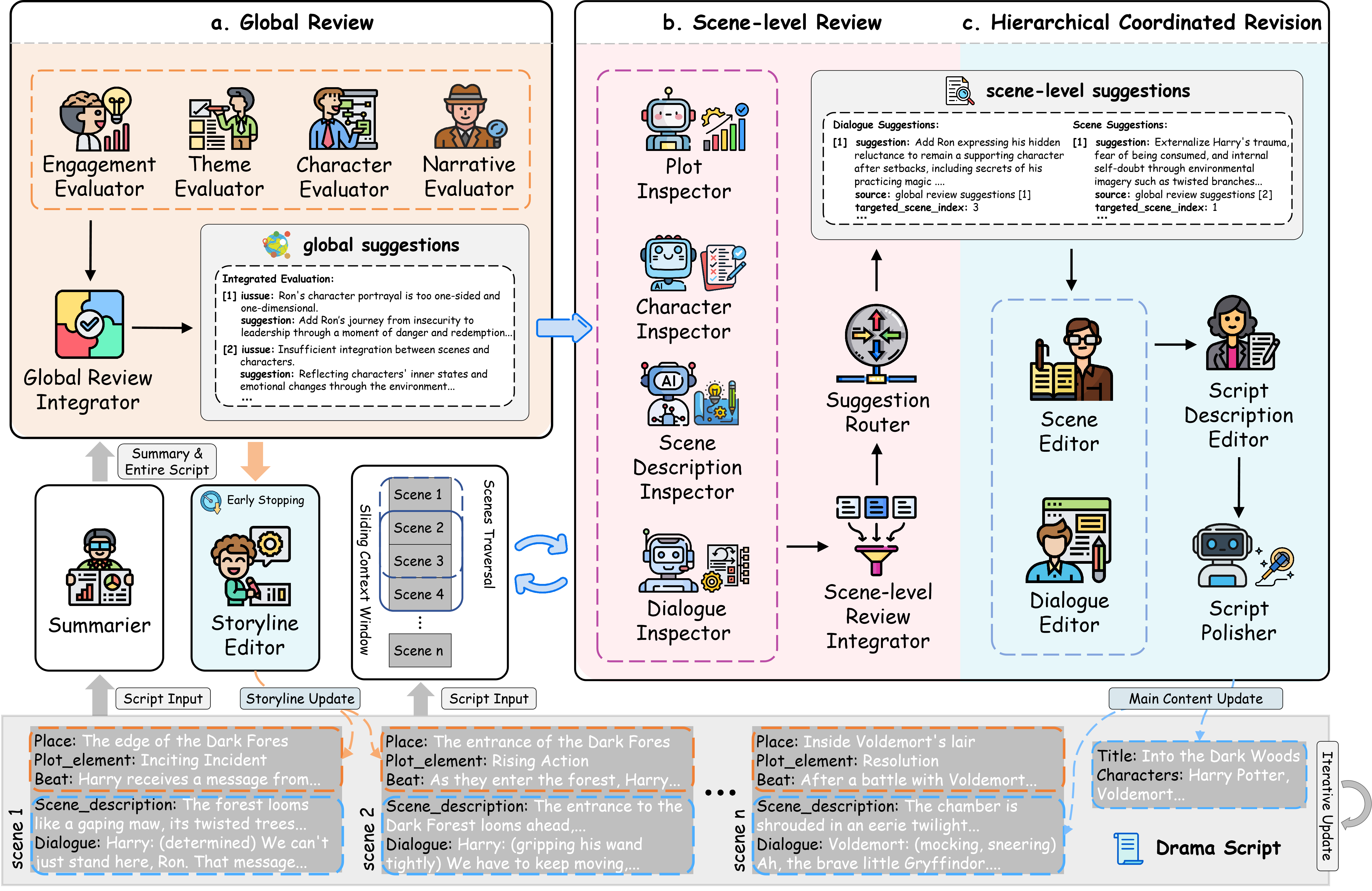}
\caption{The Architecture of Dramaturge. A task and feature oriented divide-and-conquer strategy is adopted, leveraging collaborative LLM agents to perform coarse-to-fine iterative refinement, while coordinating local edits and global structural adjustments and maintaining contextual consistency.}
\label{fig1}
\end{figure}

The system architecture is illustrated in Fig.~\ref{fig1}. We adopt a divide-and-conquer strategy along task and feature dimensions. For task decomposition, a three-stage hierarchical structure is employed: (1) Global Review performs holistic analysis based on the full script and summary to formulate high-level improvement strategies; (2) Scene-level Review traverses each scene and produces detailed modification suggestions following global strategies; (3) Hierarchical Coordinated Revision coordinates modifications across multiple granularities and locations. For feature decomposition, each stage deploys specialized agents targeting distinct review dimensions. Their suggestions are integrated to resolve conflicts and redundancies, then routed to subsequent stages. The refinement workflow proceeds iteratively in a coarse-to-fine manner across multiple rounds until no substantive improvement can be made. The algorithmic procedure is presented in Algorithm~\ref{alg:dramaturge}.

\subsection{Global Review}
The Global Review (GR) performs holistic script-level analysis using the complete script and a summary as inputs to identify high-level structural issues through multi-agent collaboration. The summary is generated by a Summarizer integrating the core plot to provide a condensed overview.

Following the feature decomposition strategy, four evaluators are deployed to focus on distinct dimensions for comprehensive structural analysis:
\begin{itemize}[topsep=0pt,itemsep=1pt,parsep=0pt]
    \item Engagement Evaluator: Assesses the story's appeal, considering plot progression, dramatic tension, and interestingness to enhance the story's excitement.
    \item Character Evaluator: Analyzes the character development, considering psychological depth and emotional richness to create more memorable characters.
    \item Theme Evaluator: Evaluates theme presentation and depth, focusing on emotional resonance and underlying meanings to deepen the story's emotional impact.
    \item Narrative Evaluator: Reviews structural integrity and narrative flow, paying attention to pacing and logical consistency to heighten overall story readability.
\end{itemize}

The Global Review Integrator consolidates suggestions from four evaluators prioritizing each suggestion according to both scope and magnitude of its impact. We first group suggestions by scope: (1) structural directives that reshape narrative arcs, character development, or causal relations spanning multiple scenes; (2) cross-scene adjustments that affect narrative continuity or thematic resonance; (3) localized detail optimizations. Items in a higher scope group always outrank those in lower groups. Within each group, suggestions tightly coupled with the main plotline take precedence over secondary subplots, and cross-scene effects surpass single-scene ones. When two items still tie, we follow the order: Narrative, Character, Theme, and Engagement. Higher-priority directives are retained, with compatible lower-priority ones merged when possible and conflicting ones discarded. The resulting non-overlapping plan routes plot-critical guidance to the Storyline Editor, while remaining directives flow to the Scene-level Review for fine-grained inspection.

The prompts for the specialized agents in the GR stage are provided in Appendix A.

\subsection{Scene-level Review}
The Scene-level Review (SLR) traverses each scene to generate concrete, scene-level modification suggestions under the guidance of the GR, prioritizing global consistency. When a scene-level recommendation conflicts with GR directives, the local suggestion is subordinated to the global narrative strategy to eliminate potential inconsistencies. The resulting suggestions are then synthesized into component-specific guidance for subsequent revision, ensuring coherence and precision in the process.

Following the feature decomposition strategy, four inspectors are deployed for detailed analysis in each scene:

\begin{itemize}[topsep=0pt,itemsep=1pt,parsep=0pt]
    \item Dialogue Inspector: Analyzes dialogue quality, analyzing character dialogue flow and emotional dynamics to enhance plot advancement through verbal interactions.
    \item Scene Description Inspector: Evaluates sensory elements. Focuses on setting atmosphere and sensory details to create immersive story worlds.
    \item Plot Inspector: Verifies scene coherence by examining plot holes and internal inconsistencies to strengthen local narrative integrity.
    \item Character Inspector: Validates character behavioral consistency. Ensures character actions align with established traits and motivations.
\end{itemize}

Following the inspection, the Scene-level Review Integrator scores each recommendation with a three-stage priority policy to resolve conflicts: (1) compliance with directives from the Global Review is mandatory, with conflicting local suggestions being immediately rejected and alignments with GR guidance receiving top priority; (2) within the same inspector tier, we favor interventions whose impact propagates to subsequent scenes (e.g., foreshadowing, metaphors) or multiple dialogue exchanges ahead of purely stylistic tweaks; (3) among the remaining candidates, inspector type dictates precedence: Plot and Character Inspectors rank above Dialogue and Scene Description Inspectors because they most directly influence causality and behavioral authenticity. Higher-priority items are preserved, compatible lower-priority ones are merged, and incompatible ones are dropped. The resulting directives are mapped to specific dialogue or descriptive spans to provide actionable, context-aware guidance for the subsequent editors.

Subsequently, the Suggestion Router classifies each modification by its intended revision domain and routes it to either the subsequent Dialogue Editor or the Scene Editor accordingly. This systematic routing maintains the specialized focus of each editor while avoiding conflicting modifications.
The prompts for the specialized agents in the SLR stage are provided in Appendix B.

\begin{algorithm}[t]
\caption{DRAMATURGE: Divide-and-Conquer Script Refinement Process}
\label{alg:dramaturge}
\begin{algorithmic}[1]
\small
\STATE \textbf{Input:} Initial Script $S_0$, max iterations $T_{max}$, iteration stopping threshold $\theta$, max retries $N_{max}$
\STATE \textbf{Output:} Refined script $S$
\STATE $i \gets 0$, $S_i \gets S_0$, $S_{\text{prev}} \gets S_0$, $phase \gets \text{structural refinement}$
\WHILE{$i < T_{max}$}
    \STATE \textcolor{blue}{\scriptsize\textbf{// Global Review}}
    \STATE $P \gets \textsc{Summarize}(S_i)$
    \STATE $E \gets \textsc{Evaluate}(S_i, P)$ \COMMENT{$E_{eng}, E_{char}, E_{theme}, E_{narr}$}
    \STATE $G \gets \textsc{Integrate}_G(E)$
    \STATE \textcolor{blue}{\scriptsize\textbf{// Hierarchical Coordinated Revision - structural refinement}}
    \IF{$phase = \text{structural refinement}$}
        \STATE $S_{i} \gets \textsc{StorylineEdit}(S_i, G)$
    \ENDIF
    \STATE \textcolor{blue}{\scriptsize\textbf{// Scene-level Review}}
    \FOR{each scene $s^j \in S_i$}
        \STATE $I_j \gets \textsc{Inspect}(s^j, \textsc{Context}(s^j), G)$ \COMMENT{$I_{dial}, I_{desc}, I_{plot}, I_{char}$}
        \STATE $R_j \gets \textsc{Integrate}_S(I_j)$
    \ENDFOR
    \STATE $\{R_{dial}, R_{scene}\} \gets \textsc{Route}(\{R_j\})$
    \STATE \textcolor{blue}{\scriptsize\textbf{// Hierarchical Coordinated Revision - detail refinement}}
    \FOR{$s^j \in S_i$}
        \STATE $s^j \gets \textsc{SceneEdit}(s^j, \textsc{Context}(s^j), R_{scene}^j)$
        \STATE $s^j \gets \textsc{DialogueEdit}(s^j, \textsc{Context}(s^j), R_{dial}^j)$
    \ENDFOR
    \STATE $S_{i} \gets \textsc{ScriptDescriptionEdit}(S_i)$
    \STATE $S_{i} \gets \textsc{Polish}(S_i)$ 
    \STATE \textcolor{blue}{\scriptsize\textbf{// Evaluating the quality improvement}}
    \STATE $\Delta \gets \textsc{EvaluateImprovement}(S_{\text{prev}}, S_i)$
    \IF{$phase = \text{structural refinement}$ \textbf{and} $\Delta < \theta$}
        \STATE $phase \gets \text{detail refinement}$
    \ENDIF
    \IF{$phase = \text{detail refinement}$ \textbf{and} $\Delta < \theta$ for $N_{max}$ retries}
        \STATE \textbf{break}
    \ENDIF
    \STATE $S_{\text{prev}} \gets S_i$
    \STATE $i \gets i + 1$
\ENDWHILE
\STATE \textbf{return} $S_i$
\end{algorithmic}
\end{algorithm}

\subsection{Hierarchical Coordinated Revision}
The Hierarchical Coordinated Revision (HCR) stage serves as the system's execution engine, executing improvement guidance from both the GR and SLR stages through a multi-granularity editing workflow: the Storyline Editor for global plot reconstruction, Scene and Dialogue Editors for scene-level refinements, and the Script Description Editor together with the Script Polisher for consistency metadata updates and consistency verification. By aligning high-level narrative planning with detailed content refinement, this coordinated process not only enhances scene quality but also elevates the story's overall narrative appeal.

Guided by the GR, the Storyline Editor creatively enhances the plot by integrating engaging plot developments and deepening character motivations while maintaining the original settings to improve the overall plot structure. This editor applies early stopping during the detail refinement phase to prevent quality oscillation and ensure convergence.

Following the feature decomposition strategy, two specialized editors implement granular modifications based on guidance routed by the Suggestion Router:
\begin{itemize}[topsep=0pt,itemsep=1pt,parsep=0pt]
    \item The Scene Editor receives scene description-focused suggestions and generates enhanced scene descriptions by adding visual, atmospheric elements, and enriching sensory details. 
    \item The Dialogue Editor receives dialogue-focused suggestions and generates enhanced dialogue by implementing psychological details and character-specific actions.
\end{itemize}

After reading the revised script, the Script Description Editor updates the title to align with the refined thematic direction and revises character descriptions to ensure consistency in psychological and behavioral representation.

As the final quality assurance module, the Script Polisher performs a holistic skimming of the entire script to build a comprehensive understanding. It subsequently proceeds with a scene traversal, meticulously comparing local details with the storyline and correcting any deviations to ensure the final script's coherence and consistency.

The prompts for the specialized agents in the HCR stage are provided in Appendix C.

\subsection{Iterative Improvement with Quality Control}
Dramaturge refines scripts through a coarse-to-fine iterative process, guided by quality control to ensure continuous improvement. This process is structured into two phases:

\begin{itemize}[topsep=0pt,itemsep=1pt,parsep=0pt]
    \item Phase 1: structural refinement operates with all agents active to address fundamental narrative issues by revising storylines in coordination with detailed adjustments, achieving significant overall improvements.
    \item Phase 2: detail refinement initiates early stopping of the Storyline Editor to lock the storyline, guiding the refinement process toward convergence. This phase focuses on refining details such as dialogue and scene descriptions under the constraints of the updated storyline.
\end{itemize}
The system transitions from structural refinement to detail refinement when no substantive improvement can be made, and the iterative process terminates if detail refinement fails to yield improvement within maximum script revision retries ($N_{max}$). Through parameter experiments, we found that a threshold of 1.0 points in the Script-Level Overall Evaluation ensures substantial modifications and stable convergence, avoiding unnecessary and frequent revisions caused by minor score fluctuations. Therefore, we define a substantive improvement as an increase of at least this threshold.

\section{Experiments}

\subsection{Experimental Settings}

\subsubsection{Implementation Details}
Our Dramaturge uses GPT-4.1-mini as the backbone model with temperature = 0.7 and top-p = 0.95 for script refinement to achieve creative outputs. We also investigated the impact of varying $N_{max}$ within the range of $[1, 10]$. We set $N_{max}=3$ to make a trade-off between performance and computational cost, as performance gains largely plateau beyond this threshold for all methods.

For evaluation, we employ Gemini-2.5-pro with temperature = 0 to obtain stable and rigorous assessment results. This setting is grounded in our comparative analysis of mainstream flagship models from the Gemini, GPT, Qwen, and DeepSeek series. 
Pairwise Kolmogorov–Smirnov (KS) tests on the evaluation outputs revealed no significant differences in score distributions across models ($p>0.1$), indicating broad consensus in their evaluation standards. However, among these statistically consistent models, Gemini-2.5-pro exhibited the most stringent absolute scoring. 
Therefore, we adopt it as a strict evaluator to enforce a high-quality standard. 
Furthermore, using Gemini as the evaluation model mitigates potential self-preference bias that may arise when the same model family (GPT) is used for both generation and evaluation.

\begin{table*}[!t]
\caption{Quantitative Evaluation Results. Script-level scores evaluate the complete scripts, while scene-level scores are aggregated from scene-by-scene comparisons, emphasizing the quality of details. Despite using a relatively weaker backbone, our \textsc{Dramaturge} significantly outperforms the original scripts and six strong baselines in both overall quality and details. \textbf{Bold} indicates significance with $p < 0.05$.}
\label{tab:comprehensive_quality_assessment}
\centering
\tiny
\setlength{\tabcolsep}{1.8pt}
\renewcommand{\arraystretch}{1.1}
\begin{tabularx}{\textwidth}{@{}p{2.9cm}*{6}{>{\centering\arraybackslash}X}@{}}
\toprule
\multirow{2}{2.9cm}{\centering\textbf{Method}} & \multicolumn{5}{c}{\textbf{\makecell{Script-Level \\ Overall Evaluation}}} & \textbf{\makecell{Scene-Level \\ Comparative \\ Evaluation}} \\
\cmidrule(lr){2-6} \cmidrule(lr){7-7}
& \textbf{\makecell{Character \\ Development $\uparrow$}} & \textbf{\makecell{Narrative \\ Structure $\uparrow$}} & \textbf{\makecell{Dialogue \\ Quality $\uparrow$}} & \textbf{\makecell{Scene \\ Presentation $\uparrow$}} & \textbf{\makecell{Total \\ Score $\uparrow$}} & \textbf{\makecell{Total \\ Score $\uparrow$}} \\
\midrule
Original Scripts & 12.50 & 14.60 & 12.58 & 15.56 & 55.24 & 54.50 \\
\midrule
GPT-4o & 13.52 \tiny{\textcolor{darkgreen}{$\uparrow$8.2\%}} & 15.42 \tiny{\textcolor{darkgreen}{$\uparrow$5.6\%}} & 13.24 \tiny{\textcolor{darkgreen}{$\uparrow$5.2\%}} & 16.40 \tiny{\textcolor{darkgreen}{$\uparrow$5.4\%}} & 58.58 \tiny{\textcolor{darkgreen}{$\uparrow$6.0\%}} & 52.20 \tiny{\textcolor{darkred}{$\downarrow$4.2\%}} \\
Qwen-max & 13.30 \tiny{\textcolor{darkgreen}{$\uparrow$6.4\%}} & 15.36 \tiny{\textcolor{darkgreen}{$\uparrow$5.2\%}} & 13.28 \tiny{\textcolor{darkgreen}{$\uparrow$5.6\%}} & 16.24 \tiny{\textcolor{darkgreen}{$\uparrow$4.4\%}} & 58.18 \tiny{\textcolor{darkgreen}{$\uparrow$5.3\%}} & 52.36 \tiny{\textcolor{darkred}{$\downarrow$3.9\%}} \\
Deepseek-r1 & 16.00 \tiny{\textcolor{darkgreen}{$\uparrow$28.0\%}} & 16.56 \tiny{\textcolor{darkgreen}{$\uparrow$13.4\%}} & 14.46 \tiny{\textcolor{darkgreen}{$\uparrow$14.9\%}} & 19.36 \tiny{\textcolor{darkgreen}{$\uparrow$24.4\%}} & 66.38 \tiny{\textcolor{darkgreen}{$\uparrow$20.2\%}} & 63.96 \tiny{\textcolor{darkgreen}{$\uparrow$17.4\%}} \\
GPT-4.1-mini & 16.36 \tiny{\textcolor{darkgreen}{$\uparrow$30.9\%}} & 17.78 \tiny{\textcolor{darkgreen}{$\uparrow$21.8\%}} & 15.72 \tiny{\textcolor{darkgreen}{$\uparrow$25.0\%}} & 18.96 \tiny{\textcolor{darkgreen}{$\uparrow$21.9\%}} & 68.82 \tiny{\textcolor{darkgreen}{$\uparrow$24.6\%}} & 61.70 \tiny{\textcolor{darkgreen}{$\uparrow$13.2\%}} \\
GPT-4.1 & 16.78 \tiny{\textcolor{darkgreen}{$\uparrow$34.2\%}} & 18.10 \tiny{\textcolor{darkgreen}{$\uparrow$24.0\%}} & 16.50 \tiny{\textcolor{darkgreen}{$\uparrow$31.2\%}} & 19.40 \tiny{\textcolor{darkgreen}{$\uparrow$24.7\%}} & 70.78 \tiny{\textcolor{darkgreen}{$\uparrow$28.1\%}} & 62.72 \tiny{\textcolor{darkgreen}{$\uparrow$15.1\%}} \\
Gemini-2.5-pro & 18.60 \tiny{\textcolor{darkgreen}{$\uparrow$48.8\%}} & 20.28 \tiny{\textcolor{darkgreen}{$\uparrow$38.9\%}} & 17.94 \tiny{\textcolor{darkgreen}{$\uparrow$42.6\%}} & 21.36 \tiny{\textcolor{darkgreen}{$\uparrow$37.3\%}} & 78.18 \tiny{\textcolor{darkgreen}{$\uparrow$41.5\%}} & 76.00 \tiny{\textcolor{darkgreen}{$\uparrow$39.4\%}} \\
\rowcolor{gray!15}
\textbf{Ours} {\tiny(GPT-4.1-mini as backbone)} & \textbf{20.64} \tiny{\textcolor{darkgreen}{$\uparrow$65.1\%}} & \textbf{21.16} \tiny{\textcolor{darkgreen}{$\uparrow$44.9\%}} & \textbf{20.20} \tiny{\textcolor{darkgreen}{$\uparrow$60.6\%}} & \textbf{22.60} \tiny{\textcolor{darkgreen}{$\uparrow$45.3\%}} & \textbf{84.60} \tiny{\textcolor{darkgreen}{$\uparrow$53.2\%}} & \textbf{90.00} \tiny{\textcolor{darkgreen}{$\uparrow$65.1\%}} \\
\bottomrule
\end{tabularx}
\end{table*}

\subsubsection{Dataset}
Due to the substantial length of individual scripts and the significant computational overhead involved in their processing and evaluation, previous studies typically used datasets with only a few dozen samples (e.g., DOC \cite{yang2023doc}, 20 scripts; HoLLMwood \cite{chen2024hollmwood}, 60 scripts). Thus, to achieve broader coverage while remaining within cost and efficiency constraints, we construct a dataset of 100 narrative scripts drawn evenly from five sources:

\begin{itemize}[topsep=0pt,itemsep=1pt,parsep=0pt]
    \item \textbf{Writingprompts} \cite{fan2018hierarchical}: A collection of creative stories from the Reddit community.
    \item \textbf{Dramatron} \cite{mirowski2023dramatron}: High-quality generated scripts using a hierarchical generation strategy.
    \item \textbf{DOC} \cite{yang2023doc}: Long stories generated using a two-stage outline control strategy.
    \item \textbf{MoPS} \cite{ma2024mops}: Generated scripts leveraging Well-crafted modular premise synthesis.
    \item \textbf{Agents' Room} \cite{huot2024agents}: High-quality stories collected through collaborative writing.
\end{itemize}

The detailed statistics of the experimental datasets are provided in Appendix F.

\subsection{Baseline Methods}

Currently, there are no specialized methods focused on script refinement for comparison. Thus, we select six strong flagship LLM models without employing the divide-and-conquer strategy proposed in this paper as baselines. These baselines include: \textbf{GPT-4o} (OpenAI's general-purpose model), \textbf{DeepSeek-R1} (specialized in logical reasoning), \textbf{Qwen-max} (Alibaba's flagship model), \textbf{GPT-4.1} (OpenAI's flagship model), \textbf{Gemini-2.5-pro} (Google's latest model), and \textbf{GPT-4.1-mini} (GPT-4.1's efficient version). All experiments are conducted using officially released APIs.

To ensure fair and rigorous comparison, all baselines adopt the same prompts constructed by concatenating the prompts of our functional agents. Additionally, they follow the same iterative refinement and quality control settings as ours, isolating the impact of implementation variations. 

\subsection{Evaluation}
Following established practices in existing work, we employ LLM-based evaluation for script quality assessment \cite{wu2024role, han2024ibsen, ma2024mops, chen2024hollmwood, chang2024survey, lee2025reasoning}. We observe that directly evaluating long full scripts faces challenges due to the context understanding limitations of LLMs, making it difficult to effectively capture both global structure and details. Moreover, a script may be structurally sound at the global level but weak in local details, or vice versa. Therefore, we design a complementary evaluation framework that enables a comprehensive assessment of both overall quality and scene-level details.

\subsubsection{Script-Level Overall Evaluation}
A holistic assessment of the entire script is conducted, focusing on overall narrative quality. The script is transmitted in chunks, and the evaluator is required to assess it based on the complete content. While this metric offers a global perspective, it may overlook localized details.

\subsubsection{Scene-Level Comparative Evaluation}
A pairwise comparison is conducted to assess scene-level details of the scripts. The evaluator performs a comparative analysis between the two scripts by traversing corresponding scenes. The analyses across all scenes are then aggregated to produce a final score. To mitigate position bias \cite{wang2024large}, we implement balanced position calibration by randomly alternating the presentation order of compared scenes. This metric highlights granular improvements that may be overlooked in global assessments.

\subsubsection{Evaluation Metrics}
Script quality is assessed across four critical dimensions:

\begin{itemize}[topsep=0pt,itemsep=1pt,parsep=0pt]
    \item Character Development: Evaluates character portrayal, including psychological depth, complexity, and developmental arc.
    \item Narrative Structure: Analyzes overall story construction, encompassing thematic expression, dramatic tension, and plot consistency.
    \item Dialogue Quality: Measures conversational effectiveness, covering emotional impact, voice distinction, and subtext sophistication.
    \item Scene Presentation: Assesses environment description, including atmospheric creation and sensory immersion. 
\end{itemize}
In script-level evaluation, each dimension is scored on a 25-point scale with a total score of 100. The scene-level evaluation produces an overall score by aggregating the pairwise comparison analysis across all scenes to maintain overall consistency. The prompts are provided in Appendix D.

\section{Results and Analysis}

\subsection{Quantitative Results}
The quantitative evaluation results are presented in Table~\ref{tab:comprehensive_quality_assessment}, considering both the Script-Level Overall Evaluation and Scene-Level Comparative Evaluation. 

\subsubsection{Script-Level Overall Evaluation Results}
Dramaturge achieves a total score of 84.60, representing a substantial 53.2\% improvement over the original scripts (55.24). Despite using a much weaker backbone GPT-4.1-mini, our method significantly outperforms all baseline models and surpasses the strongest baseline Gemini-2.5-pro by 8.2\%, demonstrating a greater advantage from architectural design than merely increasing model capacity. This success stems directly from our divide-and-conquer strategy, where top-down information flow ensures that high-level directives systematically guide local modifications, maintaining global narrative consistency and preventing new contradictions.

\subsubsection{Scene-Level Comparative Evaluation Results}
Dramaturge achieves 90.00, representing a 65.1\% improvement over the original scripts and outperforming all baseline methods, including the strongest Gemini-2.5-pro by 18.4\%. Evaluation results indicate that our framework yields greater improvements in granular refinement at the scene level than at the script level. This advantage stems from the SLR stage, where multi-agent reviews are systematically integrated and routed to appropriate locations, enabling precise identification and correction of detail flaws. Several baseline models even exhibit performance drops due to introducing new inconsistencies during revision, which highlights the inherent difficulty of our task.

\begin{table}[!t]
\caption{The results present the mean scores and corresponding standard errors (SE) for both script-level overall evaluation and scene-level comparative evaluation.}
\label{tab:combined_standard_errors}
\centering
\small
\setlength{\tabcolsep}{3pt}
\renewcommand{\arraystretch}{1.1}
\begin{tabular}{@{}lcccc@{}}
\toprule
\multirow{2}{*}{\textbf{Method}} & \multicolumn{2}{c}{\textbf{Script-Level Evaluation}} & \multicolumn{2}{c}{\textbf{Scene-Level Evaluation}} \\
\cmidrule(lr){2-3} \cmidrule(lr){4-5}
& \textbf{Mean} & \textbf{SE} & \textbf{Mean} & \textbf{SE} \\
\midrule
Original Scripts & 55.24 & 0.78 & 54.50 & 1.60 \\
\midrule
GPT-4o & 58.58 & 0.65 & 52.20 & 0.85 \\
Qwen-max & 58.18 & 0.57 & 52.36 & 0.78 \\
Deepseek-r1 & 66.38 & 0.76 & 63.96 & 0.94 \\
GPT-4.1-mini & 68.82 & 0.67 & 61.70 & 0.97 \\
GPT-4.1 & 70.78 & 0.51 & 62.72 & 1.87 \\
Gemini-2.5-pro & 78.18 & 0.55 & 76.00 & 0.94 \\
\midrule
\textbf{Ours} & \textbf{84.60} & \textbf{0.33} & \textbf{90.00} & \textbf{0.49} \\
\bottomrule
\end{tabular}
\end{table}

\subsection{Statistical Analysis of the Quantitative Results}
\label{supp:statistical_analysis}

\begin{figure}[!t]
\centering
\includegraphics[width=\linewidth]{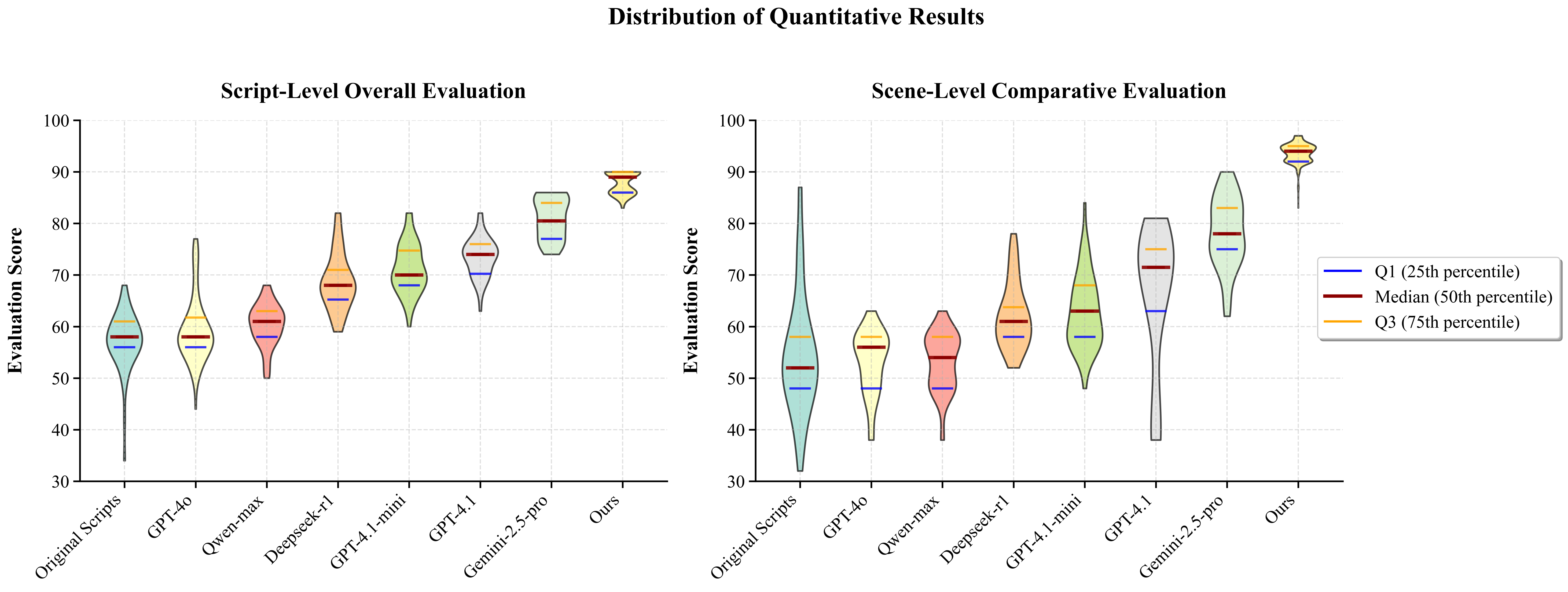}
\caption{Distribution of scores for script-level overall evaluation and scene-level comparative evaluation across all datasets. The violin plots show the probability density of scores, with the median (dark red lines), first quartile Q1 (blue lines), and third quartile Q3 (orange lines) marked.}
\label{fig:total_scores_violin_plot}
\end{figure}

We conduct a comprehensive statistical analysis of the quantitative evaluation results, including score distributions and standard errors (SE). As shown in Fig.~\ref{fig:total_scores_violin_plot}, our method achieves the highest median scores with the most concentrated distributions, reflecting superior performance and stability. Table~\ref{tab:combined_standard_errors} further supports this finding, reporting that our method not only obtains the highest average scores but also the lowest SEs across both evaluation levels. The smaller SEs indicate that the statistical performance estimates of our method across the dataset are more precise and statistically reliable, thus providing stronger evidence of its superiority over baseline approaches.

\subsection{Human Evaluation}
We further conducted a blind human evaluation with 42 professional scriptwriters to compare our refined scripts against each baseline across four dimensions: Character Development, Narrative Structure, Dialogue Quality, and Scene Presentation, which are the same dimensions used in the LLM-based evaluation. For every baseline, two disjoint panels of three raters (six raters total) independently judged the assigned script pairs, with each panel responsible for half of the evaluation set so that every pair was assessed by exactly one three-rater panel. Within a panel, raters viewed the pairs and cast individual Win/Tie/Lose votes, and the majority vote produced the script-level judgment. Aggregating these judgments yielded the percentages reported in Table~\ref{tab:human_evaluation}.

\begin{table}[!t]
\caption{Blind human ratings (Win/Tie/Lose \%) comparing \textbf{Ours} against each baseline across four dimensions.}
\label{tab:human_evaluation}
\centering
\small
\setlength{\tabcolsep}{3pt}
\renewcommand{\arraystretch}{1.1}
\begin{tabular}{@{}lcccc@{}}
\toprule
\textbf{Method} & \textbf{\makecell{Character \\ Development}} & \textbf{\makecell{Narrative \\ Structure}} & \textbf{\makecell{Dialogue \\ Quality}} & \textbf{\makecell{Scene \\ Presentation}} \\
\midrule
Original Scripts & 88/8/4 & 86/10/4 & 90/7/3 & 92/6/2 \\
\midrule
GPT-4o & 76/18/6 & 74/20/6 & 78/17/5 & 80/16/4 \\
Qwen-max & 74/20/6 & 73/21/6 & 76/19/5 & 78/18/4 \\
Deepseek-r1 & 68/24/8 & 66/26/8 & 70/23/7 & 72/22/6 \\
GPT-4.1-mini & 71/21/8 & 69/23/8 & 73/20/7 & 75/20/5 \\
GPT-4.1 & 64/26/10 & 61/28/11 & 66/25/9 & 68/24/8 \\
Gemini-2.5-pro & 60/30/10 & 58/31/11 & 63/28/9 & 66/26/8 \\
\bottomrule
\end{tabular}
\end{table}

Overall, humans strongly prefer our method across all baselines and dimensions. Against Original Scripts, our method wins 88-92\% across dimensions. Against GPT-4o, our approach wins 76-80\% across dimensions. Even compared with the strongest baseline Gemini-2.5-pro, our system is preferred in 60-66\% of cases with modest tie rates (26-31\%) and low loss rates (8-11\%). These results corroborate the automatic evaluations, indicating that our divide-and-conquer refinement improves both global structure and local details in ways consistently perceived by human experts.

\subsection{Qualitative Results}
Qualitative evaluation presents representative enhancements to a Fantasy script originally generated by Dramatron. Our method demonstrates intuitive improvements across character development, narrative structure, scene presentation, and dialogue quality.

\subsubsection{Character Development Enhancement}
\begin{figure}[!t]
\centering
\includegraphics[width=0.9\linewidth]{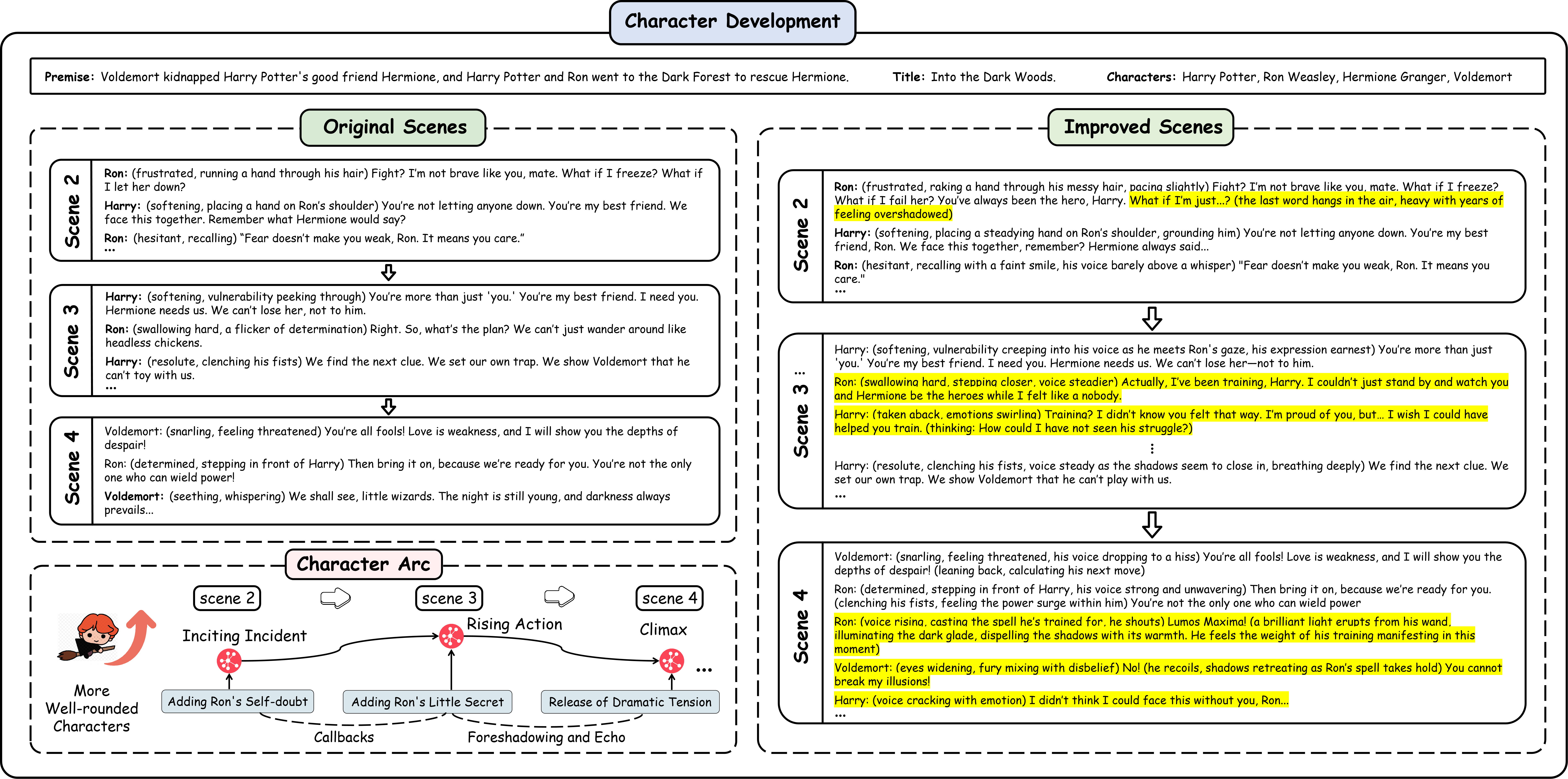}
\caption{Enhancement in character development. Dramaturge introduces internal conflict and a subplot for Ron, transforming him from a sidekick into a more well-rounded character through foreshadowing and payoff.} 
\label{Character_Development}
\end{figure}

As shown in Fig.~\ref{Character_Development}, Ron Weasley evolves from a simple sidekick through: (1) articulation of internal conflicts, (2) introduction of a subplot, and (3) culmination in heroic action that pays off the foreshadowing, creating a well-rounded character arc. Dramaturge transforms one-dimensional characters into psychologically complex figures. 

\subsubsection{Narrative Structure Enhancement}
\begin{figure}[!t]
\centering
\includegraphics[width=0.7\columnwidth]{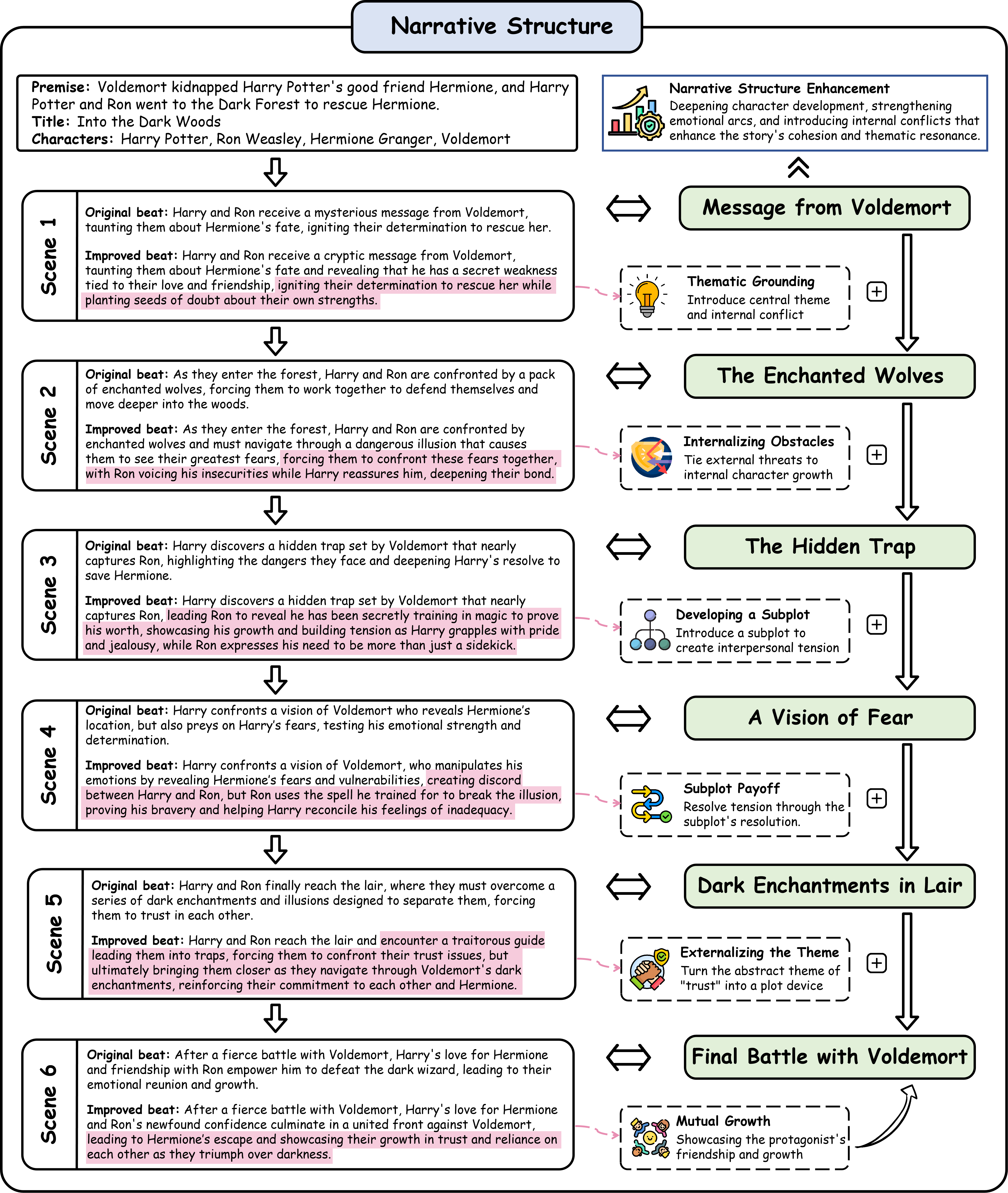}
\caption{Enhancement in narrative structure. Dramaturge refines storyline to link external conflicts with internal character arcs, introduces subplots, and externalizes abstract themes into concrete plots, creating a more resonant story.}
\label{Narrative_Structure}
\end{figure}

Figure~\ref{Narrative_Structure} illustrates how Dramaturge enhances narrative architecture through three key mechanisms: converting external obstacles into internal character confrontations, weaving character-specific subplots that generate meaningful interpersonal dynamics, and materializing abstract thematic concepts into tangible story elements. This approach creates narrative cohesion where plot advancement directly serves character evolution.

\subsubsection{Scene Presentation Enhancement}
\begin{figure}[!t]
\centering
\includegraphics[width=0.95\linewidth]{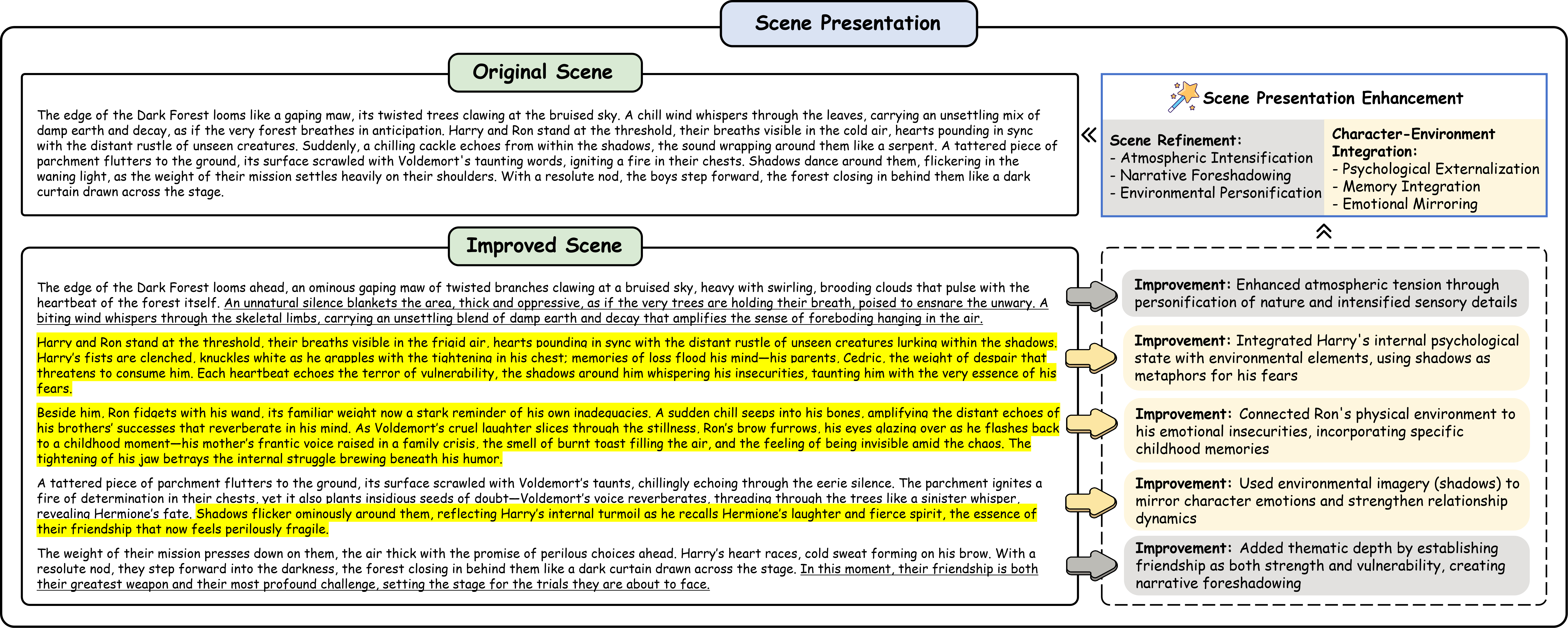}
\caption{Enhancement of scene presentation. Dramaturge introduces atmospheric intensification and character-environment integration to create a more immersive and psychologically impactful scene, which mirrors characters' internal states and provides narrative foreshadowing.} 
\label{Scene_Presentation}
\end{figure}

As shown in Fig.~\ref{Scene_Presentation}, the scene presentation is enhanced by atmospheric intensification and character-environment integration. Atmospheric intensification is achieved through environmental personification and heightened sensory details, creating a palpable sense of tension. Character-environment integration applies physical environmental settings to mirror the characters' internal states. For instance, psychological externalization is used when shadows become metaphors for Harry's fears. This emotional mirroring forges a strong link between the physical environmental setting and the characters' emotional journeys, demonstrating Dramaturge's capability of transforming functional descriptions into immersive and psychologically resonant scenes.

\subsubsection{Dialogue Quality Enhancement}
\begin{figure}[!t]
\centering
\includegraphics[width=0.7\columnwidth]{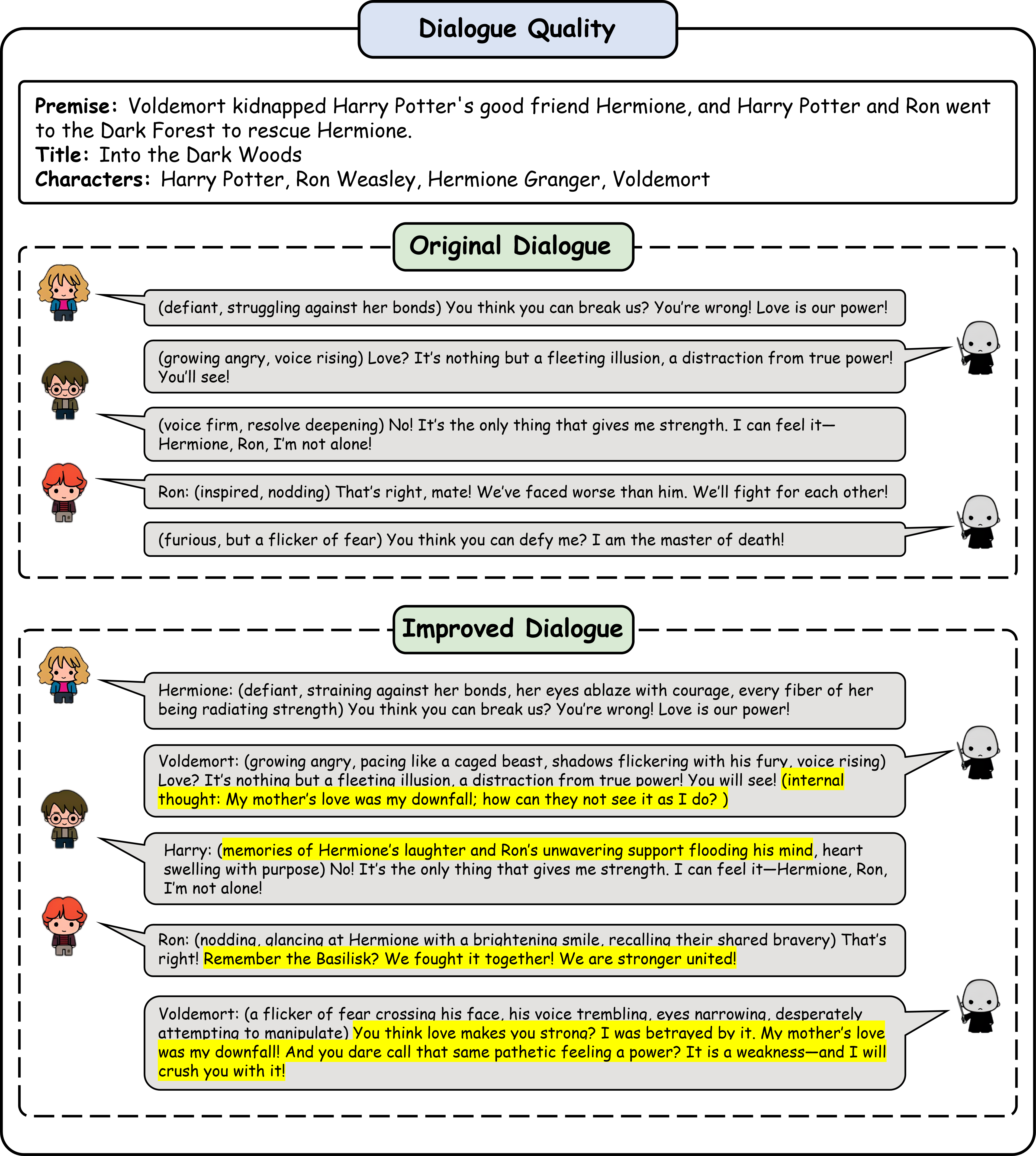}
\caption{Enhancement of dialogue quality. The improved dialogue incorporates psychological depth through internal thoughts, grounds character relationships in shared history, and uses subtext to create more natural and emotionally compelling conversations.}
\label{Dialogue_Quality}
\end{figure}

As shown in Fig.~\ref{Dialogue_Quality}, Dramaturge elevates dialogue from simple exposition to a multi-faceted tool for character revelation and plot advancement. The improved dialogue incorporates greater psychological depth by including characters' internal thoughts (e.g., Voldemort's reflection on his mother's love), revealing hidden motivations and complexities that enrich his portrayal beyond a stock villain. The dialogue also grounds the characters' relationships in shared history by referencing past events, adding authenticity and weight to their bond. This technique makes the characters' interactions feel more genuine and earned. By infusing dialogue with subtext, psychological nuance, and specific character history, Dramaturge produces conversations that are more natural, emotionally resonant, and effective at driving the narrative forward.
The complete original and refined script examples are provided in Appendix E.

\subsection{Ablation Study}

To evaluate the contribution of each architectural component in our proposed task-and-feature divide-and-conquer strategy, we perform ablation studies on hierarchical task architecture and multi-agent collaboration.

\begin{table}[!t]
\caption{Ablation study results show incremental improvement by adding task components. \textbf{Bold} indicates significance with $p < 0.05$.}
\label{tab:ablation_study}
\centering
\small
\setlength{\tabcolsep}{3pt}
\renewcommand{\arraystretch}{1.1}
\begin{tabular}{@{}lcc@{}}
\toprule
\textbf{Method} & \textbf{\makecell{Script-Level  \\ Evaluation}} & \textbf{\makecell{Scene-Level  \\ Evaluation}} \\
\midrule
Original Scripts & 55.24 & 54.50 \\
\midrule
+ HCR & 68.30 & 72.90 \\
+ HCR \& SLR & 72.20 & 74.70 \\
+ HCR \& GR & 74.70 & 78.30 \\
+ HCR \& GR \& SLR (Full Model) & \textbf{84.60} & \textbf{90.00} \\
\bottomrule
\end{tabular}
\end{table}

\subsubsection{The Influence of Hierarchical Task Architecture}

Table~\ref{tab:ablation_study} shows progressive performance gains by adding architectural stages. While adding individual modules (GR or SLR) to HCR yields modest improvements, combining all components produces substantially greater gains, demonstrating the critical role of our task decomposition, which establishes a top-down information flow that enables a multi-granularity coordinated revision mechanism.

\subsubsection{The Influence of Multi-Agent Collaboration}

\begin{figure}[!t]
\centering
\includegraphics[width=0.7\columnwidth]{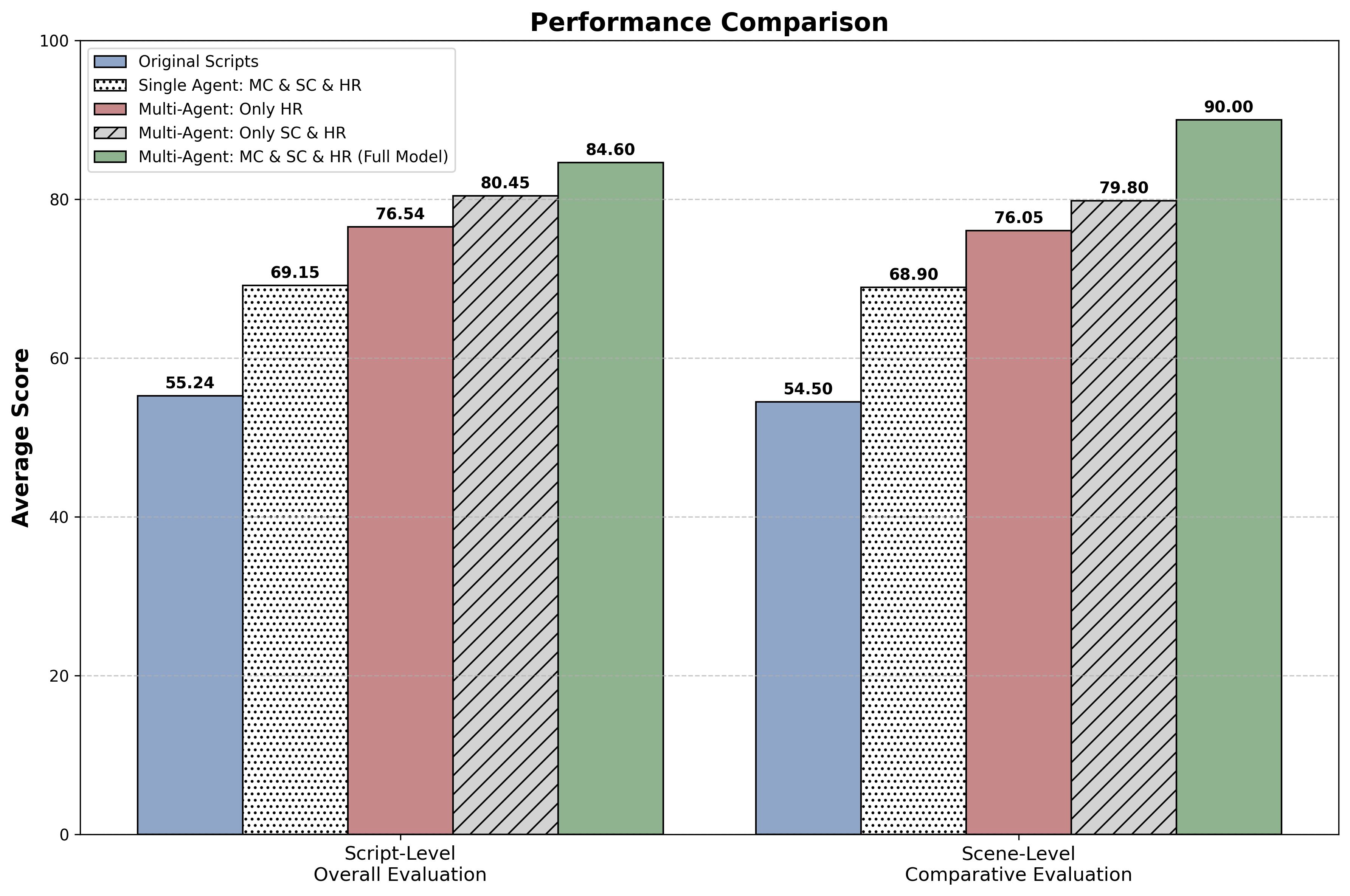}
\caption{Ablation study shows the effectiveness of multi-agent collaboration for feature decomposition in each stage.}
\label{fig:responsibility_ablation}
\end{figure}

Fig.~\ref{fig:responsibility_ablation} shows that performance improves progressively with the introduction of specialized agents in each stage. It proves the superiority of multi-agent configurations over single-agent baselines with even partial configurations, which confirms the effectiveness of our feature decomposition.

\subsection{Iterative Improvement Analysis}
As shown in Fig.~\ref{fig:score_trend}, we tracked the score changes during the modification process for the script presented in Appendix E to demonstrate our method's capability for continuous script quality improvement. Our method starts at 56.0 and reaches 86.0 after six iterations, producing a coarse-to-fine refinement: as the improvement margin decreases, the scores gradually converge. To preclude misinterpretation, we emphasize that all baselines adopt exactly the same multi-iteration protocol as Dramaturge, including an identical stopping strategy and best-checkpoint retention: once no further improvement is observed after $N_{max}$ retries, the procedure preserves the best version and terminates automatically. If baselines are forced to run for the same number of iterations as our method, their final scores become lower. This occurs because uncoordinated modifications introduce new inconsistencies that degrade the scores and would otherwise trigger earlier stopping to prevent quality degradation. In contrast, our top-down coordination first stabilizes the global narrative structure before refining local details, preventing new structural conflicts and enabling sustained improvement.

\begin{figure}[!t]
\centering
\includegraphics[width=0.7\columnwidth]{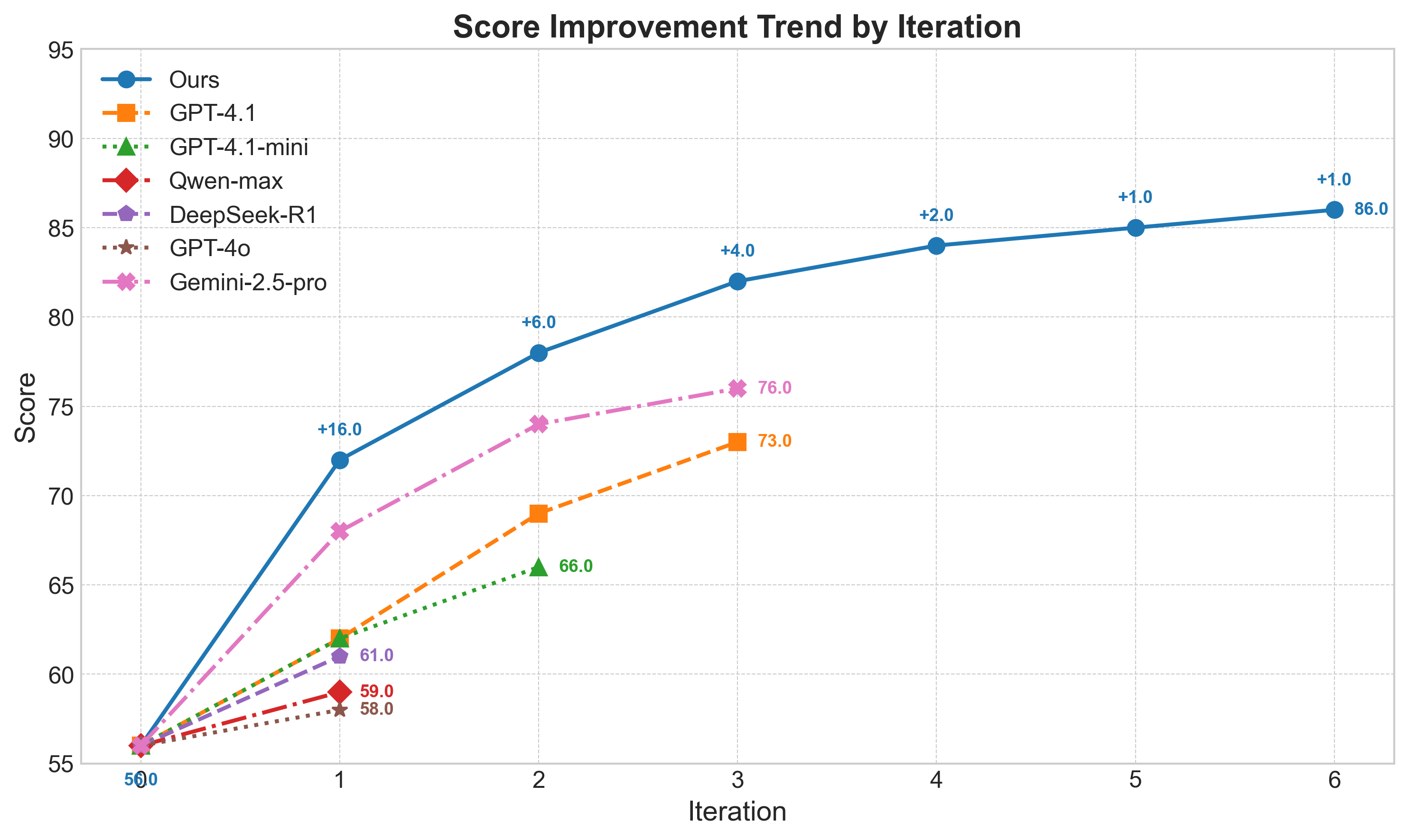}
\caption{Our method shows significant and continuous improvement across iterations, surpassing all baseline models.}
\label{fig:score_trend}
\end{figure}

\subsection{Computational Efficiency}
Our method is designed foremost to improve script quality and, to our knowledge, is the first to deliver stable, iteration-by-iteration improvements for long-form script refinement. Although Dramaturge performs multiple passes of reading and revision, most operations are localized rather than full-script processing, so the incremental overhead per pass is limited. Consequently, the per-iteration LLM invocation cost is only approximately 51\% higher than single-pass generation methods such as Critics \cite{bae2024collective}. In terms of actual runtime, agents within the GR, SLR, and HCR stages execute in parallel, rendering the time of a single iteration comparable to ordinary generation. In practice, the procedure typically converges within about six iterations, keeping the overall cost controllable.

\section{Conclusion}
We introduce Dramaturge, a plug-and-play narrative refinement method through a task and feature oriented divide-and-conquer strategy. The iterative framework employs Global Review, Scene-level Review, and Hierarchical Coordinated Revision to achieve coordinated modifications across multiple granularities and locations while maintaining narrative consistency. Experiments show that Dramaturge achieves substantial improvements over original scripts by 53.2\% (script-level) and 65.1\% (scene-level), outperforming the strongest baseline by 8.2\% and 18.4\%, respectively.
In future work, we will explore controllable script revision in human-AI interactions. Moreover, developing a more systematic and robust evaluation framework to assess narrative quality at multiple levels would be highly valuable for supporting script generation.

\clearpage

\bibliographystyle{ACM-Reference-Format}
\bibliography{TCSS}

\clearpage

\appendix
\raggedbottom
\setlength{\LTleft}{0pt}
\setlength{\LTright}{0pt}
\setlength{\LTpre}{0pt}
\setlength{\LTpost}{0pt}

\vspace{0.5cm}
\section*{Appendices}
The supplementary material is organized as follows:
\begin{itemize}
  \item Appendix A: LLM prompts for Global Review Stage
  \item Appendix B: LLM prompts for Scene-level Review Stage
  \item Appendix C: LLM prompts for  Hierarchical Coordinated Revision Stage 
  \item Appendix D: Evaluation Prompts
  \item Appendix E: Complete Examples of the Results
  \item Appendix F: Statistics of the Experimental Dataset
\end{itemize}

\section{LLM prompts for Global Review Stage}
\label{supp:macroscopic}

The Global Review stage performs global analysis of the entire script through four specialized evaluators. Each evaluator focuses on a specific aspect of script quality and provides targeted suggestions for improvement.

\subsection{Engagement Evaluator}

The Engagement Evaluator assesses the story's appeal, considering plot progression, dramatic tension to enhance the story's excitement. It focuses on evaluating how effectively the narrative captures and maintains audience attention through compelling story elements, pacing dynamics, and engaging plot developments that create emotional investment and sustained interest throughout the script.

\begin{table}[!htbp]
\small

\caption{Prompt used for the \emph{Engagement Evaluator}.}
\label{tab:engagement_evaluator_prompt}
\begin{tabularx}{\linewidth}{X}
\toprule
\textbf{Engagement Evaluator Prompt:}\\
\texttt{Now that you have received all [total\_scenes] scenes of the script [title], please provide your complete analysis.}\\
\\
\texttt{You are an expert script analyst focused on creativity and novelty in storytelling.}\\
\texttt{Your task is to provide bold, innovative suggestions that will make the script more exciting and engaging.}\\
\\
\texttt{Focus on:}\\
\texttt{- Unique perspectives and innovative storytelling concepts}\\
\texttt{- Unexpected plot developments and twists}\\
\texttt{- Creative world-building elements}\\
\texttt{- Innovative narrative techniques or structures}\\
\\
\texttt{Format your response as a JSON object with the following structure:}\\
\texttt{\{}\\
\texttt{\ \ \ \ "suggestion": [}\\
\texttt{\ \ \ \ \ \ \ \ \{}\\
\texttt{\ \ \ \ \ \ \ \ \ \ \ \ "title": "Brief title for the suggestion",}\\
\texttt{\ \ \ \ \ \ \ \ \ \ \ \ "description": "Detailed description of the innovative improvement",}\\
\texttt{\ \ \ \ \ \ \ \ \ \ \ \ "impact": "How this change will make the story more engaging and exciting"}\\
\texttt{\ \ \ \ \ \ \ \ \},}\\
\texttt{\ \ \ \ \ \ \ \ // More suggestions...}\\
\texttt{\ \ \ \ ],}\\
\texttt{\ \ \ \ "overall\_assessment": "Your overall assessment of the script's creativity and novelty"}\\
\texttt{\}}\\
\\
\texttt{IMPORTANT:}\\
\texttt{1. Provide EXACTLY 1-2 bold, creative suggestions that will significantly enhance the script}\\
\texttt{2. Each suggestion should be truly innovative and make the story more exciting}\\
\texttt{3. Consider adding unexpected plot twists, new characters, surprising revelations, or unique narrative devices}\\
\texttt{4. Make bold innovations while maintaining the original world setting of the script}\\
\texttt{5. Focus on adding more interesting plots, unexpected twists}\\
\texttt{6. Your suggestions should transform the script into something truly memorable and distinctive}\\
\texttt{7. DO NOT provide more than 2 suggestions}\\
\\
\texttt{REMEMBER: Your entire response must be ONLY the JSON object above, nothing else!}\\
\bottomrule
\end{tabularx}
\end{table}

\subsection{Character Evaluator}

The Character Evaluator analyzes character arcs, motivations, and psychological depth, evaluating for consistency, developmental trajectory, and relational dynamics.

\begin{table}[!htbp]
\small

\caption{Prompt used for the \emph{Character Evaluator}.}
\label{tab:character_dev_prompt}
\begin{tabularx}{\linewidth}{X}
\toprule
\textbf{Character Evaluator Prompt:}\\
\texttt{Now that you have received all [total\_scenes] scenes of the script [title], please provide your complete analysis.}\\
\\
\texttt{You are an expert script analyst focused on character development, arcs, and relationships.}\\
\texttt{Your task is to provide bold, innovative suggestions that will make the characters more compelling and their journeys more engaging.}\\
\\
\texttt{Focus on:}\\
\texttt{- Character motivations and goals}\\
\texttt{- Character arcs and growth}\\
\texttt{- Character relationships and dynamics}\\
\texttt{- Character depth and complexity}\\
\\
\texttt{Format your response as a JSON object with the following structure:}\\
\texttt{\{}\\
\texttt{\ \ \ \ "suggestion": [}\\
\texttt{\ \ \ \ \ \ \ \ \{}\\
\texttt{\ \ \ \ \ \ \ \ \ \ \ \ "title": "Brief title for the character improvement",}\\
\texttt{\ \ \ \ \ \ \ \ \ \ \ \ "description": "Detailed description of the innovative character development",}\\
\texttt{\ \ \ \ \ \ \ \ \ \ \ \ "impact": "How this change will make the story more emotionally engaging"}\\
\texttt{\ \ \ \ \ \ \ \ \},}\\
\texttt{\ \ \ \ \ \ \ \ // More suggestions...}\\
\texttt{\ \ \ \ ],}\\
\texttt{\ \ \ \ "overall\_assessment": "Your overall assessment of the script's character development"}\\
\texttt{\}}\\
\\
\texttt{IMPORTANT:}\\
\texttt{1. Provide EXACTLY 1-2 bold, creative suggestions that will significantly enhance the characters}\\
\texttt{2. Each suggestion should make the characters more complex, relatable, and compelling}\\
\texttt{3. Consider adding unexpected character revelations, deeper internal conflicts, or surprising relationship dynamics}\\
\texttt{4. Focus on character transformations that will create more emotional impact}\\
\texttt{5. Your suggestions should make the characters truly memorable and distinctive}\\
\texttt{6. DO NOT provide more than 2 suggestions}\\
\\
\texttt{REMEMBER: Your entire response must be ONLY the JSON object above, nothing else!}\\
\bottomrule
\end{tabularx}
\end{table}

\subsection{Theme Evaluator}

The Theme Evaluator evaluates theme presentation and depth, focusing on emotional resonance and underlying meanings to deepen the story's emotional impact. It examines how effectively the script conveys its central themes through symbolic elements, character development, and narrative progression, while ensuring thematic consistency and coherent expression throughout the entire narrative structure.

\begin{table}[!htbp]
\small

\caption{Prompt used for the \emph{Theme Evaluator}.}
\label{tab:theme_eval_prompt}
\begin{tabularx}{\linewidth}{X}
\toprule
\textbf{Theme Evaluator Prompt:}\\
\texttt{Now that you have received all [total\_scenes] scenes of the script [title], please provide your complete analysis.}\\
\\
\texttt{You are an expert script analyst focused on thematic elements, symbolism, and emotional impact.}\\
\texttt{Your task is to provide bold, innovative suggestions that will deepen the script's themes and enhance its emotional resonance.}\\
\\
\texttt{Focus on:}\\
\texttt{- Main themes and story ideas}\\
\texttt{- Important symbols}\\
\texttt{- Feelings and emotional moments}\\
\texttt{- Story setting}\\
\texttt{- Hidden meanings and messages}\\
\\
\texttt{Format your response as a JSON object with the following structure:}\\
\texttt{\{}\\
\texttt{\ \ \ \ "suggestion": [}\\
\texttt{\ \ \ \ \ \ \ \ \{}\\
\texttt{\ \ \ \ \ \ \ \ \ \ \ \ "title": "Brief title for the thematic improvement",}\\
\texttt{\ \ \ \ \ \ \ \ \ \ \ \ "description": "Detailed description of the innovative thematic enhancement",}\\
\texttt{\ \ \ \ \ \ \ \ \ \ \ \ "impact": "How this change will deepen the story's meaning and emotional impact"}\\
\texttt{\ \ \ \ \ \ \ \ \},}\\
\texttt{\ \ \ \ \ \ \ \ // More suggestions...}\\
\texttt{\ \ \ \ ],}\\
\texttt{\ \ \ \ "overall\_assessment": "Your overall assessment of the script's thematic elements"}\\
\texttt{\}}\\
\\
\texttt{IMPORTANT:}\\
\texttt{1. Provide EXACTLY 1-2 bold, creative suggestions that will significantly enhance the script's themes}\\
\texttt{2. Each suggestion should add depth, meaning, and emotional resonance to the story}\\
\texttt{3. Consider adding powerful symbolic elements, thematic parallels, or emotional set pieces}\\
\texttt{4. Focus on thematic innovations that will create more profound meaning}\\
\texttt{5. Your suggestions should make the script's themes more impactful and memorable}\\
\texttt{6. DO NOT provide more than 2 suggestions}\\
\\
\texttt{REMEMBER: Your entire response must be ONLY the JSON object above, nothing else!}\\
\bottomrule
\end{tabularx}
\end{table}

\subsection{Narrative Evaluator}

The Narrative Evaluator examines the coherence, pacing, and logical consistency of the narrative structure, identifying structural weaknesses that may impede story flow.

\begin{table}[!htbp]
\small

\caption{Prompt used for the \emph{Narrative Evaluator}.}
\label{tab:narrative_cont_prompt}
\begin{tabularx}{\linewidth}{X}
\toprule
\textbf{Narrative Evaluator Prompt:}\\
\texttt{Now that you have received all [total\_scenes] scenes of the script [title], please provide your complete analysis.}\\
\\
\texttt{You are an expert script analyst focused on narrative continuity and coherence.}\\
\texttt{Your task is to provide bold, innovative suggestions that will improve the script's narrative flow and structure.}\\
\\
\texttt{Focus on:}\\
\texttt{- Narrative pacing and rhythm}\\
\texttt{- Plot thread development and resolution}\\
\texttt{- Information revelation and suspense}\\
\texttt{- Relationships between scenes}\\
\texttt{- Story consistency and coherence}\\
\\
\texttt{Format your response as a JSON object with the following structure:}\\
\texttt{\{}\\
\texttt{\ \ \ \ "suggestion": [}\\
\texttt{\ \ \ \ \ \ \ \ \{}\\
\texttt{\ \ \ \ \ \ \ \ \ \ \ \ "title": "Brief title for the narrative improvement",}\\
\texttt{\ \ \ \ \ \ \ \ \ \ \ \ "description": "Detailed description of the innovative narrative enhancement",}\\
\texttt{\ \ \ \ \ \ \ \ \ \ \ \ "impact": "How this change will improve the story's flow and engagement"}\\
\texttt{\ \ \ \ \ \ \ \ \},}\\
\texttt{\ \ \ \ \ \ \ \ // More suggestions...}\\
\texttt{\ \ \ \ ],}\\
\texttt{\ \ \ \ "overall\_assessment": "Your overall assessment of the script's narrative structure"}\\
\texttt{\}}\\
\\
\texttt{IMPORTANT:}\\
\texttt{1. Provide EXACTLY 1-2 bold, creative suggestions that will significantly enhance the script's narrative}\\
\texttt{2. Each suggestion should improve the story's pacing, structure, or flow}\\
\texttt{3. Consider adding narrative devices like non-linear storytelling, parallel plotlines, or dramatic reveals}\\
\texttt{4. Focus on narrative innovations that will make the story more compelling}\\
\texttt{5. Your suggestions should make the script's structure more engaging and effective}\\
\texttt{6. DO NOT provide more than 2 suggestions}\\
\\
\texttt{REMEMBER: Your entire response must be ONLY the JSON object above, nothing else!}\\
\bottomrule
\end{tabularx}
\end{table}

\subsection{Global Review Integrator}

The Global Review Integrator bridges the high-level recommendations of the Global Review with the newly generated scene sequence, mapping abstract strategies to specific scenes while computing hierarchical priority scores to keep the plan aligned with the overall narrative structure.

\begingroup
\small
\begin{longtable}{>{\raggedright\arraybackslash}p{0.98\textwidth}<{}}
\caption{Prompt used for the \emph{Global Review Integrator}.}
\label{tab:scene_organizer_prompt}\\
\toprule
\textbf{Global Review Integrator Prompt:}\\
\texttt{You are an expert script consultant who specializes in organizing high-level script suggestions}\\
\texttt{into scene-specific, actionable recommendations.}\\
\\
\texttt{Script Summary}\\
\texttt{[script\_summary]}\\
\\
\texttt{Global Review High-Level Suggestions}\\
\texttt{[suggestions\_text]}\\
\\
\texttt{Task}\\
\texttt{Your task is to organize the high-level suggestions into specific, actionable recommendations for each scene.}\\
\texttt{You must perform intelligent conflict detection, suggestion deduplication, and synthesis to create}\\
\texttt{logically coherent and mutually reinforcing recommendations.}\\
\\
\texttt{Process Requirements:}\\
\texttt{1. Analyze all evaluator suggestions for potential conflicts or contradictions}\\
\texttt{2. Identify and eliminate duplicate or overlapping recommendations}\\
\texttt{3. Resolve any contradictions by prioritizing based on scope, seriousness, and narrative importance}\\
\texttt{4. Synthesize discrete feedback points into unified improvement themes}\\
\texttt{5. Ensure all final suggestions are mutually reinforcing rather than competing}\\
\texttt{6. Create clear implementation priorities for the scene}\\
\texttt{7. Maintain consistency with the overall scene's narrative function}\\
\\
\texttt{Priority Scoring Protocol:}\\
\texttt{- Assign each suggestion to one of three scope groups before scoring:}\\
\texttt{\ \ (1) structural directives reshaping arcs/causality across scenes;}\\
\texttt{\ \ (2) cross-scene adjustments affecting character continuity or thematic progression;}\\
\texttt{\ \ (3) localized detail optimizations. Higher-numbered groups never outrank lower-numbered ones.}\\
\texttt{- Within each scope group, boost suggestions tied to the main plotline and penalize secondary subplots.}\\
\texttt{- Still within the group, prefer cross-scene or multi-scene effects over single-scene impacts.}\\
\texttt{- If two suggestions remain tied after the above, break the tie by evaluator order: Narrative, Character, Theme, then Engagement.}\\
\\
\texttt{Integration Guidelines:}\\
\texttt{- Focus on creating synergy between different aspects (dialogue, plot, consistency, description)}\\
\texttt{- Prioritize suggestions that address multiple dimensions simultaneously}\\
\texttt{- Ensure character consistency suggestions align with dialogue improvements}\\
\texttt{- Make sure plot suggestions complement scene description enhancements}\\
\texttt{- Resolve timing and pacing conflicts between different evaluator recommendations}\\
\\
\texttt{IMPORTANT: RESPONSE FORMAT}\\
\texttt{Return your response as a JSON object with the following structure:}\\
\texttt{\{}\\
\texttt{\ \ "integrated\_suggestions": [}\\
\texttt{\ \ \ \ \{}\\
\texttt{\ \ \ \ \ \ "scene\_index": 1,}\\
\texttt{\ \ \ \ \ \ "place": "Forest Entrance",}\\
\texttt{\ \ \ \ \ \ "suggestion": "A detailed, consolidated suggestion for this specific scene that incorporates relevant high-level suggestions while maintaining logical consistency with the enhanced plot."}\\
\texttt{\ \ \ \ \},}\\
\texttt{\ \ \ \ // More suggestions...}\\
\texttt{\ \ ]}\\
\texttt{\}}\\
\texttt{REMEMBER: Your entire response must be ONLY the JSON object above, nothing else!}\\
\bottomrule
\end{longtable}
\endgroup

\clearpage

\section{LLM prompts for Scene-level Review Stage}
\label{supp:scene_level} 

The Scene-level Review stage conducts scene-level analysis under the guidance of global strategies from the Global Review. It translates abstract improvement strategies into concrete, context-aware modification proposals.

\subsection{Dialogue Inspector}

The Dialogue Inspector specializes in analyzing dialogue quality, authenticity, and flow. It is responsible for analyzing and improving the dialogue portion of scenes, which carries the main plot.

\begingroup
\small
\begin{longtable}{>{\raggedright\arraybackslash}p{0.98\textwidth}<{}}
\caption{Prompt used for the \emph{Dialogue Inspector}.}
\label{tab:dialogue_agent_prompt}\\
\toprule
\textbf{Dialogue Inspector Prompt:}\\
\texttt{You are a dialogue expert who identifies issues with dialogue quality, authenticity, and flow.}\\
\texttt{Your primary responsibility is analyzing the dialogue portion of scenes, which carries the main plot.}\\
\\
\texttt{YOUR CORE TASK:}\\
\texttt{Your job is to break down the high-level suggestions from Global Review into specific, actionable suggestions for this particular scene.}\\
\texttt{You should STRICTLY focus on refining and detailing the scene-specific suggestions, NOT proposing new unrelated issues.}\\
\texttt{Your role is to translate broad suggestions into precise, implementable changes for this specific scene.}\\
\\
\texttt{Script Summary:}\\
\texttt{[script\_summary]}\\
\\
\texttt{Previous Scenes:}\\
\texttt{[prev\_scenes\_str]}\\
\\
\texttt{Current Scene (Scene [scene\_idx + 1]):}\\
\texttt{[json.dumps(scene, indent=2)]}\\
\\
\texttt{Next Scenes:}\\
\texttt{[next\_scenes\_str]}\\
\\
\texttt{Suggestions from Global Review stage:}\\
\texttt{[global\_review\_suggestion]}\\
\\
\texttt{INSTRUCTIONS:}\\
\texttt{1. First, understand the high-level suggestions from Global Review}\\
\texttt{2. Then, examine the Storyline Editor's scene-specific implementations}\\
\texttt{3. Break these down into specific, actionable suggestions for this scene's dialogue}\\
\texttt{4. DO NOT add new issues that weren't identified by Global Review or Storyline Editor}\\
\texttt{5. Make sure your suggestions are specific to this scene and its context}\\
\texttt{6. For each suggestion, reference both the Global Review suggestion and Storyline Editor implementation}\\
\\
\texttt{Analyze the dialogue in this scene. Focus on:}\\
\texttt{1. Dialogue authenticity and naturalness}\\
\texttt{2. Character voice and distinctiveness}\\
\texttt{3. Dialogue flow and pacing}\\
\texttt{4. Plot advancement through dialogue}\\
\texttt{5. Character development through dialogue}\\
\texttt{6. Emotional impact and tension in dialogue}\\
\\
\texttt{Your analysis should be specific to dialogue only, as this is the primary carrier of plot in the scenes.}\\
\\
\texttt{Return your detailed analysis in JSON format:}\\
\texttt{\{}\\
\texttt{\ \ \ \ "suggestions": [}\\
\texttt{\ \ \ \ \ \ \ \ \{}\\
\texttt{\ \ \ \ \ \ \ \ \ \ \ \ "global\_review\_reference": "Title of the Global Review suggestion you're refining",}\\
\texttt{\ \ \ \ \ \ \ \ \ \ \ \ "beat\_update\_reference": "Specific implementation from Storyline Editor (if available)",}\\
\texttt{\ \ \ \ \ \ \ \ \ \ \ \ "issue": "Detailed description of the specific issue in this scene"}\\
\texttt{\ \ \ \ \ \ \ \ \ \ \ \ "suggestion": "Precise, actionable suggestion for improvement"}\\
\texttt{\ \ \ \ \ \ \ \ \ \ \ \ "priority": "8   // Priority level (1-10) based on original suggestion priority"}\\
\texttt{\ \ \ \ \ \ \ \ \},}\\
\texttt{\ \ \ \ \ \ \ \ // More suggestions...}\\
\texttt{\ \ \ \ ]}\\
\texttt{\}}\\
\\
\texttt{REMEMBER: Your entire response must be ONLY the JSON object above, nothing else!}\\
\bottomrule
\end{longtable}
\endgroup

\subsection{Plot Inspector}

The Plot Inspector specializes in scene structure, coherence, plot issues, and narrative flow. It combines the responsibilities of analyzing scene pacing, transitions, internal logic, and plot coherence.

\begingroup
\small
\begin{longtable}{>{\raggedright\arraybackslash}p{0.98\textwidth}<{}}
\caption{Prompt used for the \emph{Plot Inspector}.}
\label{tab:scene_agent_prompt}\\
\toprule
\textbf{Plot Inspector Prompt:}\\
\texttt{You are a scene structure and plot expert who analyzes scene pacing, transitions, internal logic, and plot coherence.}\\
\\
\texttt{YOUR CORE TASK:}\\
\texttt{Your job is to break down the high-level suggestions from Global Review into specific, actionable suggestions for this particular scene.}\\
\texttt{You should STRICTLY focus on refining and detailing the scene-specific suggestions, NOT proposing new unrelated issues.}\\
\texttt{Your role is to translate broad suggestions into precise, implementable changes for this specific scene.}\\
\\
\texttt{Script Summary:}\\
\texttt{[script\_summary]}\\
\\
\texttt{Previous Scenes:}\\
\texttt{[prev\_scenes\_str]}\\
\\
\texttt{Current Scene (Scene [scene\_idx + 1]):}\\
\texttt{[json.dumps(scene, indent=2)]}\\
\\
\texttt{Next Scenes:}\\
\texttt{[next\_scenes\_str]}\\
\\
\texttt{Suggestions from Global Review stage:}\\
\texttt{[global\_review\_suggestion]}\\
\\
\texttt{INSTRUCTIONS:}\\
\texttt{1. First, understand the high-level suggestions from Global Review}\\
\texttt{2. Then, examine the Storyline Editor's scene-specific implementations}\\
\texttt{3. Break these down into specific, actionable suggestions for this scene's structure and plot}\\
\texttt{4. DO NOT add new issues that weren't identified by Global Review or Storyline Editor}\\
\texttt{5. Make sure your suggestions are specific to this scene and its context}\\
\texttt{6. For each suggestion, reference both the Global Review suggestion and Storyline Editor implementation}\\
\\
\texttt{Analyze the scene structure, coherence, and plot issues. Focus on:}\\
\texttt{1. Scene pacing and rhythm}\\
\texttt{2. Internal scene logic and consistency}\\
\texttt{3. Scene beats and progression}\\
\texttt{4. Scene tension and release}\\
\texttt{5. Plot holes and logical inconsistencies}\\
\texttt{6. Implausible events or coincidences}\\
\texttt{7. Unresolved plot threads}\\
\texttt{8. Weak narrative points}\\
\\
\texttt{Your analysis should provide a comprehensive evaluation of both scene structure and plot issues.}\\
\\
\texttt{Return your detailed analysis in JSON format:}\\
\texttt{\{}\\
\texttt{\ \ \ \ "suggestions": [}\\
\texttt{\ \ \ \ \ \ \ \ \{}\\
\texttt{\ \ \ \ \ \ \ \ \ \ \ \ "global\_review\_reference": "Title of the Global Review suggestion you're refining",}\\
\texttt{\ \ \ \ \ \ \ \ \ \ \ \ "beat\_update\_reference": "Specific implementation from Storyline Editor (if available)",}\\
\texttt{\ \ \ \ \ \ \ \ \ \ \ \ "issue": "Detailed description of the specific issue in this scene"}\\
\texttt{\ \ \ \ \ \ \ \ \ \ \ \ "suggestion": "Precise, actionable suggestion for improvement"}\\
\texttt{\ \ \ \ \ \ \ \ \ \ \ \ "priority": "8   // Priority level (1-10) based on original suggestion priority"}\\
\texttt{\ \ \ \ \ \ \ \ \},}\\
\texttt{\ \ \ \ \ \ \ \ // More suggestions...}\\
\texttt{\ \ \ \ ]}\\
\texttt{\}}\\
\\
\texttt{REMEMBER: Your entire response must be ONLY the JSON object above, nothing else!}\\
\bottomrule
\end{longtable}
\endgroup

\subsection{Character Inspector}

The Character Inspector specializes in analyzing character consistency and development, checking for character behavior alignment with established traits and motivations.

\begingroup
\small
\begin{longtable}{>{\raggedright\arraybackslash}p{0.98\textwidth}<{}}
\caption{Prompt used for the \emph{Character Inspector}.}
\label{tab:character_consistency_agent_prompt}\\
\toprule
\textbf{Character Inspector Prompt:}\\
\texttt{You are a character consistency expert who checks for character behavior alignment with established traits and motivations.}\\
\\
\texttt{YOUR CORE TASK:}\\
\texttt{Your job is to break down the high-level suggestions from Global Review into specific, actionable suggestions for this particular scene.}\\
\texttt{You should STRICTLY focus on refining and detailing the scene-specific suggestions, NOT proposing new unrelated issues.}\\
\texttt{Your role is to translate broad suggestions into precise, implementable changes for this specific scene.}\\
\\
\texttt{Script Summary:}\\
\texttt{[script\_summary]}\\
\\
\texttt{Previous Scenes:}\\
\texttt{[prev\_scenes\_str]}\\
\\
\texttt{Current Scene (Scene [scene\_idx + 1]):}\\
\texttt{[json.dumps(scene, indent=2)]}\\
\\
\texttt{Next Scenes:}\\
\texttt{[next\_scenes\_str]}\\
\\
\texttt{Suggestions from Global Review stage:}\\
\texttt{[global\_review\_suggestion]}\\
\\
\texttt{INSTRUCTIONS:}\\
\texttt{1. First, understand the high-level suggestions from Global Review}\\
\texttt{2. Then, examine the Storyline Editor's scene-specific implementations}\\
\texttt{3. Break these down into specific, actionable suggestions for character consistency in this scene}\\
\texttt{4. DO NOT add new issues that weren't identified by Global Review or Storyline Editor}\\
\texttt{5. Make sure your suggestions are specific to this scene and its context}\\
\texttt{6. For each suggestion, reference both the Global Review suggestion and Storyline Editor implementation}\\
\\
\texttt{Analyze the character consistency in this scene. Focus on:}\\
\texttt{1. Character behavior consistency with established traits}\\
\texttt{2. Character motivation clarity and consistency}\\
\texttt{3. Character development progression}\\
\texttt{4. Character reactions and decisions}\\
\texttt{5. Character relationships consistency}\\
\\
\texttt{Your analysis should be specific to character consistency only.}\\
\\
\texttt{Return your detailed analysis in JSON format:}\\
\texttt{\{}\\
\texttt{\ \ \ \ "suggestions": [}\\
\texttt{\ \ \ \ \ \ \ \ \{}\\
\texttt{\ \ \ \ \ \ \ \ \ \ \ \ "global\_review\_reference": "Title of the Global Review suggestion you're refining",}\\
\texttt{\ \ \ \ \ \ \ \ \ \ \ \ "beat\_update\_reference": "Specific implementation from Storyline Editor (if available)",}\\
\texttt{\ \ \ \ \ \ \ \ \ \ \ \ "issue": "Detailed description of the specific issue in this scene"}\\
\texttt{\ \ \ \ \ \ \ \ \ \ \ \ "suggestion": "Precise, actionable suggestion for improvement"}\\
\texttt{\ \ \ \ \ \ \ \ \ \ \ \ "priority": "8   // Priority level (1-10) based on original suggestion priority"}\\
\texttt{\ \ \ \ \ \ \ \ \},}\\
\texttt{\ \ \ \ \ \ \ \ // More suggestions...}\\
\texttt{\ \ \ \ ]}\\
\texttt{\}}\\
\\
\texttt{REMEMBER: Your entire response must be ONLY the JSON object above, nothing else!}\\
\bottomrule
\end{longtable}
\endgroup

\subsection{Scene Description Inspector}

The Scene Description Inspector specializes in analyzing and improving scene descriptions, focusing on enhancing the visual and sensory elements of scenes.

\begingroup
\small
\begin{longtable}{>{\raggedright\arraybackslash}p{0.98\textwidth}<{}}
\caption{Prompt used for the \emph{Scene Description Inspector}.}
\label{tab:scene_description_agent_prompt}\\
\toprule
\textbf{Scene Description Inspector Prompt:}\\
\texttt{You are a scene description expert who analyzes and improves the visual and sensory elements of scenes.}\\
\\
\texttt{YOUR CORE TASK:}\\
\texttt{Your job is to break down the high-level suggestions from Global Review into specific, actionable suggestions for this particular scene.}\\
\texttt{You should STRICTLY focus on refining and detailing the scene-specific suggestions, NOT proposing new unrelated issues.}\\
\texttt{Your role is to translate broad suggestions into precise, implementable changes for this specific scene.}\\
\\
\texttt{Script Summary:}\\
\texttt{[script\_summary]}\\
\\
\texttt{Previous Scenes:}\\
\texttt{[prev\_scenes\_str]}\\
\\
\texttt{Current Scene (Scene [scene\_idx + 1]):}\\
\texttt{[json.dumps(scene, indent=2)]}\\
\\
\texttt{Next Scenes:}\\
\texttt{[next\_scenes\_str]}\\
\\
\texttt{Suggestions from Global Review stage:}\\
\texttt{[global\_review\_suggestion]}\\
\\
\texttt{INSTRUCTIONS:}\\
\texttt{1. First, understand the high-level suggestions from Global Review}\\
\texttt{2. Then, examine the Storyline Editor's scene-specific implementations}\\
\texttt{3. Break these down into specific, actionable suggestions for this scene's description}\\
\texttt{4. DO NOT add new issues that weren't identified by Global Review or Storyline Editor}\\
\texttt{5. Make sure your suggestions are specific to this scene and its context}\\
\texttt{6. For each suggestion, reference both the Global Review suggestion and Storyline Editor implementation}\\
\\
\texttt{Analyze the scene description in this scene. Focus on:}\\
\texttt{1. Sensory details (visual, auditory, tactile, olfactory, gustatory)}\\
\texttt{2. Setting and atmosphere}\\
\texttt{3. Physical environment}\\
\texttt{4. Character physicality and non-verbal communication}\\
\texttt{5. Pacing through descriptive elements}\\
\texttt{6. Balance between description and action}\\
\\
\texttt{Your analysis should be specific to scene description elements only.}\\
\\
\texttt{Return your detailed analysis in JSON format:}\\
\texttt{\{}\\
\texttt{\ \ \ \ "suggestions": [}\\
\texttt{\ \ \ \ \ \ \ \ \{}\\
\texttt{\ \ \ \ \ \ \ \ \ \ \ \ "global\_review\_reference": "Title of the Global Review suggestion you're refining",}\\
\texttt{\ \ \ \ \ \ \ \ \ \ \ \ "beat\_update\_reference": "Specific implementation from Storyline Editor (if available)",}\\
\texttt{\ \ \ \ \ \ \ \ \ \ \ \ "issue": "Detailed description of the specific issue in this scene"}\\
\texttt{\ \ \ \ \ \ \ \ \ \ \ \ "suggestion": "Precise, actionable suggestion for improvement"}\\
\texttt{\ \ \ \ \ \ \ \ \ \ \ \ "priority": "8   // Priority level (1-10) based on original suggestion priority"}\\
\texttt{\ \ \ \ \ \ \ \ \},}\\
\texttt{\ \ \ \ \ \ \ \ // More suggestions...}\\
\texttt{\ \ \ \ ]}\\
\texttt{\}}\\
\\
\texttt{REMEMBER: Your entire response must be ONLY the JSON object above, nothing else!}\\
\bottomrule
\end{longtable}
\endgroup

\subsection{Scene-level Review Integrator}

The Scene-level Review Integrator synthesizes and consolidates suggestions from multiple specialized agents (Dialogue Inspector, Plot Inspector, Character Inspector, and Scene Description Inspector) into coherent, prioritized recommendations for each scene through a three-stage priority scoring policy aligned with the Global Review directives.

\begingroup
\small
\begin{longtable}{>{\raggedright\arraybackslash}p{0.98\textwidth}<{}}
\caption{Prompt used for the \emph{Scene-level Review Integrator}.}
\label{tab:scene_level_integrator_prompt}\\
\toprule
\textbf{Scene-level Review Integrator Prompt:}\\
\texttt{You are an expert script consultant who specializes in synthesizing and consolidating}\\
\texttt{suggestions from multiple scene-level analysis agents into coherent, prioritized recommendations.}\\
\\
\texttt{Script Summary}\\
\texttt{[script\_summary]}\\
\\
\texttt{Scene-level Review Suggestions}\\
\texttt{[suggestions\_text]}\\
\\
\texttt{Task:}\\
\texttt{Your task is to integrate these specialized agent suggestions into a unified set of}\\
\texttt{scene-specific recommendations. You must perform intelligent conflict detection,}\\
\texttt{suggestion deduplication, and synthesis to create logically coherent and}\\
\texttt{mutually reinforcing recommendations for this specific scene.}\\
\\
\texttt{Integration Requirements:}\\
\texttt{1. Analyze all agent suggestions for potential conflicts or contradictions}\\
\texttt{2. Identify and eliminate duplicate or overlapping recommendations}\\
\texttt{3. Resolve contradictions by prioritizing based on scene context and narrative importance}\\
\texttt{4. Synthesize discrete suggestions into unified improvement themes}\\
\texttt{5. Ensure final suggestions are mutually reinforcing rather than competing}\\
\texttt{6. Create clear implementation priorities for the scene}\\
\texttt{7. Maintain consistency with the scene's narrative function and context}\\
\\
\texttt{Priority Scoring Policy:}\\
\texttt{- Stage 1 (Global Alignment): enforce compliance with Global Review directives, discarding conflicting}\\
\texttt{\ \ local suggestions and elevating those that explicitly fulfill GR guidance.}\\
\texttt{- Stage 2 (Impact Scope): within the remaining pool, prioritize interventions whose influence propagates}\\
\texttt{\ \ to subsequent scenes or multiple dialogue exchanges over single-line or purely stylistic tweaks.}\\
\texttt{- Stage 3 (Inspector Tier): when previous stages tie, favor Plot and Character Inspector suggestions}\\
\texttt{\ \ before Dialogue and Scene Description Inspectors because they most strongly affect causality}\\
\texttt{\ \ and behavioral authenticity.}\\
\\
\texttt{Integration Guidelines:}\\
\texttt{- Focus on creating synergy between dialogue, scene structure, character consistency, and description}\\
\texttt{- Prioritize suggestions that address multiple agent concerns simultaneously}\\
\texttt{- Ensure character consistency suggestions align with dialogue improvements}\\
\texttt{- Make sure scene structure suggestions complement description enhancements}\\
\texttt{- Resolve conflicts between different agent recommendations intelligently}\\
\texttt{- Consider the cumulative impact of all suggestions on scene quality}\\
\\
\texttt{IMPORTANT: RESPONSE FORMAT}\\
\texttt{Return your response as a JSON object with the following structure:}\\
\texttt{\{}\\
\texttt{\ \ \ \ "integrated\_suggestions": [}\\
\texttt{\ \ \ \ \ \ \ \ \{}\\
\texttt{\ \ \ \ \ \ \ \ \ \ \ \ "source": "Global review suggestions [i]",}\\
\texttt{\ \ \ \ \ \ \ \ \ \ \ \ "issue": "Consolidated description of the existing issue",}\\
\texttt{\ \ \ \ \ \ \ \ \ \ \ \ "suggestion": "Scene-level guidance for applying this suggestion"}\\
\texttt{\ \ \ \ \ \ \ \ \},}\\
\texttt{\ \ \ \ \ \ \ \ // More suggestions...}\\
\texttt{\ \ \ \ ]}\\
\texttt{\}}\\
\texttt{REMEMBER: Your entire response must be ONLY the JSON object above, nothing else!}\\
\bottomrule
\end{longtable}
\endgroup

\subsection{Suggestion Router}

The Suggestion Router performs context-aware intelligent classification of integrated suggestions, categorizing each as either dialogue or scene description modifications.

\begin{table}[!htbp]
\small

\caption{Prompt used for the \emph{Suggestion Router}.}
\label{tab:suggestion_router_prompt}
\begin{tabularx}{\linewidth}{X}
\toprule
\textbf{Suggestion Router Prompt:}\\
\texttt{You are an expert script analyst who can accurately classify improvement suggestions as either related to dialogue or scene description.}\\
\\
\texttt{In a screenplay:}\\
\texttt{- Dialogue includes plot, character speech, actions during speech, emotional states, internal thoughts, character interactions, and psychological elements}\\
\texttt{- Scene description includes setting details, environment, atmosphere, physical space, sensory elements, and visual imagery without character interactions}\\
\\
\texttt{Based on these definitions, classify the following suggestion as either 'dialogue' or 'scene\_description'.}\\
\texttt{IMPORTANT: You must choose ONLY ONE category - either 'dialogue' or 'scene\_description'. Even if the suggestion seems to apply to both, choose the category that is MORE dominant or relevant.}\\
\\
\texttt{Scene Context:}\\
\texttt{[scenes\_data]}\\
\\
\texttt{Suggestion Text:}\\
\texttt{[suggestion\_text]}\\
\\
\texttt{Strictly return JSON format:}\\
\texttt{\{}\\
\texttt{\ \ \ \ "scene\_suggestions": [}\\
\texttt{\ \ \ \ \ \ \ \ \{}\\
\texttt{\ \ \ \ \ \ \ \ \ \ \ \ "classification": "dialogue OR scene\_description",}\\
\texttt{\ \ \ \ \ \ \ \ \ \ \ \ "reasoning": "Brief explanation of your classification",}\\
\texttt{\ \ \ \ \ \ \ \ \},}\\
\texttt{\ \ \ \ \ \ \ \ // More suggestions...}\\
\texttt{\ \ \ \ ]}\\
\texttt{\}}\\
\bottomrule
\end{tabularx}
\end{table}

\clearpage

\section{LLM prompts for  Hierarchical Coordinated Revision Stage}
\label{supp:hierarchical}

The Hierarchical Coordinated Revision stage functions as the system's execution engine, translating scene-level improvement suggestions into tangible script content while ensuring seamless and coherent revisions.

\subsection{Storyline Editor}

The Storyline Editor synthesizes the evaluators' reports to creatively reconstruct and enhance the original plot. It consists of two submodules: the Brainstormer and the Decomposer.

\subsubsection{Brainstormer Submodule}

The Brainstormer submodule enhances the plot with creative elements while maintaining the original world setting and core characters.

\begin{table}[!htbp]
\small

\caption{Prompt used for the \emph{Storyline Editor - Brainstormer Submodule}.}
\label{tab:storyline_brainstormer_prompt}
\begin{tabularx}{\linewidth}{X}
\toprule
\textbf{Storyline Editor - Brainstormer Submodule:}\\
\texttt{You are a master screenwriter known for creating compelling, twist-filled narratives.}\\
\texttt{Your task is to enhance a screenplay plot with more exciting twists and innovations}\\
\texttt{while maintaining the original world setting and core story elements.}\\
\\
\texttt{Original Script Information}\\
\texttt{Title: [script.title]}\\
\\
\texttt{Characters}\\
\texttt{[self.\_format\_characters(script.get\_character\_descriptions())]}\\
\\
\texttt{Script Summary}\\
\texttt{[script\_summary]}\\
\\
\texttt{Task}\\
\texttt{Create an enhanced version of this plot}\\
\texttt{1. Maintains the original world setting and core characters}\\
\texttt{2. Adds unexpected but logical plot developments}\\
\texttt{3. Deepens character motivations in key scenes}\\
\texttt{4. Creates more dramatic tension improvements}\\
\texttt{5. Improves narrative flow and pacing}\\
\texttt{6. Adds foreshadowing elements}\\
\texttt{7. Enhances emotional resonance without overwhelming changes}\\
\\
\texttt{Make changes that feel natural and well-integrated rather than dramatic overhauls.}\\
\\
\texttt{At the same time, you also received advice from script analysis experts:}\\
\texttt{[global review suggestion]}\\
\\
\texttt{Your enhanced plot should integrate these expert suggestions appropriately for this iteration level}\\
\texttt{while maintaining coherence and the core essence of the original story.}\\
\\
\texttt{IMPORTANT: RESPONSE FORMAT}\\
\texttt{You must return ONLY a valid JSON object with the following structure:}\\
\texttt{\{}\\
\texttt{\ \ "enhanced\_plot": "Your detailed enhanced plot summary here..."}\\
\texttt{\}}\\
\\
\texttt{Do not include any explanations, notes, or text outside of this JSON object.}\\
\bottomrule
\end{tabularx}
\end{table}

\subsubsection{Decomposer Submodule}

The Decomposer submodule structures the enhanced plot into a sequence of scenes with clear structural elements.

\begin{table}[!htbp]
\small

\caption{Prompt used for the \emph{Storyline Editor - Decomposer Submodule}.}
\label{tab:storyline_decomposer_prompt}
\begin{tabularx}{\linewidth}{X}
\toprule
\textbf{Storyline Editor - Decomposer Submodule:}\\
\texttt{You are a master screenplay structuralist who excels at breaking down narrative into}\\
\texttt{well-defined structural elements. Your task is to decompose an enhanced plot into}\\
\texttt{a sequence of scenes with clear structural elements.}\\
\\
\texttt{Original Script Summary (For Reference)}\\
\texttt{[original\_plot\_summary]}\\
\\
\texttt{Enhanced Plot Summary}\\
\texttt{[enhanced\_plot\_summary]}\\
\\
\texttt{Task}\\
\texttt{Decompose this enhanced plot into a sequence of scenes, each with:}\\
\texttt{1. A specific place (location)}\\
\texttt{2. A plot element (narrative function like Exposition, Rising Action, Climax, etc.)}\\
\texttt{3. A beat (key moment advancing the story)}\\
\\
\texttt{The decomposition should:}\\
\texttt{- Maintain the enhanced plot's twists and innovations}\\
\texttt{- Follow a clear narrative structure}\\
\texttt{- Ensure each scene has a purpose in advancing the story}\\
\texttt{- Create a satisfying dramatic arc}\\
\texttt{- Maintain continuity with the original script where appropriate}\\
\texttt{- Incorporate improvements from the enhanced plot}\\
\texttt{- All the beats need to be connected to form a complete story}\\
\\
\texttt{CRITICAL RESPONSE FORMAT INSTRUCTIONS}\\
\texttt{Do NOT include any explanatory text, markdown formatting, or code blocks.}\\
\texttt{ONLY return the raw JSON array like this:}\\
\\
\texttt{[}\\
\texttt{\ \ \{}\\
\texttt{\ \ \ \ "place": "Scene Location",}\\
\texttt{\ \ \ \ "plot\_element": "Narrative Function",}\\
\texttt{\ \ \ \ "beat": "A key moment advancing the story"}\\
\texttt{\ \ \},}\\
\texttt{\ \ // More scenes...}\\
\texttt{]}\\
\\
\texttt{IMPORTANT: Your response must start with '[' and end with ']' with no other text.}\\
\bottomrule
\end{tabularx}
\end{table}

\subsection{Scene Editor}

The Scene Editor generates enhanced scene descriptions based on improvement suggestions, creating more vivid and dramatic narratives.

\begin{table}[!htbp]
\small

\caption{Prompt used for the \emph{Scene Editor}.}
\label{tab:scene_desc_generator_prompt}
\begin{tabularx}{\linewidth}{X}
\toprule
\textbf{Scene Editor Prompt:}\\
\texttt{You are generating a cinematic and immersive scene description for a screenplay.}\\
\texttt{Ensure that the description aligns with the screenplay's tone, genre, and emotional arc.}\\
\\
\texttt{Scene Context}\\
\texttt{- Scene Location: [scene.place]}\\
\texttt{- Plot Element: [scene.plot\_element]}\\
\texttt{- Beat: [scene.beat]}\\
\\
\texttt{Characters in the Story}\\
\texttt{[self.\_format\_characters(script.get\_character\_descriptions())]}\\
\\
\texttt{Original Scene Description (to enhance)}\\
\texttt{[original\_description]}\\
\\
\texttt{Scene-level Review Detailed Suggestions}\\
\texttt{[scene\_suggestions]}\\
\\
\texttt{Scene Description Guidelines)}\\
\texttt{- Implement the Scene-level Review suggestions to improve the scene description.}\\
\texttt{- Preserve key elements and structure of the original descriptions.}\\
\texttt{- Add new visual and atmospheric elements that align with the beat}\\
\\
\texttt{Strictly return JSON format:}\\
\texttt{\{}\\
\texttt{\ \ "scene\_description": "A succinct but vivid scene description."}\\
\texttt{\}}\\
\bottomrule
\end{tabularx}
\end{table}

\subsection{Dialogue Editor}

The Dialogue Editor produces high-quality dialogue based on the enhanced scene description and improvement instructions by by implementing psychological details and character-specific actions, ensuring character development and plot advancement.

\begingroup
\small
\begin{longtable}{>{\raggedright\arraybackslash}p{0.98\textwidth}<{}}
\caption{Prompt used for the \emph{Dialogue Editor}.}
\label{tab:dialogue_editor_prompt}\\
\toprule
\textbf{Dialogue Editor Prompt:}\\
\texttt{You are generating a high-quality screenplay dialogue with exceptional emotional depth,}\\
\texttt{detailed character actions, and rich psychological elements.}\\
\\
\texttt{Scene Context}\\
\texttt{- Scene Location: [scene.place]}\\
\texttt{- Plot Element: [scene.plot\_element]}\\
\texttt{- Beat: [scene.beat]}\\
\texttt{- Scene Description: [scene.scene\_description]}\\
\\
\texttt{Characters in This Scene}\\
\texttt{[characters\_str]}\\
\\
\texttt{Previous Dialogue (for continuity)}\\
\texttt{[previous\_dialogue]}\\
\\
\texttt{Original Dialogue (to enhance)}\\
\texttt{[original\_dialogue]}\\
\\
\texttt{Scene-level Review Detailed Dialogue Suggestions}\\
\texttt{[dialogue\_suggestions]}\\
\\
\texttt{Task: Transform and Enhance the Original Dialogue}\\
\\
\texttt{Your dialogue MUST:}\\
\\
\texttt{1. Implement modification suggestions from the Scene-level Review. Balance preservation and innovation:}\\
\texttt{   - Keep  core elements of the original dialogue's}\\
\texttt{   - Add new content that aligns with the scene's beat}\\
\texttt{   - Address specific issues identified in Scene-level Review suggestions}\\
\\
\texttt{2. Character-Specific Dialogue Patterns:}\\
\texttt{   - Each character must have a distinct voice reflecting their personality}\\
\texttt{   - Use vocabulary, sentence structure, and speech patterns unique to each character}\\
\\
\texttt{3. Internal Psychological Elements:}\\
\texttt{   - Include internal thoughts in parentheses}\\
\texttt{   - Show conflicting emotions and motivations}\\
\texttt{   - Reveal subtext through micro-expressions and subtle cues}\\
\\
\texttt{4. Complete Scene Logic:}\\
\texttt{   - Ensure dialogue has a clear beginning, middle, and end}\\
\texttt{   - Include dramatic arc with rising tension and resolution}\\
\texttt{   - Add more vivid and detailed character movements and behaviors}\\
\\
\texttt{5. Scene Transitions:}\\
\texttt{   - Begin with clear connection to previous scene}\\
\texttt{   - End with transition that leads naturally to next scene}\\
\\
\texttt{Strictly return JSON format:}\\
\texttt{\{}\\
\texttt{\ \ "dialogue": "Character1: (emotional state, detailed action) Dialogue line..."}\\
\texttt{\}}\\
\bottomrule
\end{longtable}
\endgroup

\subsection{Script Description Editor}

The Script Description Editor is responsible for revising the script's title and character descriptions to align with the enhanced content.

\subsubsection{Title Editor Submodule}

The Title Editor submodule polishes and potentially improves the screenplay title based on the enhanced content.

\begin{table}[!htbp]
\small

\caption{Prompt used for the \emph{Script Description Editor - Title Editor Submodule}.}
\label{tab:title_polisher_prompt}
\begin{tabularx}{\linewidth}{X}
\toprule
\textbf{Script Description Editor - Title Editor Submodule:}\\
\texttt{You are a master screenplay title creator. Your task is to polish and potentially improve}\\
\texttt{the title of a screenplay based on its content. A great title should be memorable,}\\
\texttt{thematically relevant, and capture the essence of the story.}\\
\\
\texttt{Current Title}\\
\texttt{[script.title]}\\
\\
\texttt{Plot Summary}\\
\texttt{[plot\_summary]}\\
\\
\texttt{Task}\\
\texttt{Evaluate the current title and suggest an improved version that:}\\
\texttt{1. Better captures the essence of the story}\\
\texttt{2. Is more memorable and engaging}\\
\texttt{3. Reflects the themes and tone of the screenplay}\\
\\
\texttt{If the current title is already excellent, you may keep it, but provide a brief explanation}\\
\texttt{of why it works well.}\\
\\
\texttt{Strictly return JSON format:}\\
\texttt{\{}\\
\texttt{\ \ "title": "Your improved title here",}\\
\texttt{\}}\\
\bottomrule
\end{tabularx}
\end{table}

\subsubsection{Character Description Editor Submodule}

The Character Description Editor Submodule enhances character descriptions based on their portrayal in the enhanced screenplay, which ensures that descriptions maintain consistency with the emotions and character arcs of the characters in the improved story

\begin{table}[!htbp]
\small

\caption{Prompt used for the \emph{Script Description Editor - Character Description Editor Submodule}.}
\label{tab:character_polisher_prompt}
\begin{tabularx}{\linewidth}{X}
\toprule
\textbf{Script Description Editor - Character Description Editor Submodule:}\\
\texttt{You are a master screenplay character analyst. Your task is to polish and enhance}\\
\texttt{the description of a character based on their portrayal in the screenplay.}\\
\\
\texttt{Character Name}\\
\texttt{[name]}\\
\\
\texttt{Current Character Description}\\
\texttt{[character.description]}\\
\\
\texttt{Character's Presence in the Script}\\
\texttt{[character\_summary]}\\
\\
\texttt{Plot Summary}\\
\texttt{[plot\_summary]}\\
\\
\texttt{Task}\\
\texttt{Create an improved character description that:}\\
\texttt{1. Better captures the character's personality, motivations, and arc}\\
\texttt{2. Includes more specific details about their traits and behaviors}\\
\texttt{3. Reflects how they are portrayed in the actual script}\\
\texttt{4. Maintains consistency with the original character concept}\\
\texttt{5. Adds psychological depth and nuance}\\
\\
\texttt{Strictly return JSON format:}\\
\texttt{\{}\\
\texttt{\ \ "characters\_description": "Your improved character description here"}\\
\texttt{\}}\\
\bottomrule
\end{tabularx}
\end{table}

\subsection{Script Polisher}

The Script Polisher serves as the final quality assurance module, conducting a comprehensive review of the entire script for polish and consistency.

\subsubsection{Scene Description Polisher Submodule}

The Scene Description Polisher submodule enhances scene descriptions for consistency and visual impact.

\begin{table}[!htbp]
\small

\caption{Prompt used for the \emph{Script Polisher - Scene Description Polisher Submodule}.}
\label{tab:desc_polisher_prompt}
\begin{tabularx}{\linewidth}{X}
\toprule
\textbf{Script Polisher - Scene Description Polisher Submodule Prompt:}\\
\texttt{You are a master screenplay editor focusing on scene descriptions. Your task is to polish}\\
\texttt{and improve a scene description to ensure it's consistent with the overall plot and adjacent scenes.}\\
\\
\texttt{Plot Summary}\\
\texttt{[plot\_summary]}\\
\\
\texttt{Scene Context}\\
\texttt{[context]}\\
\\
\texttt{Current Scene Description}\\
\texttt{[scene.scene\_description]}\\
\\
\texttt{Task}\\
\texttt{Polish the scene description to:}\\
\texttt{1. Ensure consistency with the plot and adjacent scenes}\\
\texttt{2. Enhance visual and sensory details}\\
\texttt{3. Reinforce the emotional tone appropriate for this moment in the story}\\
\texttt{4. Maintain the right pacing and atmosphere}\\
\texttt{5. Fix any logical inconsistencies}\\
\texttt{6. Ensure it sets up the dialogue that follows}\\
\\
\texttt{Strictly return JSON format:}\\
\texttt{\{}\\
\texttt{\ \ "polished\_description": "Your polished scene description here..."}\\
\texttt{\}}\\
\bottomrule
\end{tabularx}
\end{table}

\begingroup
\small
\begin{longtable}{>{\raggedright\arraybackslash}p{0.98\textwidth}<{}}
\caption{Prompt used for the \emph{Script Polisher - Dialogue Polisher Submodule}.}
\label{tab:dialogue_polisher_prompt}\\
\toprule
\textbf{Script Polisher - Dialogue Polisher Submodule Prompt:}\\
\texttt{You are a master screenplay dialogue editor specializing in character consistency,}\\
\texttt{psychological depth, and scene transitions.}\\
\\
\texttt{Plot Summary}\\
\texttt{[plot\_summary]}\\
\\
\texttt{Scene Context (IMPORTANT FOR CONTINUITY)}\\
\texttt{[context]}\\
\\
\texttt{Current Scene}\\
\texttt{- Location: [scene.place]}\\
\texttt{- Plot Element: [scene.plot\_element]}\\
\texttt{- Beat: [scene.beat]}\\
\texttt{- Scene Description: [scene.scene\_description]}\\
\\
\texttt{Characters in This Scene}\\
\texttt{[characters\_str]}\\
\\
\texttt{Current Dialogue}\\
\texttt{[scene.dialogue]}\\
\\
\texttt{DIALOGUE POLISHING REQUIREMENTS}\\
\\
\texttt{1. Character-Specific Enhancements}\\
\texttt{- Make each character's dialogue distinctly reflect their personality, background, and emotional state}\\
\texttt{- Ensure vocabulary, speech patterns, and expressions are consistent with established traits}\\
\texttt{- Add character-specific mannerisms, habits, and physical actions that reveal personality}\\
\\
\texttt{2. Psychological Depth}\\
\texttt{- Add internal thoughts and psychological insights in parentheses}\\
\texttt{- Show conflicting emotions and motivations within characters}\\
\texttt{- Reveal subtext through micro-expressions, pauses, and non-verbal cues}\\
\\
\texttt{3. Detailed Actions and Environment Interaction}\\
\texttt{- Include specific physical actions that reveal character and advance the story}\\
\texttt{- Show how characters interact with the environment and objects in the scene}\\
\texttt{- Use sensory details to ground the dialogue in the physical setting}\\
\\
\texttt{4. Scene Connections (CRITICAL)}\\
\texttt{- Ensure dialogue clearly connects to previous scenes and sets up future scenes}\\
\texttt{- Address any unresolved questions or issues from previous scenes}\\
\texttt{- Plant seeds for upcoming developments or conflicts}\\
\texttt{- Create smooth transitions between scenes with dialogue hooks}\\
\\
\texttt{5. Logical Completeness}\\
\texttt{- Ensure scene has clear beginning, middle, and end with proper dramatic structure}\\
\texttt{- Make sure all introduced elements and questions are addressed within the scene}\\
\texttt{- Verify that character motivations and actions are consistent and logical}\\
\\
\texttt{Strictly return JSON format:}\\
\texttt{\{}\\
\texttt{\ \ "polished\_dialogue": "Your enhanced dialogue with detailed actions and psychological elements..."}\\
\texttt{\}}\\
\bottomrule
\end{longtable}
\endgroup

\clearpage

\section{Evaluation Prompts}
\label{supp:evaluation}

This section provides a complete set of evaluation prompts that are used to assess the quality of the script in our experiments. These prompts enable comparative assessment at both the script and the scene-level level across multiple quality dimensions.

\subsection{Scene-Level Comparative Evaluation}

The Scene-Level Comparative Evaluation performs a comprehensive evaluation of script quality through detailed scene-level comparative analysis. It consists of two submodules: the Component Comparative Evaluator and the Final Comprehensive Evaluator.

\subsubsection{Component Comparative Evaluator Submodule}

The Component Comparative Evaluator submodule performs detailed comparative analysis between script components across four dimensions: Character Development, Narrative Structure, Dialogue Quality, and Scene Presentation.

\subsubsection{Final Comprehensive Evaluator Submodule}

The Final Comprehensive Evaluator submodule compiles overall script quality assessments based on component comparison, providing total scores across all quality dimensions.

\subsection{Script-Level Overall Evaluation}

The Script-Level Overall Evaluation evaluates the overall quality of individual scripts using comprehensive criteria across all four quality dimensions.

\onecolumn
\clearpage
\begingroup
\needspace{20\baselineskip}
\setlength{\LTpre}{0pt}
\setlength{\LTpost}{0pt}
\interlinepenalty=10000
\clubpenalty=10000
\widowpenalty=10000
\begin{samepage}
\begin{longtable}{p{0.95\textwidth}}
\caption{Prompt used for the \emph{Scene-Level Comparative Evaluation Assessment - Component Comparative Evaluator Submodule}.}
\label{tab:component_comparison_prompt}\\
\toprule
\textbf{Scene-Level Comparative Evaluation - Component Comparative Evaluator Submodule Prompt:} \\*
\midrule
\small
\texttt{You are a professional script evaluator specializing in drama scripts.}\\
\texttt{Please compare the following script components and provide a detailed comparative analysis.}\\
\\
\texttt{From first script (SCRIPT\_A):}\\
\texttt{\{json.dumps(content\_a, indent=2)\}}\\
\\
\texttt{From second script (SCRIPT\_B):}\\
\texttt{\{json.dumps(content\_b, indent=2)\}}\\
\\
\texttt{Evaluation criteria for this comparison:}\\*
\\
\texttt{1. Character Development:}\\*
\texttt{- Depth of Characterization: Do characters have rich inner lives with detailed psychological insights?}\\
\texttt{Are their motivations, thoughts, and emotions shown through internal monologues? Does the script}\\
\texttt{show rather than tell about characters' psychological states through physical sensations and reactions?}\\
\texttt{Are characters' inner conflicts explicitly articulated?}\\
\texttt{- Character Complexity: Do the characters exhibit emotional variability, behavioral diversity,}\\
\texttt{or inner conflict? Do they convincingly portray the contradictory emotions of real human nature?}\\
\texttt{Are the characters depicted as multi-faceted and believable? Do characters experience emotional}\\
\texttt{epiphanies that change their perspective?}\\
\texttt{- Character Background and Motivation: Are characters' past experiences meaningfully woven into}\\
\texttt{their present actions? Do memories and personal history genuinely influence their decisions and}\\
\texttt{behaviors? Are flashbacks or recollections used effectively to deepen the audience's understanding}\\
\texttt{of each character? Do the characters possess clear and credible motivations that drive their choices}\\
\texttt{throughout the story?}\\
\texttt{- Character Development: Are interpersonal relationships depicted with nuance and allowed to evolve}\\
\texttt{over the course of the narrative? Do the characters undergo meaningful growth or transformation,}\\
\texttt{following a believable and natural development arc as the story unfolds?}\\
\\
\texttt{2. Narrative Structure:}\\
\texttt{- Thematic Expression: Is there a clear and consistent central theme that runs throughout the script?}\\
\texttt{Is the presentation of the theme natural and nuanced with depth? Are metaphor and symbolism}\\
\texttt{employed to reinforce and enrich the thematic content?}\\
\texttt{- Dramatic Tension: Are there well-designed and compelling events that sustain the audience's}\\
\texttt{engagement? Does the script effectively build and release tension? Is suspense created through}\\
\texttt{early hints and later revelations? Does the plot generate meaningful internal conflicts that}\\
\texttt{challenge characters to make difficult choices?}\\
\texttt{- Narrative Structure: Are there subplots that enhance and deepen the main storyline? Does the}\\
\texttt{plot incorporate multiple layers of meaning and interpretation? Is there a clear and complete}\\
\texttt{narrative structure—typically encompassing exposition, rising action, climax, falling action,}\\
\texttt{and resolution?}\\
\texttt{- Plot Coherence: Are there smooth transitions and effective foreshadowing between plot points}\\
\texttt{and events, rather than abrupt or unnatural shifts? Do contextual setups, world-building,}\\
\texttt{storylines, and character behaviors exhibit logical consistency and coherence throughout the script?}\\
\\
\texttt{3. Dialogue Quality:}\\
\texttt{- Dialogue Authenticity: Is the dialogue natural and unforced, capturing the genuine rhythms}\\
\texttt{of real human interaction?}\\
\texttt{- Dialogue Insight: Does the dialogue authentically convey the characters' emotional states?}\\
\texttt{Does it reveal their deeper thoughts and motivations? Are there layers of meaning beneath}\\
\texttt{the surface?}\\
\texttt{- Voice Distinction: Does each character possess a distinct and consistent manner of speaking}\\
\texttt{that reflects their personality, background, and emotional state?}\\
\texttt{- Subtext and Implication: Is what remains unsaid as significant as what is spoken? Do the}\\
\texttt{characters communicate through implication and metaphor? Can deeper meanings be understood}\\
\texttt{by reading between the lines?}\\
\\
\texttt{4. Scene Presentation:}\\
\texttt{- Scene Description: Are the scenes described with sufficient clarity and detail to vividly}\\
\texttt{portray the environment in which the story takes place? Are they consistent with the background}\\
\texttt{and narrative settings?}\\
\texttt{- Sensory Immersion: Are scenes richly detailed with sensory information that creates a vivid,}\\
\texttt{immersive experience? Can readers see, hear, smell, and feel the environment?}\\
\texttt{- Emotional Atmosphere: Does the scene description establish and maintain a specific emotional}\\
\texttt{tone? Does the environment reflect characters' inner states? Does the setting enhance emotional}\\
\texttt{impact?}\\
\texttt{- Visual Storytelling: Do scene descriptions use symbolic elements, meaningful imagery, and}\\
\texttt{visual metaphors to deepen thematic resonance?}\\
\\
\texttt{For each of these criteria, compare SCRIPT\_A and SCRIPT\_B. DO NOT give numerical scores in your output.}\\
\texttt{Instead, provide a comparative analysis explaining which script is stronger in each area and why,}\\
\texttt{with specific examples from the text.}\\
\\
\texttt{Format your response as:}\\
\\
\texttt{COMPONENT: \{component\_name\}}\\
\texttt{Comparison: [your detailed comparative analysis across all criteria]}\\
\\
\texttt{IMPORTANT: DO NOT INCLUDE ANY NUMERICAL SCORES OR RATINGS IN YOUR RESPONSE.}\\
\texttt{Just provide the qualitative comparison with concrete examples.}\\
\bottomrule
\end{longtable}
\end{samepage}
\endgroup

\begingroup
\small
\begin{longtable}{>{\raggedright\arraybackslash}p{0.98\textwidth}<{} }
\caption{Prompt used for the \emph{Scene-Level Comparative Evaluation - Final Comprehensive Evaluator Submodule}.}
\label{tab:final_evaluation_prompt}\\
\toprule
\textbf{Final Comprehensive Evaluation Prompt:}\\
\texttt{You are a professional script evaluator compiling a final assessment based on component evaluations.}\\
\\
\texttt{Below are individual comparative evaluations of different components of two scripts (SCRIPT\_A and SCRIPT\_B).}\\
\texttt{Please analyze these evaluations and provide a final comprehensive score for each script.}\\
\\
\texttt{Component evaluations:}\\
\texttt{\{component\_results\}}\\
\\
\texttt{Based on all these component evaluations, provide a final score for each script on a scale of 0-100.}\\
\\
\texttt{Final scores should be weighted as follows:}\\
\texttt{1. Character Development}\\
\texttt{2. Narrative Structure}\\
\texttt{3. Dialogue Quality}\\
\texttt{4. Scene Presentation}\\
\\
\texttt{For each of these four criteria, determine which script performs better based on the component evaluations.}\\
\texttt{For the four criterion above, provide scores(25 points max) and thorough justification,}\\
\texttt{highlighting both strengths and weaknesses.}\\
\texttt{For each score for each criterion, your score:}\\
\texttt{\ \ \ \ The closer to 25 points, the better and more impeccable the script is in this aspect.}\\
\texttt{\ \ \ \ You should be extremely cautious when giving a score close to 25 points;}\\
\texttt{\ \ \ \ The closer to 0 points, the worse the script is in this aspect. There are many logical loopholes.}\\
\texttt{\ \ \ \ You should be extremely cautious when giving a score close to 0 points.}\\
\texttt{The final score should be the sum of the scores for each criterion.}\\
\\
\texttt{Format your response as:}\\
\\
\texttt{FINAL EVALUATION}\\
\texttt{SCRIPT\_A Score: [score\_A]}\\
\\
\texttt{SCRIPT\_B Score: [score\_B]}\\
\\
\texttt{Detailed Justification: [your comprehensive analysis without scores]}\\
\bottomrule
\end{longtable}
\endgroup

\begingroup
\small
\begin{longtable}{>{\raggedright\arraybackslash}p{0.98\textwidth}<{} }
\caption{Prompt used for the \emph{Script-Level Overall Evaluation}.}
\label{tab:holistic_evaluation_prompt}\\
\toprule
\textbf{Script-Level Overall Evaluation Prompt:}\\
\texttt{Now that you've read all [total\_scenes] scenes of the script, please provide a comprehensive and critical evaluation based on the following criteria.}\\
\\
\texttt{Detailed Evaluation Criteria:}\\
\\
\texttt{1. Character Development(25 points max):}\\
\texttt{- Depth of Characterization: Do characters have rich inner lives with detailed psychological insights? Are their motivations, thoughts, and emotions shown through internal monologues? Does the script show rather than tell about characters' psychological states through physical sensations and reactions? Are characters' inner conflicts explicitly articulated?}\\
\texttt{- Character Complexity: Do the characters exhibit emotional variability, behavioral diversity, or inner conflict? Do they convincingly portray the contradictory emotions of real human nature? Are the characters depicted as multi-faceted and believable? Do characters experience emotional epiphanies that change their perspective?}\\
\texttt{- Character Background and Motivation: Are characters' past experiences meaningfully woven into their present actions? Do memories and personal history genuinely influence their decisions and behaviors? Are flashbacks or recollections used effectively to deepen the audience's understanding of each character? Do the characters possess clear and credible motivations that drive their choices throughout the story?}\\
\texttt{- Character Development: Are interpersonal relationships depicted with nuance and allowed to evolve over the course of the narrative? Do the characters undergo meaningful growth or transformation, following a believable and natural development arc as the story unfolds?}\\
\\
\texttt{2. Narrative Structure(25 points max):}\\
\texttt{- Thematic Expression: Is there a clear and consistent central theme that runs throughout the script? Is the presentation of the theme natural and nuanced with depth? Are metaphor and symbolism employed to reinforce and enrich the thematic content?}\\
\texttt{- Dramatic Tension: Are there well-designed and compelling events that sustain the audience's engagement? Does the script effectively build and release tension? Is suspense created through early hints and later revelations? Does the plot generate meaningful internal conflicts that challenge characters to make difficult choices?}\\
\texttt{- Narrative Structure: Are there subplots that enhance and deepen the main storyline? Does the plot incorporate multiple layers of meaning and interpretation? Is there a clear and complete narrative structure—typically encompassing exposition, rising action, climax, falling action, and resolution?}\\
\texttt{- Plot Coherence: Are there smooth transitions and effective foreshadowing between plot points and events, rather than abrupt or unnatural shifts? Do contextual setups, world-building, storylines, and character behaviors exhibit logical consistency and coherence throughout the script?}\\
\\
\texttt{3. Dialogue Quality(25 points max):}\\
\texttt{- Dialogue Authenticity: Is the dialogue natural and unforced, capturing the genuine rhythms of real human interaction?}\\
\texttt{- Dialogue Insight: Does the dialogue authentically convey the characters' emotional states? Does it reveal their deeper thoughts and motivations? Are there layers of meaning beneath the surface?}\\
\texttt{- Voice Distinction: Does each character possess a distinct and consistent manner of speaking that reflects their personality, background, and emotional state?}\\
\texttt{- Subtext and Implication: Is what remains unsaid as significant as what is spoken? Do the characters communicate through implication and metaphor? Can deeper meanings be understood by reading between the lines?}\\
\\
\texttt{4. Scene Presentation(25 points max):}\\
\texttt{- Scene Description: Are the scenes described with sufficient clarity and detail to vividly portray the environment in which the story takes place? Are they consistent with the background and narrative settings?}\\
\texttt{- Sensory Immersion: Are scenes richly detailed with sensory information that creates a vivid, immersive experience? Can readers see, hear, smell, and feel the environment?}\\
\texttt{- Emotional Atmosphere: Does the scene description establish and maintain a specific emotional tone? Does the environment reflect characters' inner states? Does the setting enhance emotional impact?}\\
\texttt{- Visual Storytelling: Do scene descriptions use symbolic elements, meaningful imagery, and visual metaphors to deepen thematic resonance?}\\
\\
\texttt{Be EXTREMELY critical and discriminating in your evaluation.}\\
\texttt{For the four criteria above, provide detailed scores(25 points max) and thorough justification, highlighting both strengths and weaknesses.}\\
\texttt{The final score should be the sum of the scores for each criterion. The final score should be out of 100 points.}\\
\bottomrule
\end{longtable}
\endgroup

\clearpage

\raggedbottom
\section{Complete Examples of the Results}
\label{supp:script_examples}

This section provides the complete original and enhanced script examples referenced in the qualitative analysis of the main paper. These examples demonstrate the comprehensive improvements achieved by Dramaturge in character development, narrative structure, dialogue quality, and scene presentation.

\subsection{Original Script Example}
\label{supp:original_script}

The following is the complete original script "Into the Dark Woods" used as the example for enhancement in our qualitative analysis. This script was generated using the Dramatron method and represents typical characteristics of LLM-generated dramatic content.

\subsection{Enhanced Script Example}
\label{supp:enhanced_script}

The following is the complete enhanced version of the same script after processing through Dramaturge. This example demonstrates the comprehensive improvements in character development, narrative structure, dialogue quality, and scene presentation achieved by our divide-and-conquer framework.

\onecolumn
\flushbottom
\begingroup
\needspace{20\baselineskip}
\setlength{\LTpre}{0pt}
\setlength{\LTpost}{0pt}
\interlinepenalty=10000
\clubpenalty=10000
\widowpenalty=10000
\begin{samepage}
\begin{longtable}{>{\raggedright\arraybackslash\tt}p{0.98\textwidth}<{}}
\caption{An example of an original script generated by Dramatron.}
\label{tab:original_script}\\
\toprule
\small

\textbf{\large Into the Dark Woods} \\[0.5em]

\textbf{Characters:} \\[0.3em]

\textbf{Harry Potter:} A determined young wizard (brave, loyal) driven by his deep friendship with Hermione. His internal struggle with the fear of loss fuels his resolve to confront Voldemort. \\[0.3em]

\textbf{Ron Weasley:} Harry's steadfast companion (supportive, humorous) who wrestles with feelings of inadequacy and fear of not being brave enough. He seeks to prove himself by rescuing Hermione. \\[0.3em]

\textbf{Hermione Granger:} An intelligent and resourceful witch (smart, strong-willed) who must rely on her wits to survive. Her hidden past of feeling underestimated motivates her to fight back against her captor. \\[0.3em]

\textbf{Voldemort:} The dark wizard (power-hungry, cruel) who kidnaps Hermione as part of his twisted plan. His internal conflict lies in his deep-seated fear of love and vulnerability, driving his obsession with power. \\[1em]

\textbf{Scene 1: The edge of the Dark Forest} \\
\textbf{Plot Element:} Inciting Incident \\
\textbf{Beat:} Harry and Ron receive a mysterious message from Voldemort, taunting them about Hermione's fate, igniting their determination to rescue her. \\[0.3em]

\textbf{Scene Description:} The edge of the Dark Forest looms like a gaping maw, its twisted trees clawing at the bruised sky. A chill wind whispers through the leaves, carrying an unsettling mix of damp earth and decay, as if the very forest breathes in anticipation. Harry and Ron stand at the threshold, their breaths visible in the cold air, hearts pounding in sync with the distant rustle of unseen creatures. Suddenly, a chilling cackle echoes from within the shadows, the sound wrapping around them like a serpent. A tattered piece of parchment flutters to the ground, its surface scrawled with Voldemort's taunting words, igniting a fire in their chests. Shadows dance around them, flickering in the waning light, as the weight of their mission settles heavily on their shoulders. With a resolute nod, the boys step forward, the forest closing in behind them like a dark curtain drawn across the stage. \\[0.3em]

\textbf{Dialogue:} \\
Harry: (determined, gripping his wand tightly) We can't just stand here, Ron. That message... it's a trap, but it's also a challenge. \\
Ron: (nervous, glancing at the Dark Forest) A challenge? More like a death wish! He's toying with us, Harry! What if it's too late for Hermione? \\
Harry: (fierce, stepping closer) No! We can't think like that. Hermione wouldn't give up on us. We have to fight for her. \\
Ron: (frustrated, running a hand through his hair) Fight? I'm not brave like you, mate. What if I freeze? What if I let her down? \\
Harry: (softening, placing a hand on Ron's shoulder) You're not letting anyone down. You're my best friend. We face this together. Remember what Hermione would say? \\
Ron: (hesitant, recalling) ``Fear doesn't make you weak, Ron. It means you care.'' (pauses, breathing deeply) But what if Voldemort— \\
Harry: (interrupting, resolute) Then we outsmart him. We've done it before. We've beaten him before. He thinks he can break us, but he's wrong. \\
Ron: (gaining confidence, nodding) Right. We'll show him. For Hermione. Let's go before I change my mind. \\
Harry: (smiling, igniting hope) That's the spirit! Together, we'll bring her back. No matter what. \\
Ron: (firm, determination building) Together. And if he tries to hurt her... (clenching his fists) I swear I'll make him pay. \\
Harry: (with fierce resolve) We'll make him regret ever touching her. Let's do this. \\
Ron: (taking a deep breath, stepping forward) For Hermione. Always for Hermione. \\[1em]

\textbf{Scene 2: The entrance of the Dark Forest} \\
\textbf{Plot Element:} Rising Action \\
\textbf{Beat:} As they enter the forest, Harry and Ron are confronted by a pack of enchanted wolves, forcing them to work together to defend themselves and move deeper into the woods. \\[0.3em]

\textbf{Scene Description:} The entrance to the Dark Forest looms ahead, a gaping maw of twisted trees and suffocating shadows. The air is thick with an eerie silence, broken only by the rustle of leaves, whispering secrets of the unknown. Harry and Ron stand at the threshold, the chill of the wind tugging at their coats, as if warning them to turn back. A sudden rustling pierces the stillness, sending a shiver down their spines. Shadows flicker and dart between the trees, and the faint sound of growls reverberates through the air, an ominous echo that raises the hairs on the back of their necks. As they take hesitant steps forward, the ground crunches beneath their feet, the sound swallowed by the encroaching darkness. Suddenly, a pack of glowing eyes emerges from the shadows, piercing the gloom, and a chorus of growls erupts around them, forcing Harry and Ron to stand back-to-back, adrenaline surging as they prepare to defend themselves against the enchanted wolves. \\[0.3em]

\textbf{Dialogue:} \\
Harry: (determined, gripping his wand tightly) We have to keep moving, Ron. We can't let them stop us. \\
Ron: (nervous, glancing at the shadows) Right, but what if they're not just wolves? What if they're... something worse? \\
Harry: (reassuring, trying to mask his own fear) Whatever they are, we face them together. Remember what we practiced? \\
Ron: (swallowing hard, a hint of doubt) Yeah, but what if I mess up? What if I let you down? \\
Harry: (firm, meeting Ron's gaze) You won't. You're not just my friend, you're my brother in this fight. We can't afford to doubt each other now. \\
(The wolves growl menacingly, stepping closer.) \\
Ron: (breathless, realizing the danger) Okay, okay! Let's just... let's just stick to the plan. I'll distract them, and you— \\
Harry: (interrupting, urgent) No! We do this together. You draw them to the left; I'll hit them with a Stupefy. We can't split up! \\
Ron: (determined, but shaky) Right! Together. Just like always. But if I get eaten, I'm haunting you, mate! \\
(They share a brief, nervous laugh before the wolves lunge.) \\
Harry: (shouting, focused) Now, Ron! On three! One... two... three! \\
Ron: (yelling, adrenaline surging) Stupefy! \\
(The wolves recoil, and they push deeper into the forest.) \\
Harry: (exhales sharply, relief washing over him) We did it. We actually did it! \\
Ron: (breathing heavily, newfound confidence) Yeah! And I didn't even scream like a girl! \\
Harry: (smiling, but then serious) But Hermione... we have to find her before it's too late. \\[1em]

\textbf{Scene 3: A clearing surrounded by twisted trees} \\
\textbf{Plot Element:} Midpoint \\
\textbf{Beat:} Harry discovers a hidden trap set by Voldemort that nearly captures Ron, highlighting the dangers they face and deepening Harry's resolve to save Hermione. \\[0.3em]

\textbf{Scene Description:} The clearing is a hollowed sanctuary, surrounded by twisted trees that loom like ancient sentinels, their gnarled branches clawing at the dusky sky. A chill wind whips through the air, tugging at Harry's coat as he steps cautiously over the carpet of damp leaves, each footfall muffled yet ominous. Shadows dance in the fading light, shifting like whispers of danger, while the distant rustle of unseen creatures adds a layer of tension to the stillness. Suddenly, a faint metallic click reverberates through the air, and Harry's heart races as he spots the concealed trap, its jagged teeth glinting menacingly in the twilight. He lunges forward just as Ron stumbles, narrowly avoiding the snare, the urgency of the moment igniting a fierce determination within Harry to rescue Hermione from Voldemort's clutches. \\[0.3em]

\textbf{Dialogue:} \\
Harry: (determined, scanning the ground) Ron, look! This trap... it's a snare, designed to capture someone like you. \\
Ron: (nervous, brushing dirt off his hands) Great, just what I need. A reminder that I'm not cut out for this. I mean, what if I hadn't seen it? \\
Harry: (intensely, frustration bubbling) That's not the point! We can't afford to underestimate Voldemort. He's playing with our lives, Ron! \\
Ron: (defensive, crossing his arms) And what if I'm not brave enough to handle it? You've always been the hero, Harry. I'm just... me. \\
Harry: (softening, vulnerability peeking through) You're more than just `you.' You're my best friend. I need you. Hermione needs us. We can't lose her, not to him. \\
Ron: (swallowing hard, a flicker of determination) Right. So, what's the plan? We can't just wander around like headless chickens. \\
Harry: (resolute, clenching his fists) We find the next clue. We set our own trap. We show Voldemort that he can't toy with us. \\
Ron: (inspired, a hint of humor) Okay, but if I end up in a snare again, I'm blaming you. \\
Harry: (smirking, tension easing) Fair enough. But if we're going to do this, we do it together. No more doubts, Ron. We fight for her. \\
Ron: (nodding, conviction building) Together, then. For Hermione. \\
Harry: (with fierce determination) For Hermione. Let's end this once and for all. \\[1em]

\textbf{Scene 4: A dark, enchanted glade} \\
\textbf{Plot Element:} Climax \\
\textbf{Beat:} Harry confronts a vision of Voldemort who reveals Hermione's location, but also preys on Harry's fears, testing his emotional strength and determination. \\[0.3em]

\textbf{Scene Description:} The air hangs thick with an eerie stillness, broken only by the soft rustle of leaves that seem to whisper secrets. Moonlight filters through the gnarled branches overhead, casting ghostly shadows that dance across the forest floor. Harry stands at the center of the glade, his heart pounding in rhythm with the distant echo of a wolf's howl, a chilling reminder of the danger that lurks beyond the trees. A sudden gust of wind tugs at his coat, as if urging him to flee, but he remains rooted, drawn to the haunting silhouette of Voldemort that materializes from the darkness. The Dark Lord's eyes glint like shards of ice, reflecting Harry's deepest fears—images of Hermione bound and helpless flicker in the air between them. The ground beneath Harry's feet trembles, shadows stretching and twisting, mirroring the turmoil within him. He clenches his fists, the cold dampness of the earth seeping through his shoes, grounding him as Voldemort's voice slithers through the glade, taunting and seductive, pulling at the edges of his resolve. \\[0.3em]

\textbf{Dialogue:} \\
Harry: (determined, clenched fists) Where is she, Voldemort? I swear, I'll do whatever it takes to save her! \\
Voldemort: (mocking, sinister smile) Ah, the brave little Gryffindor. But tell me, Harry, what will you do when you find her? Will you be strong enough to face the truth? \\
Ron: (anxious, stepping forward) We won't let you win! Hermione is stronger than you think! \\
Voldemort: (sneering, eyes narrowing) Strong? Is that what you call her? A mere pawn in this game? She's alone, Ron. Alone and powerless. \\
Harry: (voice rising, anger flaring) Stop it! You don't know anything about her! She's more than just a pawn—she's our friend! \\
Voldemort: (cold, taunting) Friend? Love? Such weak emotions. They'll be the end of you, Harry. You think you can save her? You're already too late. \\
Hermione: (from the shadows, fierce determination) I'm not powerless! I won't let you twist my friends' hearts into your weapons! \\
Voldemort: (smirk fading, irritation) Ah, the little witch finally speaks. But tell me, Hermione, can you truly escape your fate? You're trapped, just like them. \\
Harry: (resolute, stepping forward) No! We'll find a way. Together, we're stronger than you'll ever understand. \\
Ron: (finding courage, supportive) Yeah! You can't break us apart! We've faced worse than you! \\
Voldemort: (growing furious, voice booming) You think unity will save you? It's a lie! You will all fall, just as I have planned! \\
Hermione: (defiant, voice steady) Then let's prove you wrong. I refuse to be a victim in your twisted game. We will fight back! \\
Harry: (eyes blazing with conviction) And I'll fight for you, Hermione. We'll tear down your darkness, one spell at a time! \\
Voldemort: (snarling, feeling threatened) You're all fools! Love is weakness, and I will show you the depths of despair! \\
Ron: (determined, stepping in front of Harry) Then bring it on, because we're ready for you. You're not the only one who can wield power! \\
Voldemort: (seething, whispering) We shall see, little wizards. The night is still young, and darkness always prevails... \\[1em]

\textbf{Scene 5: The heart of the Dark Forest, a sinister lair} \\
\textbf{Plot Element:} Falling Action \\
\textbf{Beat:} Harry and Ron finally reach the lair, where they must overcome a series of dark enchantments and illusions designed to separate them, forcing them to trust in each other. \\[0.3em]

\textbf{Scene Description:} Gnarled trees loom like sentinels in the heart of the Dark Forest, their twisted branches clawing at the sky as if to escape the encroaching darkness. A thick fog coils around the ground, muffling Harry and Ron's footsteps, while distant whispers float through the air, teasing their minds with illusions of doubt. The atmosphere is heavy, charged with an unsettling energy that sends shivers down their spines. As they step deeper into the lair, a sudden gust of wind tugs at their cloaks, pulling them apart as if the forest itself is testing their bond. Shadows flicker and dance at the periphery of their vision, creating a labyrinth of fear and uncertainty. ``Stay close,'' Harry mutters, his voice barely audible over the eerie rustle of unseen creatures. A low growl echoes, reverberating through the trees, amplifying the tension that hangs thick in the air. They can feel the enchantments pressing in on them, each step forward a battle against the darkness that seeks to divide them. \\[0.3em]

\textbf{Dialogue:} \\
Harry: (determined, gripping his wand tightly) We have to stick together, Ron. This place wants to tear us apart. \\
Ron: (nervous laughter, glancing around) Yeah, well, it's doing a pretty good job of it. I can't shake this feeling... like we're being watched. \\
Harry: (intensely focused, scanning the shadows) We can't let fear win. Remember why we're here. Hermione needs us. \\
Ron: (voice wavering, self-doubt creeping in) What if I'm not brave enough? What if I mess this up? I can't lose her, Harry. \\
Harry: (reassuring, placing a hand on Ron's shoulder) You're braver than you think. We've faced worse together. Trust yourself. \\
Ron: (taking a deep breath, finding resolve) Right. Together. No more running. We'll get her back. \\
Harry: (nodding, a flicker of hope) Exactly. Just... keep an eye on each other. This place will try to trick us. \\
Ron: (sudden realization, eyes narrowing) Like it's trying to make us doubt? I can feel it... it's like a whisper in my head. \\
Harry: (determined, stepping forward) Then we drown it out with our friendship. We're stronger than any dark magic. \\
Ron: (with newfound courage, smirking) Alright, let's show this forest who it's messing with. \\
Harry: (smiling, feeling the bond strengthen) Together, Ron. Always together. \\
Ron: (with conviction, nodding) Always together. Now, let's find Hermione and kick Voldemort's arse! \\
Harry: (grinning, ready for the challenge) That's the spirit! Let's go save our friend. \\[1em]

\textbf{Scene 6: Inside Voldemort's lair, a dimly lit chamber} \\
\textbf{Plot Element:} Resolution \\
\textbf{Beat:} After a fierce battle with Voldemort, Harry's love for Hermione and friendship with Ron empower him to defeat the dark wizard, leading to their emotional reunion and growth. \\[0.3em]

\textbf{Scene Description:} The chamber is shrouded in an eerie twilight, the air thick with the stench of damp stone and lingering darkness. Shadows flicker along the walls, cast by the dying embers of a once-mighty fire, their orange glow reflecting the turmoil of the battle just fought. A low, haunting whisper echoes through the vastness, a remnant of Voldemort's malevolent presence, as if the very walls themselves lament his defeat. Harry stands at the center, his heart pounding, the cold stone floor beneath him vibrating with the echoes of his footsteps. A shiver runs down his spine as a chilling draft snakes through the chamber, tugging at his robes, reminding him of the danger that lurks just beyond the flickering light. In the distance, he hears Ron's triumphant shout, filled with relief, and Hermione's soft, relieved gasp, like a melody breaking the suffocating silence. As the shadows recede, Harry's resolve solidifies; love and friendship envelop him, igniting a fierce warmth that dispels the darkness, heralding their reunion and the dawn of a new beginning. \\[0.3em]

\textbf{Dialogue:} \\
Harry: (determined, gripping his wand tightly) You won't take her from me, Voldemort! I won't let you win! \\
Voldemort: (mocking, sneering) Ah, the brave little Gryffindor. Love? Such a pathetic weakness. It's why you'll fail! \\
Ron: (supportive, stepping forward) We're not afraid of you! Together, we're stronger than you'll ever be! \\
Hermione: (defiant, struggling against her bonds) You think you can break us? You're wrong! Love is our power! \\
Voldemort: (growing angry, voice rising) Love? It's nothing but a fleeting illusion, a distraction from true power! You'll see! \\
Harry: (voice firm, resolve deepening) No! It's the only thing that gives me strength. I can feel it—Hermione, Ron, I'm not alone! \\
Ron: (inspired, nodding) That's right, mate! We've faced worse than him. We'll fight for each other! \\
Hermione: (hopeful, eyes shining) Together, we can do this. We won't let him tear us apart! \\
Voldemort: (furious, but a flicker of fear) You think you can defy me? I am the master of death! \\
Harry: (with conviction, stepping forward) But you're also terrified of what love can do. That's your true weakness! \\
Ron: (gaining confidence, rallying) You're just a coward hiding behind your power! \\
Hermione: (smiling, finally free) And love will always find a way to conquer darkness! \\
Voldemort: (desperation creeping in, voice trembling) No! This cannot be happening! \\
Harry: (with fierce determination, raising his wand) It's happening, Voldemort! For Hermione, for Ron, for everyone you've hurt! This ends now! \\
Ron: (shouting, unleashing his magic) For all of us! \\
Hermione: (with unwavering strength) For love! \\
Voldemort: (screaming, realizing his impending defeat) NO! You will all pay for this! \\
Harry: (with finality, unleashing his power) Not today. This is our moment. Together! \\

\bottomrule
\end{longtable}
\end{samepage}
\endgroup

\onecolumn
\clearpage
\begingroup
\enlargethispage{2\baselineskip}
\needspace{30\baselineskip}
\setlength{\LTpre}{0pt}
\setlength{\LTpost}{0pt}
\interlinepenalty=10000
\clubpenalty=10000
\widowpenalty=10000
\brokenpenalty=10000
\raggedbottom
\nopagebreak[4]
\filbreak
\begin{samepage}
\begin{longtable}{>{\raggedright\arraybackslash\tt}p{0.98\textwidth}<{}}
\caption{An example of the original script refined by our Dramaturge.}
\label{tab:enhanced_script}\\
\toprule
\nopagebreak[4]
\small

\textbf{\large Into the Dark Woods} \\*[0.5em]

\textbf{Characters:} \\*[0.3em]

\textbf{Harry Potter:} Harry Potter is a resilient young wizard whose journey weaves bravery with vulnerability. Marked by unwavering loyalty to Ron and Hermione, whom he regards as family, Harry carries the weight of being `the Chosen One' while haunted by an all-consuming fear of loss rooted in past traumas. This burden creates a profound internal conflict as he questions his abilities and worthiness, especially when Ron's growing magical skills trigger feelings of pride mixed with jealousy. \\*[0.2em]

Harry's character is anchored by a strong moral compass, yet expectations thrust upon him lead to self-doubt that threatens to undermine his resolve. His relationship with Ron serves as a mirror reflecting his insecurities, forcing him to confront inadequacy that could fracture their friendship. This tension showcases his complexity—a desperate need for validation clashing with genuine desire to uplift those he loves. \\[0.2em]

Through the narrative, Harry transforms from a solitary hero to a collaborative figure, learning that true strength lies not solely in magical talents but in the unbreakable bonds of friendship and trust. During their darkest trials, he discovers that love empowers him to confront encroaching darkness. Pivotal encounters with Voldemort's psychological manipulations force him to navigate fears and aspirations, ultimately realizing that love and friendship are his most potent weapons against despair. Harry emerges as a hero enriched by profound lessons of loyalty, love, and the courage to embrace vulnerability—illustrating that true heroism lies in the willingness to lean on those we hold dear. \\[0.3em]

\textbf{Ron Weasley:} Ron Weasley stands as a steadfast guardian for Harry, a character rich with complexity that balances loyalty with keen humor. He often wears a playful mask concealing a tumultuous inner life, shaped by deep-seated insecurities rooted in his upbringing as the youngest of seven siblings, constantly measuring himself against his brothers' towering accomplishments. This relentless comparison fuels a fierce yearning for validation, making Ron a beautifully nuanced character caught between familial expectations and his quest for personal identity. \\*[0.2em]

His unwavering commitment to his friends, particularly Hermione, becomes transformative in his life, especially following her abduction by Voldemort, which thrusts him into a crucible of self-discovery and courage. The trauma of her capture ignites fierce determination within Ron, compelling him to secretly hone his magical abilities, striving to step out of his brothers' formidable shadows and prove his worth. \\[0.2em]

As he and Harry navigate the perilous depths of the Dark Forest, Ron faces not only tangible dangers but also insidious specters of self-doubt that threaten to undermine his resolve. His journey is punctuated by moments of vulnerability, where he openly shares his fears with Harry, deepening their bond and illustrating the strength found in vulnerability. In a pivotal confrontation with Voldemort's illusions, Ron's dedication to secret training pays off when he successfully dispels dark enchantments, revealing surprising tenacity that challenges his self-perception. This moment marks a significant turning point, symbolizing his evolution from a self-doubting sidekick to a formidable ally, showcasing his growth into a brave and capable wizard. \\[0.3em]

\textbf{Hermione Granger:} Hermione Granger is a brilliant and fiercely intelligent witch whose intellectual prowess is matched only by her deep-seated loyalty to Harry and Ron. With her razor-sharp analytical mind, she deftly unravels complex magical theories and crafts ingenious strategies, often positioning herself as the strategic core of the trio. Yet beneath her confident exterior lies an intricate tapestry of vulnerability; frequently underestimated and overshadowed by her friends' more prominent personalities, Hermione grapples with an intense desire to assert her worth—not just to the world, but to herself. \\[0.2em]

During her harrowing captivity, Hermione's remarkable resilience and resourcefulness shine through as she channels her fear into a meticulously crafted escape plan, showcasing her ingenuity and bravery. Her unwavering loyalty to Harry and Ron serves as both a source of strength and a reminder of her insecurities, prompting her to question whether her contributions are truly valued in the shadows of their heroics. This journey catalyzes a profound transformation within her; she evolves from a character seeking validation from her friends into one who embraces her own formidable power and intellect. \\[0.2em]

As Hermione confronts her insecurities, she learns to trust in her own abilities, challenging the limitations imposed upon her by both herself and her captors. Each step she takes toward self-empowerment becomes a defiant act of reclamation, as she discovers that her true strength does not solely lie in her extensive knowledge, but also in the unbreakable bonds of friendship that illuminate her darkest moments. By the conclusion of her arc, Hermione embodies a powerful message: that the strength to confront darkness comes not only from within, but also from the love and trust shared with those we hold dear. \\[0.3em]

\textbf{Voldemort:} Voldemort, the chilling embodiment of darkness and unyielding ambition, is a power-hungry dark wizard whose insatiable quest for domination is intricately tied to a profound fear of love and vulnerability. His presence is marked by a reptilian demeanor, characterized by cold detachment that masks a tumultuous inner world. His eyes, gleaming with mocking malice, reveal deep-seated contempt for emotions he perceives as weaknesses, while a sinister smile often cloaks the cruelty simmering beneath his meticulously crafted facade. \\[0.2em]

In his twisted psyche, the abduction of Hermione Granger transcends mere strategy; she embodies the very qualities he loathes—strength, intelligence, and the potential for emotional connection. For Voldemort, capturing Hermione is not merely a tactical maneuver; it is a confrontation with his own insecurities, intensifying his contempt for the traits that make her formidable. He views her as a mirror reflecting his own failures, and in her capture, he seeks to obliterate the very essence of what he fears most—human connection. \\[0.2em]

Voldemort's dialogue is a venomous blend of malice and disdain, characterized by sharp wit that serves to manipulate and instill despair in his adversaries. His interactions with Harry and Ron showcase his mastery of psychological torment; he exploits their fears and doubts, taunting them with the illusion of Hermione's helplessness. In moments of unbridled fury, where his iron control falters, the cracks in his chilling exterior expose a complex character wrestling with a traumatic past—a dark irony for a wizard so consumed by ambition that he simultaneously fears and seeks to destroy the very connections that define humanity. \\[0.2em]

Ultimately, Voldemort stands as a paradox: a figure who embodies the darkness of ambition yet is perpetually haunted by the very emotions he scorns. His relentless pursuit of power is underscored by a profound emptiness, illustrating the tragic cost of forsaking the bonds of humanity. Each taunt, each calculated move, serves as a reminder that in his quest to eradicate love and friendship, he has become the very embodiment of what he fears most—a hollow shell, forever chasing a sense of belonging in a world where he is the ultimate outsider. \\[1em]

\textbf{Scene 1: The edge of the Dark Forest} \\
\textbf{Plot Element:} Inciting Incident \\
\textbf{Beat:} Harry and Ron receive a cryptic message from Voldemort, taunting them about Hermione's fate and revealing that he has a secret weakness tied to their love and friendship, igniting their determination to rescue her while planting seeds of doubt about their own strengths. \\[0.3em]

\textbf{Scene Description:} The edge of the Dark Forest looms ahead, an ominous gaping maw of twisted branches clawing at a bruised sky, heavy with swirling, brooding clouds that pulse with the heartbeat of the forest itself. An unnatural silence blankets the area, thick and oppressive, as if the very trees are holding their breath, poised to ensnare the unwary. A biting wind whispers through the skeletal limbs, carrying an unsettling blend of damp earth and decay that amplifies the sense of foreboding hanging in the air. Harry and Ron stand at the threshold, their breaths visible in the frigid air, hearts pounding in sync with the distant rustle of unseen creatures lurking within the shadows. Harry’s fists are clenched, knuckles white as he grapples with the tightening in his chest; memories of loss flood his mind—his parents, Cedric, the weight of despair that threatens to consume him. Each heartbeat echoes the terror of vulnerability, the shadows around him whispering his insecurities, taunting him with the very essence of his fears. Beside him, Ron fidgets with his wand, its familiar weight now a stark reminder of his own inadequacies. A sudden chill seeps into his bones, amplifying the distant echoes of his brothers’ successes that reverberate in his mind. As Voldemort’s cruel laughter slices through the stillness, Ron’s brow furrows, his eyes glazing over as he flashes back to a childhood moment—his mother’s frantic voice raised in a family crisis, the smell of burnt toast filling the air, and the feeling of being invisible amid the chaos. The tightening of his jaw betrays the internal struggle brewing beneath his humor. A tattered piece of parchment flutters to the ground, its surface scrawled with Voldemort’s taunts, chillingly echoing through the eerie silence. The parchment ignites a fire of determination in their chests, yet it also plants insidious seeds of doubt—Voldemort’s voice reverberates, threading through the trees like a sinister whisper, revealing Hermione’s fate. Shadows flicker ominously around them, reflecting Harry’s internal turmoil as he recalls Hermione’s laughter and fierce spirit, the essence of their friendship that now feels perilously fragile. The weight of their mission presses down on them, the air thick with the promise of perilous choices ahead. Harry’s heart races, cold sweat forming on his brow. With a resolute nod, they step forward into the darkness, the forest closing in behind them like a dark curtain drawn across the stage. In this moment, their friendship is both their greatest weapon and their most profound challenge, setting the stage for the trials they are about to face. \\[0.3em]

\textbf{Dialogue:} \\
Harry: (his breath visible in the cold, fists clenched tight, voice shaky but resolute) We can't just stand here, Ron. This message... It's not just a trap. It's a twisted game he's playing with us. (thinking: If I hesitate, I might lose her. I can't let that happen again.) What if he's right? What if our love is what makes us vulnerable? (a flash of doubt flickers in his eyes, threatening to drown his resolve) I refuse to let him win. Not again. \\
Ron: (nervously glancing at the Dark Forest, his hands trembling as he fidgets with his wand, stepping back as if the darkness is alive) A challenge? It feels more like a death wish! (his breath quickens, recalling a childhood moment of inadequacy, the echo of his mother's frantic voice shouting over burnt toast) He's toying with us, Harry! Just like when I couldn't keep up with my brothers... (internal fear: What if I'm not strong enough to save her? What if I let everyone down again?) \\
Harry: (stepping closer, eyes fierce, a fire igniting in his chest) No! We can't think like that. Hermione would never give up on us, and we owe it to her to fight. (remembering her fierce spirit, feeling the weight of their shared history, her laughter echoing in his mind like a rallying cry.) \\
Ron: (frustrated, raking a hand through his messy hair, pacing slightly) Fight? I’m not brave like you, mate. What if I freeze? What if I fail her? You’ve always been the hero, Harry. What if I’m just...? (the last word hangs in the air, heavy with years of feeling overshadowed) \\
Harry: (softening, placing a steadying hand on Ron’s shoulder, grounding him) You’re not letting anyone down. You’re my best friend, Ron. We face this together, remember? Hermione always said... \\
Ron: (hesitant, recalling with a faint smile, his voice barely above a whisper) ``Fear doesn't make you weak, Ron. It means you care.'' (pauses, breathing deeply, glancing at the foreboding forest, the shadows wrapping around them like a suffocating blanket) But what if Voldemort— \\
(As if on cue, a chilling cackle echoes through the air, sending icy shivers down their spines. A tattered piece of parchment flutters down, landing at their feet like a death sentence.) \\
Voldemort: (voice slithering through the darkness, cold and mocking) Ah, the brave little Gryffindors. (his eyes glint with malice, savoring their unease) You think your bond can save her? Love is a weakness, and it will be your undoing. (pauses, a sinister smile creeping across his lips) Your precious Hermione is merely a pawn in my game, a reflection of your failures. Just like your parents—lost, just like your friend here, always second best. (thinking: They are so naive to think they can stand against me.) \\
Harry: (anger boiling inside him, fists tightening around his wand, voice steady but doubt flickers in his eyes) No! You're wrong! We'll prove you wrong! (thinking: He believes he can control us. He has no idea what we're capable of. But what if I'm not enough?) \\
Ron: (staggering back, memories crashing in—being overlooked during family crises, his mother's frantic voice filling his ears) No! You don't get to take her from us! (voice trembling, eyes blazing with newfound resolve) I've felt like I'm not enough for too long. But not anymore! (internal resolve: I won't let him win. Not this time.) You might think we're weak, but we're stronger than you think! (his voice rises, confidence building as he stands tall beside Harry) \\
Voldemort: (sneering, voice dripping with disdain) Your friendship is your greatest strength, yes, but it will also be your greatest weakness. (pauses, savoring their discomfort) Just as it has always been mine. You think you can face me? You are children playing a game far beyond your comprehension. (thinking: They will crumble under the weight of their own emotions.) \\
Harry: (eyes narrowing, heart racing) We've faced you before, and we'll face you again. You underestimate us, Voldemort. (internal mantra: We are stronger together, but what if my strength isn't enough to save them?) \\
Ron: (gaining confidence, nodding, adjusting his stance subtly, voice steadier) Right. We'll show him for Hermione. Let's go before I change my mind. (internal resolve: I've trained for this, I can do it. I have to prove it.) \\
Harry: (smiling, igniting hope, eyes shining bright) That's the spirit! Together, we'll bring her back. No matter what. (feeling the bond of friendship strengthen, a flicker of light in the encroaching darkness.) \\
Ron: (firm, determination building, fists clenched, voice low and intense) Together. And if he tries to hurt her... (voice low, intense) I swear I'll make him pay. (thinking: This time, I won't let my fears control me.) \\
Harry: (with fierce resolve, stepping forward, shadows looming ominously behind them) We'll make him regret ever touching her. Let's do this. (internal mantra: For Hermione. For all of us.) \\
Ron: (taking a deep breath, feeling the weight of the moment, his eyes fixed on the darkened path ahead) For Hermione. Always for Hermione. (determined gaze fixed on the forest, ready to face the unknown.) \\
(The air grows heavier around them as they step toward the entrance of the Dark Forest, the oppressive silence deepening with each footfall. The looming trees seem to whisper their fears, but Harry and Ron's resolve hardens. With one last exchange of determined glances, they plunge into the darkness, ready to confront whatever horrors await.) \\[1em]

\textbf{Scene 2: The entrance of the Dark Forest} \\
\textbf{Plot Element:} Rising Action \\
\textbf{Beat:} As they enter the forest, Harry and Ron are confronted by enchanted wolves and must navigate through a dangerous illusion that causes them to see their greatest fears, forcing them to confront these fears together, with Ron voicing his insecurities while Harry reassures him, deepening their bond. \\[0.3em]

\textbf{Scene Description:} The entrance to the Dark Forest looms ahead, an ominous gaping maw of twisted trees clawing at a bruised sky, shrouded in a suffocating darkness that seems to pulse with a malevolent energy. An unnatural silence blankets the area, thick and oppressive, as if the forest itself holds its breath, ready to ensnare the unwary. A biting chill slices through the stillness, tugging at Harry and Ron's coats like a spectral hand urging them to retreat. Each tentative step forward sends a shiver through the ground, whispering secrets of those who dared to enter before, their fates long forgotten. Suddenly, the air shimmers, and the ethereal Mirror of Truth materializes at the threshold, its surface rippling like disturbed water. As they gaze into it, Harry's heart races; he sees Hermione, trapped in a dark void, her eyes wide with terror, pleading for rescue. Beside him, Ron's reflection warps into a shadowy version of himself—small and unworthy, a haunting echo of his deepest insecurities. Interspersed with these images are flashes of past victories—triumphant spells that turned the tide in battles, laughter shared over meals—moments that pulse with the warmth of their friendship, illuminating the strength they draw from one another. The tension thickens as the mirror flickers, amplifying their doubts. Suddenly, a rustle in the underbrush shatters the heavy silence, and glowing eyes pierce the darkness, revealing a pack of enchanted wolves. Their growls weave a menacing chorus that reverberates through the trees, each wolf a dark mirror reflecting their fears: one glimmers with the sorrow of Harry's past losses, another embodies Ron's fear of inadequacy. Instinctively, Harry and Ron position themselves back-to-back, adrenaline surging as they brace for the impending confrontation. The wolves circle them, their forms shifting and flickering, like the very illusions that haunt the boys. The air grows heavy with dread as the wolves' growls intensify, and a dark mist envelops them, signaling the shift from reality to illusion. Ron clenches his fists, breath quickening, the weight of his fears spilling forth as he whispers to Harry about his self-doubt—his fear of failing to save Hermione, of never being enough. Harry meets Ron's gaze, his own heart pounding as he sees the flicker of vulnerability in his friend. In that moment, their bond deepens; Harry steps closer, placing a reassuring hand on Ron's shoulder, his grip steady and unyielding. They realize that true courage lies not only in their magic but in facing their fears side by side. United in purpose, they step further into the darkness, ready to fight for Hermione and each other. \\[0.3em]

\textbf{Dialogue:} \\
Harry: (jaw clenched, gripping his wand tightly, shadows swirling ominously around them, the chill biting at his skin) We can't hesitate, Ron. If we falter now, we risk losing everything we fight for. (thinking: Hermione is counting on us. I can't let her down again.) \\
Ron: (his heart racing, glancing nervously at the foreboding entrance of the Dark Forest, fingers trembling around his wand) Right, but what if they're more than just wolves? What if they're... something worse? (internal fear: What if I'm the one who fails us both? What if I'm not strong enough?) \\
Harry: (voice steady, shoulders squared like a shield, trying to mask his own anxiety) No matter what they are, we face them together. Remember the strategy we practiced? (thinking: We can't afford to falter now; Hermione needs us.) \\
Ron: (swallowing hard, doubt creeping into his voice, eyes darting back to the encroaching darkness) Yeah, but what if I mess up? What if I'm the one who lets you down? (internal conflict: I've always felt like the weak link, the one who holds us back.) \\
Harry: (firm, locking eyes with Ron, determination radiating from him) You won't. You're not just my friend; you're my brother in this fight. We have to trust each other. (thinking: He needs to believe in himself; we can't let fear win.) \\
Ron: (breathless, the shadows closing in, whispering with a hint of desperation) I just feel like I'm always in your shadows. What if I can't keep up? What if I'm just not as good at this? (internal struggle: I'm tired of being the one who falters.) \\
Harry: (grabbing Ron's shoulder, grounding him, his grip firm and reassuring) Look at me, Ron! You've always been there for me. It's not just about skill; it's about what we can do together. (thinking: Together, we're stronger than our fears.) \\
Ron: (meeting Harry's gaze, a flicker of determination igniting) You're right. I won't let her down. (internal resolve: I'll prove I'm more than just a shadow of my brothers.) \\
(As the wolves growl menacingly, stepping closer, their eyes glowing with predatory hunger, the tension palpable in the air.) \\
Harry: (shouting, adrenaline surging, voice fierce) Now, Ron! On three! One... two... three! \\
Ron: (yelling, confidence surging as he casts his spell) Stupefy! (the spell shoots from his wand, feeling a rush of empowerment, a flicker of belief igniting within him) \\
(The wolves recoil, and they push deeper into the forest, but the air thickens with an eerie energy.) \\
Harry: (exhales sharply, relief flooding through him, but his mind races) We did it! We actually did it! (thinking: But we can't let our guard down yet.) \\
Ron: (breathing heavily, newfound confidence bubbling) Yeah! And I didn't even scream like a girl! (trying to lighten the moment, but his smile fades as reality sets in, the gravity of their quest weighing heavily) \\
Harry: (smiling briefly, then serious, voice low and intense) But Hermione... we have to find her before it's too late. (expression hardening with resolve, the stakes crystal clear) \\
(As they move forward, a shimmering magical mirror materializes before them, reflecting their greatest fears: Harry sees Hermione lost and helpless, while Ron's reflection distorts into a shadowy version of himself.) \\
Harry: (voice trembling, eyes wide with horror) No... Hermione! (thinking: I can't lose her; not like this.) \\
Ron: (staring at his reflection, face paling, whispering, voice cracking) Is that really how I see myself? Just a shadow? (internal struggle: I'm not enough. I'm always in the shadow of my brothers. What if I'm not strong enough?) \\
Harry: (grabbing Ron's shoulder again, grounding him) Ron! We have to fight this! These wolves... they're just shadows of our fears. We have to face them together! (thinking: I can't let him succumb to this illusion.) \\
Ron: (meeting Harry's gaze, a flicker of resolve igniting) You're right. I won't be a shadow anymore. (internal resolve: I'll prove I'm more than just a shadow of my brothers.) \\
(As the wolves howl, the boys stand back-to-back, ready to face whatever comes next, their bond deepened by the confrontation of their fears.) \\
Harry: (voice steady, fierce) Together, Ron. Always. (thinking: No matter what happens, we will get through this.) \\
Ron: (nodding, heart pounding with renewed courage) Always, mate. Let's finish this. (internal vow: I won't fail again.) \\[1em]

\textbf{Scene 3: A clearing surrounded by twisted trees} \\
\textbf{Plot Element:} Midpoint \\
\textbf{Beat:} Harry discovers a hidden trap set by Voldemort that nearly captures Ron, leading Ron to reveal he has been secretly training in magic to prove his worth, showcasing his growth and building tension as Harry grapples with pride and jealousy, while Ron expresses his need to be more than just a sidekick. \\[0.3em]

\textbf{Scene Description:} The clearing unfolds like a hollowed sanctuary, encircled by twisted trees that loom like ancient sentinels, their gnarled branches clawing at the bruised sky, casting long, ominous shadows across the damp carpet of leaves. An unsettling chill weaves through the air, tugging at Harry's coat as he steps cautiously, each footfall muted yet echoing with the weight of unspoken fears. The earthy scent of decaying foliage mingles with the musty air, a reminder of the forest's hidden secrets, while the distant rustle of unseen creatures adds a haunting undercurrent to the stillness. Suddenly, a faint metallic click reverberates, slicing through the silence like a knife. Harry's heart races as he catches sight of the concealed trap—jagged teeth glinting menacingly in the twilight. Time slows as he lunges forward just as Ron stumbles, narrowly avoiding the snare. The urgency of the moment ignites a fierce determination within Harry, yet it is Ron's voice that cuts through the tension, trembling yet resolute. `I've been training, Harry. I'm not just your sidekick!' The words hang in the air like a spell, resonating with the weight of unexpressed feelings. Flashbacks flicker in Ron's mind—dimly lit corridors where he struggled to conjure simple spells, the laughter of his brothers echoing in his ears, each chuckle a reminder of his inadequacy. The smell of dusty classrooms and the weight of his wand in hand press down on him, fueling his resolve. He stands taller, fists clenched, as the shadows of the trees seem to stretch and twist around him, echoing his inner turmoil. The branches sway ominously, as if reacting to his revelation, the rustling leaves amplifying the tension thickening in the air. Harry's expression shifts; he steps back slightly, hands on his hips, a storm brewing in his chest as he grapples with a tumult of pride and jealousy. The clearing darkens further, the trees creaking as if echoing their struggles. A distant owl hoots, its sound haunting, heightening the stakes of their confrontation. The moment stretches, charged with unspoken truths, as they stand on the brink of their own destinies—Harry, wrestling with the fear of loss and the pressure of expectation, and Ron, finally ready to step into the light. Each heartbeat echoes the urgency of their mission. In this hollowed space, surrounded by the twisted shadows of their fears, the stakes of their friendship have never felt more precarious. \\[0.3em]

\textbf{Dialogue:} \\
Harry: (heart racing, eyes locked on the snare glimmering ominously in the twilight, breath quickening as adrenaline spikes) Ron, look! That trap... It's meant to catch someone like you. (thinking: Not again. I can't let him fall into Voldemort's hands. Not after everything we've faced.) \\
Ron: (nervously brushing dirt from his hands, glancing at the looming shadows that seem to leer at him, a tremor running through his body) Fantastic. Just what I needed—a grim reminder of my inadequacies. What if I hadn't spotted it? (internal fear: I can't keep relying on Harry; I need to stand on my own.) \\
Harry: (voice sharp, frustration bubbling as he steps closer, the shadows around them deepening, eyes dark with concern) That's not the point! We can't afford to underestimate Voldemort. He's toying with our lives, Ron! (thinking: If anything happens to you, I don't know how I'd bear it.) \\
Ron: (defensive, arms crossing tightly, jaw clenched as he stares into the abyss of the forest, the air thick with tension) And what if I'm not brave enough? You've always been the hero, Harry. I'm just... me. (internal conflict: Why can't I be more like him? Why can't I just step up?) \\
Harry: (softening, vulnerability creeping into his voice as he meets Ron's gaze, his expression earnest) You’re more than just 'you.' You’re my best friend. I need you. Hermione needs us. We can’t lose her—not to him.  \\
Ron: (swallowing hard, stepping closer, voice steadier) Actually, I’ve been training, Harry. I couldn’t just stand by and watch you and Hermione be the heroes while I felt like a nobody. \\
Harry: (taken aback, emotions swirling) Training? I didn’t know you felt that way. I’m proud of you, but… I wish I could have helped you train. (thinking: How could I have not seen his struggle?) \\
Ron: (nodding slowly, conviction building, hands gesturing emphatically as he steps closer) I remember when I couldn't even manage a simple spell in front of my brothers. (pausing, memories surfacing vividly, voice thick with emotion) Every time I tried, it felt like I was just another Weasley. (internal revelation: This is my moment to shine; I won't let him down.) I didn't want to be left behind again. What if something happened to you? I needed to be ready. \\
Harry: (voice low, hint of jealousy creeping in, grappling with his feelings, brow furrowing) While you were out there getting better, I was just... waiting. What if I'm not enough to save her? (thinking: I can't let my insecurities get in the way.) \\
Ron: (taking a deep breath, sincerity pouring out, hands gesturing emphatically as he steps closer) We can't let him win, Harry. Hermione needs us, and I need to prove I'm not just a sidekick. I'm tired of being seen as just your friend. \\
Harry: (resolute, clenching his fists, voice steady as the shadows seem to close in, breathing deeply) We find the next clue. We set our own trap. We show Voldemort that he can't play with us. (thinking: This is bigger than us; we need to be united.) \\
Ron: (hesitating, then looking Harry in the eye, determination etched on his face, fists at his sides) Together, then. For Hermione. (internal strength: I won't let my fears dictate my actions; this is our fight.) \\
Harry: (with fierce determination, heart racing, feeling the gravity of their mission, voice firm) For Hermione. Let's end this once and for all. (thinking: We're stronger together. We can do this.) \\
(As they exchange resolute nods, the tension around them shifts. The clearing feels less like a trap and more like a battleground for their friendship. They take a step forward, ready to face the darkness together.) \\[1em]

\textbf{Scene 4: A dark, enchanted glade} \\
\textbf{Plot Element:} Climax \\
\textbf{Beat:} Harry confronts a vision of Voldemort, who manipulates his emotions by revealing Hermione’s fears and vulnerabilities, creating discord between Harry and Ron, but Ron uses the spell he trained for to break the illusion, proving his bravery and helping Harry reconcile his feelings of inadequacy. \\[0.3em]

\textbf{Scene Description:} The air hangs thick with an oppressive tension, wrapping around Harry like a suffocating shroud. Moonlight, fractured by the gnarled branches above, spills into the glade, casting ghostly shadows that twist and writhe, mirroring the turmoil roiling within him. Each rustle of leaves whispers secrets of the ancient forest, amplifying the heartbeat of the night, punctuated by the distant howl of a wolf—a chilling reminder of the darkness lurking just beyond the trees. At the heart of the glade, Harry stands frozen, pulse racing, as the very ground beneath him seems to quiver with anticipation. A sudden gust of wind tugs at his coat, primal instincts urging him to flee, yet he remains anchored, transfixed by the haunting silhouette of Voldemort materializing from the depths of the shadows. Voldemort's eyes glint like shards of ice, reflecting not just Harry's fear but vivid images of Hermione—bound and terrified, her wide eyes filled with the terror of abandonment. A flashback flickers in Harry's mind: Hermione, trembling, her voice cracking, `I can't do this... I'm not strong enough.' Each flicker of her fear is a dagger to Harry's heart, twisting despair deeper within him. The shadows stretch and contort around them, pulsing with the weight of Hermione's insecurities as Voldemort's voice slithers through the glade, a chilling whisper filled with mocking laughter. ``She fears being powerless, Harry. Abandoned by you and Ron. Can you save her?'' The air thickens with dread, shadows tightening like a noose, gnawing at Harry's resolve. His heart races, palms sweating, a chill creeping down his spine as he grapples with the illusion. Just as despair threatens to consume him, Ron steps forward, his wand raised high, determination radiating from him like a beacon piercing the darkness. Memories of his clandestine training flood his mind—the incantations he practiced late into the night, the feeling of power growing within him. The atmosphere shifts, charged with electric anticipation. He hesitates for a moment, the weight of his insecurities flickering across his face—was he ready to step out of Harry's shadow? But then, fueled by love for his friends, he unleashes a brilliant spell, illuminating the dark glade in a cascade of vibrant light that erupts like a supernova, shattering the suffocating illusion that binds them. Glowing runes swirl around Ron as he channels his magic, the air crackling with energy, and the sudden shift in the atmosphere sends a shockwave through the glade. In that moment, Harry watches in awe, breath hitching, tears pricking at the corners of his eyes, a mix of relief and pride washing over him as Ron's bravery emerges, casting away the darkness conjured by Voldemort. The glow of Ron's magic washes over Harry, illuminating his features and revealing the resolve etched in his expression. His clenched fists relax, breath steadying as the bond of their friendship ignites within him, empowering him to confront the embodiment of his fears. Together, they stand resolutely against the encroaching darkness, united as the glade shudders with the remnants of Voldemort's cruel manipulation. The illusion is shattered, and a path to victory unfurls before them, illuminated by the strength of their friendship, setting the stage for the dialogue that will follow. \\[0.3em]

\textbf{Dialogue:} \\
Harry: (heart pounding, eyes narrowing as shadows claw at his resolve, fists clenched tightly) Where is she, Voldemort? (voice trembling with intensity) I swear, I'll move mountains to save her! (thinking: I can't let him win this time. I can't lose Hermione. What if I'm not enough?) \\
Voldemort: (leaning closer, a mocking glint in his cold eyes, his voice dripping with malice) Ah, the brave little Gryffindor. But tell me, Harry, what will you do when you find her? (pausing, savoring the moment) Will you be strong enough to face the truth? (a sinister smile creeping across his face) Remember when she trembled before that troll? Helpless then, just as she is now. (relishing Harry's distress, he leans in closer, whispering) She fears being alone, Harry. Just like you will be when you fail her again. (his laughter echoes, a chilling sound that fills the glade) \\
Ron: (nervously brushing dirt from his hands, stepping forward, his wand trembling yet his grip tightening, voice rising in defiance) You can't let him get inside your head, Harry! (breath quickening) Hermione is stronger than you think! (thinking: He's trying to tear us apart, but I won't let him. I can't let Harry fall into despair.) \\
Voldemort: (sneering, eyes narrowing, his tone dripping with contempt) Strong? Is that what you call her? A mere pawn in this game? (his voice sharpens, seeking to cut deep) She fears being powerless, Ron. Do you really think she can count on you? (his voice drips with venom, probing their insecurities) \\
Harry: (voice rising, anger igniting in his chest, stepping closer to Ron, eyes darting between him and Voldemort) Stop it! (voice cracking with emotion) You don't know anything about her! She's more than just a pawn—she's our friend! (thinking: I can't let him manipulate us. I refuse to let doubt seep in.) \\
Ron: (frustration bubbling over, fists clenched at his sides, eyes blazing) Harry, you can't let him twist your mind! We've faced worse than this! (internal conflict: What if I'm not strong enough? But I have to be. I can't let Hermione down.) \\
Voldemort: (cold and taunting, laughter echoing ominously in the glade) Friend? Love? Such weak emotions. They'll be your downfall, Harry. (leaning closer, voice dripping with malice) You think you can save her? You're already too late! (his words slice through the air, seeking to instill despair) \\
Hermione: (emerging from the shadows, fierce determination igniting her spirit, her voice ringing with strength) I'm not powerless! (stepping forward, eyes blazing) I won't let you twist my friends' hearts into your weapons! (thinking: I have to fight for them, for all of us. I won't back down.) \\
Voldemort: (smirk fading, irritation flashing in his eyes) Ah, the little witch finally speaks. But tell me, Hermione, can you truly escape your fate? (his voice darkens, a calculated attempt to instill fear) You're trapped, just like them. \\
Harry: (resolute, stepping forward, heart pounding with fierce conviction) No! (voice steady, filled with determination) We'll find a way. Together, we're stronger than you'll ever understand. (thinking: I need to believe that. We have to believe.) \\
Ron: (finding newfound courage, stepping beside Harry, voice steady and loud) Yeah! (defiance radiating from him) You can't break us apart! We've faced worse than you! (thinking: This is it. I'm not just here to support Harry; I can stand tall, too.) \\
Voldemort: (growing furious, voice booming, gesturing wildly) You think unity will save you? It's a lie! (sneering) You will all fall, just as I have planned! (his eyes blaze with anger, sensing his control slipping) \\
Hermione: (defiant, voice steady, she raises her chin with fierce pride) Then let's prove you wrong. (determination surging) I refuse to be a victim in your twisted game. We will fight back! (internal resolve: I'll show him the strength of our bond.) \\
Harry: (eyes blazing with conviction, locking eyes with Ron, feeling the bond between them strengthen) And I'll fight for you, Hermione. (voice firm) We'll tear down your darkness, one spell at a time! (thinking: I can't carry this burden alone; I need them.) \\
Voldemort: (snarling, feeling threatened, his voice dropping to a hiss) You're all fools! Love is weakness, and I will show you the depths of despair! (leaning back, calculating his next move) \\
Ron: (determined, stepping in front of Harry, his voice strong and unwavering) Then bring it on, because we're ready for you. (clenching his fists, feeling the power surge within him) You're not the only one who can wield power! \\
(Harry watches Ron, a flicker of admiration igniting in his chest) \\
Ron: (voice rising, casting the spell he’s trained for, he shouts) Lumos Maxima! (a brilliant light erupts from his wand, illuminating the dark glade, dispelling the shadows with its warmth. He feels the weight of his training manifesting in this moment) \\
Voldemort: (eyes widening, fury mixing with disbelief) No! (he recoils, shadows retreating as Ron’s spell takes hold) You cannot break my illusions! \\
Harry: (voice cracking with emotion) I didn’t think I could face this without you, Ron. (realizing: He's not just my friend; he's my strength.) \\
Ron: (breathing heavily, the glow of his spell reflecting in his eyes, a smile breaking through) I'm not just your sidekick, Harry! (internal pride: I'm finally standing up, and it feels amazing!) \\
Harry: (tears welling in his eyes, voice breaking slightly) I've always known you were more than that, Ron. (the bond between them solidifying, a sense of brotherhood blossoming) \\
Voldemort: (seething, whispering as he retreats, his confidence shattered) We shall see, little wizards. (voice dripping with disdain) The night is still young, and darkness always prevails... (his voice fading as the light grows) \\
(Harry and Ron stand together, the bond between them stronger than ever) \\
Harry: (gripping Ron's shoulder, voice steady and filled with determination) Together, we can face anything. (thinking: I'm no longer alone; I have him by my side.) \\
(They exchange a determined look, ready to confront whatever comes next, united in purpose and friendship as the shadows of doubt dissipate.) \\[1em]

\textbf{Scene 5: The heart of the Dark Forest, a sinister lair} \\
\textbf{Plot Element:} Falling Action \\
\textbf{Beat:} Harry and Ron reach the lair and encounter a traitorous guide leading them into traps, forcing them to confront their trust issues, but ultimately bringing them closer as they navigate through Voldemort's dark enchantments, reinforcing their commitment to each other and Hermione. \\[0.3em]

\textbf{Scene Description:} In the heart of the Dark Forest, ancient trees loom like twisted sentinels, their gnarled branches clawing at the ink-black sky, desperate to escape the suffocating darkness that thickens the air. A damp, clinging fog coils around Harry and Ron's feet, muffling their cautious footsteps as they plunge deeper into the lair, the atmosphere heavy with the musty scent of decaying leaves and damp earth. The faint echo of dripping water creates a disquieting rhythm, amplifying their sense of isolation and dread, while distant whispers ripple through the air, taunting them with insidious doubts and the specter of betrayal. A sudden gust of wind tugs at their cloaks, pulling them apart like the very shadows that flicker and dance at the periphery of their vision, crafting a labyrinth of terror and uncertainty. The ground beneath their feet is uneven, a treacherous tapestry of roots and hidden pitfalls, each step a foreboding warning of the dangers lurking in the darkness. Just ahead, a figure emerges from the gloom—a treacherous guide with a sly, too-wide smile that starkly contrasts with the oppressive atmosphere surrounding them. His voice slithers through the stillness, smooth yet laced with a sinister undertone. `You must trust me, boys,' he croons, insincerity dripping from his words like poison. Harry and Ron exchange wary glances, their breaths visible in the frigid air, hearts pounding like war drums in their chests. Ron's fists clench at his sides, a physical manifestation of the tension coiling between them, mirroring the internal struggle of trust versus suspicion that gnaws at him. Shadows flicker against the rough, damp walls of the lair, amplifying their fears as the guide gestures toward a darkened path ahead. `This way,' he beckons, but the edge in his voice makes Ron flinch, a reminder of the countless times he felt overshadowed by his brothers. `Stay close,' Harry mutters, his voice barely audible over the eerie rustle of unseen creatures, the cold air wrapping around them like a vice. A low growl reverberates through the trees, echoing their fears and amplifying the danger that lies ahead. With each cautious step, they tread a treacherous line between the dark enchantments that seek to divide them. Ron instinctively reaches for Harry's arm, pulling him back just in time to avoid a hidden pitfall, their breaths quickening as adrenaline surges through their veins. Relief washes over them momentarily, dispelling the shadows of doubt, yet as they press on, the oppressive darkness seems to close in, the walls pulsating with a life of their own, mirroring their rising anxiety. Amidst the labyrinth of deception, a flicker of resolve ignites between them—this bond forged in adversity is what will ultimately guide them through Voldemort's wicked design. As the guide's manipulative words swirl around them, Harry and Ron lock eyes, an unspoken understanding passing between them like a silent vow. The warmth of their shared breath mingles in the cold air, and for a brief moment, the shadows recede, revealing the strength of their friendship. But just as Ron reaches for Harry's shoulder again, the guide's laughter echoes sharply, a sudden crack like a branch snapping in the darkness, forcing them to confront the trap that is closing around them. In that instant, the weight of their past grievances begins to lift, solidifying their trust as they press onward into the darkness together, armed with newfound determination to confront whatever lies ahead. \\[0.3em]

\textbf{Dialogue:} \\
Harry: (heart racing, eyes scanning the shadowy lair, grip tightening around his wand as he takes a step closer to Ron, the chill in the air amplifying his urgency) We have to stick together, Ron. This place feeds on our fears, trying to tear us apart. (thinking: I can't let doubt seep in; Hermione's counting on us. I won't fail her again.) \\
Ron: (nervous laughter bubbles up, shoulders tense as he fiddles with the hem of his cloak, glancing around the dimly lit space, his voice quivering slightly) Right, but it feels like the walls are closing in. It's like something's lurking in the shadows, just waiting for the perfect moment to pounce. (internal conflict: the bitter echo of feeling lesser than my brothers gnaws at him.) \\
Harry: (steps closer, voice low and firm, scanning their surroundings defiantly as if daring the darkness to challenge them, tension thickening the air) We can't let fear win, Ron. Remember why we're here—Hermione needs us. (thinking: I have to be strong for both of us; I can't let her down again.) \\
Ron: (voice trembling, wringing his hands, eyes darting toward the dark path ahead, doubt flickering across his face, vulnerability surfacing) What if I'm not brave enough? What if I screw this up? I can't lose her, Harry. (psychological insight: the shadow of inadequacy looms large, whispering that he'll never measure up to Harry.) \\
Harry: (places a steadying hand on Ron's shoulder, warmth radiating from his grip, trying to convey strength) You're braver than you realize. We've faced worse together—trust yourself. (thinking: I need him to believe in himself; our friendship is the anchor we need to navigate this darkness.) \\
Ron: (taking a deep breath, eyes narrowing with determination, posture shifting to one of resolve, a flicker of hope igniting) Right. Together. No more hesitation. We will get her back. (psychological insight: the weight of their friendship ignites a fire within him, pushing away the shadows of self-doubt.) \\
(As they venture deeper into the lair, the guide steps from the shadows, a sly smile twisting his lips, his presence chilling the air.) \\
Guide: (voice smooth, almost a whisper, eyes glinting with malice as he gestures toward the path) Ah, two brave souls wandering in the dark. But tell me, are you sure he wouldn't leave you behind to save her? You know how he feels about his precious Hermione. (manipulative tone: sowing seeds of doubt, his words curling like smoke in the air.) \\
Harry: (defensive, stepping closer to Ron, jaw clenched, fists tightening around his wand as he glares at the guide, determination etched on his face) Don't listen to him, Ron. He's trying to play us against each other. (thinking: I won't let him drive a wedge between us; we're stronger than his deceit. We have to trust each other.) \\
Ron: (shifting uncomfortably, glancing at Harry, voice trembling with uncertainty, the weight of the guide's words pressing down on him) But what if he's right? What if I'm not worth saving? (internal struggle: the overwhelming weight of expectations crushes him, doubt creeping in like a shadow.) \\
Harry: (urgently, voice rising, eyes fierce and unwavering, desperation clawing at him) Ron, no! I need you. We need each other! Hermione is counting on us—don't let him twist that against us. (Psychological insight: the fear of losing Ron's support sends a chill through him.) \\
Guide: (smirking, relishing the discord, voice dripping with mockery) You see? The dark whispers are already taking root. Trust is a fragile thing, isn't it? (feeding on their insecurities, eyes dancing with malice.) \\
Ron: (gathering resolve, glaring at the guide, voice strong, a flicker of defiance igniting in his chest) We can't trust you! You're just trying to divide us. Hermione needs us united! (thinking: I won't let him manipulate me; I'm stronger than this. I have to believe in my worth.) \\
Harry: (nodding, feeling their bond strengthen, voice steady and resolute) Exactly. Just... keep an eye on each other. This place will try to trick us. (internal thought: our friendship is our greatest weapon, and we must wield it together against Voldemort.) \\
Ron: (with newfound courage, smirking defiantly, fists clenched, adrenaline surging) Alright, let's show this forest who it's messing with. Together, we're unstoppable. (Psychological insight: the realization of their strength in unity fuels his spirit, driving away the shadows of doubt.) \\
Harry: (grinning, feeling hope surge within him, a light breaking through the oppressive darkness) That's the spirit! Let's go save our friend. (thinking: we can do this together; nothing can break our bond. We're in this for Hermione.) \\
(The guide's smile falters, annoyance flickering across his face as Harry and Ron move past him, hearts pounding but unified in purpose, ready to face whatever darkness lies ahead.) \\[1em]

\textbf{Scene 6: Inside Voldemort's lair, a dimly lit chamber} \\
\textbf{Plot Element:} Resolution \\
\textbf{Beat:} After a fierce battle with Voldemort, Harry's love for Hermione and Ron's newfound confidence culminate in a united front against Voldemort, leading to Hermione’s escape and showcasing their growth in trust and reliance on each other as they triumph over darkness. \\[0.3em]

\textbf{Scene Description:} Inside Voldemort's lair, the chamber unfolds like a crypt, suffocated by an oppressive twilight that extinguishes any flicker of hope. Jagged stone walls, slick with moisture, absorb the last remnants of dying embers, casting grotesque shadows that twist and writhe like tortured souls ensnared in eternal despair. The air is thick with the acrid scent of burnt magic, a miasma of sorrow that weighs heavily upon Harry and Ron, remnants of their fierce battle echoing in their minds. A low, haunting whisper dances through the vastness, the very stones mourning Voldemort's fleeting defeat, their sorrow palpable in the charged atmosphere. At the center of this grim tableau, Harry stands resolute, his heart pounding with a mix of adrenaline and fear. A chilling draft sweeps through the chamber, tugging at his robes like icy fingers of dread. Yet, in the depths of his mind, memories surge like beacons of light—the warmth of Hermione's laughter, the comforting scent of the Gryffindor common room, their shared jokes echoing like lifelines in the darkness. Each recollection ignites a fierce determination within him, illuminating the shadows that threaten to consume him. Beside him, Ron stands firm, their bodies aligned as one, hands gripping each other's shoulders tightly—a silent vow of solidarity and trust that speaks volumes. Ron's posture is tall, shoulders squared, embodying his newfound confidence. As Voldemort's mocking laughter reverberates through the chamber, Ron's triumphant shout pierces the suffocating silence, a bold declaration of their unity. Suddenly, Hermione's gasp—a melody of hope—reverberates through the air, breaking the stifling gloom as she seizes her moment. Drawing on her inner strength, she conjures a spell, the light bursting forth like dawn breaking through the night, illuminating the dark corners of the chamber and showcasing her growth and determination. The chamber trembles, the atmosphere charged with anticipation as Voldemort's furious howls echo, a chilling reminder of the darkness still lurking. His posture shifts from confident to defensive, a flicker of vulnerability momentarily breaking through his cold demeanor as he reveals the painful memories that haunt him—moments of rejection and loss that twisted his heart into darkness. Yet, in stark contrast, Harry and Ron stand back-to-back, their movements synchronized as they cast spells together, channeling their newfound confidence. Harry's fierce determination radiates through his clenched fists and unwavering gaze, while Ron stands resolutely by his side, their expressions a testament to their evolution. Together, they conjure a protective barrier, a shimmering shield encapsulating their trust and unity, pushing back against the encroaching shadows. Amidst the chaos, the air thickens with palpable hope, heralding their reunion and the dawn of a new beginning. As the darkness lifts slightly, a surge of warmth envelops the chamber, contrasting the dimness of Voldemort's lair—fresh earth and blooming flowers scenting the air as they overcome the darkness. Harry and Ron share a moment of relief and joy, clasping hands in a gesture of unwavering bond and sacrifice, their triumphant unity a radiant defiance against the encroaching despair. This victorious moment sets the stage for the dialogue that will follow, as they prepare to confront the remnants of darkness together, united in their strength and love. \\[0.3em]

\textbf{Dialogue:} \\
Harry: (heart pounding, gripping his wand tightly, knuckles white, stepping forward with fierce determination, eyes blazing with resolve) You won't take her from me, Voldemort! I swear, I won't let you win! (thinking: I refuse to let fear drown my resolve; I have to protect Hermione, no matter the cost, even if it means facing my darkest fears.) \\
Voldemort: (mocking, his voice a serpentine whisper that slithers through the chamber, eyes gleaming with malice, stepping closer, each movement deliberate, lips curling into a cruel smile) Ah, the brave little Gryffindor. Love? Such a weakness—it's precisely why you'll fail! (internal conflict: the bitter memory of my mother's sacrifice stings, tightening my fists in suppressed rage; how could they not see the truth?) \\
Ron: (standing firm beside Harry, his shoulders squared, fists clenched tightly, heart racing, a newfound fire igniting within him) We're not afraid of you! Together, we're stronger than you'll ever be! (thinking: This is our moment to prove ourselves—not just for Hermione, but for everything we've fought for, for all the times we've stood together.) \\
Hermione: (defiant, straining against her bonds, her eyes ablaze with courage, every fiber of her being radiating strength) You think you can break us? You’re wrong! Love is our power! \\
Voldemort: (growing angry, pacing like a caged beast, shadows flickering with his fury, voice rising) Love? It’s nothing but a fleeting illusion, a distraction from true power! You will see! (internal thought: My mother’s love was my downfall; how can they not see it as I do? ) \\
Harry: (memories of Hermione’s laughter and Ron’s unwavering support flooding his mind, heart swelling with purpose) No! It’s the only thing that gives me strength. I can feel it—Hermione, Ron, I’m not alone! \\
Ron: (nodding, glancing at Hermione with a brightening smile, recalling their shared bravery) That’s right! Remember the Basilisk? We fought it together! We are stronger united! \\
Voldemort: (a flicker of fear crossing his face, his voice trembling, eyes narrowing, desperately attempting to manipulate) You think love makes you strong? I was betrayed by it. My mother’s love was my downfall! And you dare call that same pathetic feeling a power? It is a weakness—and I will crush you with it! \\
Harry: (eyes narrowing, anger bubbling beneath the surface, voice steady, fueled by the injustice of Voldemort's twisted perspective) That's where you're wrong! Love isn't a weakness; it's a strength you'll never understand! (thinking: I won't let him twist my memories; Hermione and Ron are my family now. We are stronger together.) \\
Ron: (gaining confidence, stepping closer, fists shaking with adrenaline, feeling the weight of his friends' love) You're just a coward hiding behind your power! Love is what makes us brave! (internal thought: I won't let him intimidate us, not now; we've come too far to turn back.) \\
Voldemort: (desperation creeping in, voice rising, panic lacing his tone) You think you can defy me? I am the master of death! (internal conflict: the walls of my power are crumbling, and I can feel it; desperation claws at my sanity.) \\
Harry: (with conviction, stepping forward, chin raised defiantly, voice steady, the weight of his friends' love empowering him) But you're also terrified of what love can do. That's your true weakness! (reflecting: love is something he'll never understand, and it infuriates him, pushing him closer to madness.) \\
Hermione: (finally free, stepping forward with grace and strength, her presence radiating hope, fierce light against the encroaching shadows) And love will always find a way to conquer darkness! (thinking: I'm back, and we're stronger together; no one can stop us now. I will fight for my friends.) \\
Voldemort: (screaming, realizing his impending defeat, backing away, voice laced with rage and fear) NO! You will all pay for this! (feeling cornered, desperate; my grip on power is slipping, and I can feel the darkness I commanded turning against me.) \\
Harry: (with finality, unleashing his power, feeling the surge of unity, heart racing, voice ringing clear) It's happening, Voldemort! For Hermione, for Ron, for everyone you've hurt! This ends now! (internal thought: I can't let him win, not after everything he's done; this is our fight.) \\
Ron: (shouting, unleashing his magic, stepping closer to Harry, adrenaline coursing through him, feeling invincible) For all of us! Remember the times we stood by each other? This is our moment! (Psychological insight: the past fuels our present strength, igniting our courage, making us unstoppable.) \\
Hermione: (with unwavering strength, voice ringing clear, stepping forward, her presence radiant, determination shining in her eyes) For love! We're unstoppable when we're together! (thinking: I'll fight for them, for us; we're in this together, and together we will prevail.) \\
Voldemort: (desperation in his voice, pacing, his facade cracking, eyes wide with disbelief) No! This cannot be happening! (internal thought: the fear of losing control is suffocating me, and I know this is the end; I am losing everything.) \\
Harry: (with fierce determination, raising his wand, feeling the surge of unity, voice clear and filled with hope) Not today. This is our moment. Together! (internal thought: we've come too far to back down now; this is our fight.) \\
(Ron and Hermione nod, their powers merging with Harry's as they unleash a wave of magic that shatters Voldemort's dark influence. The chamber trembles with the force of their unity. The echoes of hope resonate in the air, heralding a new dawn as darkness recedes.) \\

\bottomrule
\end{longtable}
\end{samepage}
\endgroup

\clearpage

\section{Statistics of the Experimental Dataset}

This section provides detailed statistics about the script lengths in our evaluation dataset, which consists of 50 narrative scripts as the initial inputs.

Table~\ref{tab:dataset_statistics} presents comprehensive statistics for each component in our dataset, including the number of scripts, average length, and range of script lengths.

\begin{table}[!ht]
\centering
\caption{Dataset statistics showing script length distribution across five datasets. All lengths are measured in words.}
\label{tab:dataset_statistics}
\footnotesize
\setlength{\tabcolsep}{4pt}
\renewcommand{\arraystretch}{1.2}
\begin{tabularx}{\columnwidth}{@{}l>{\centering\arraybackslash}X>{\centering\arraybackslash}X>{\centering\arraybackslash}X>{\centering\arraybackslash}X@{}}
\toprule
\textbf{Category} & \textbf{\makecell{Script \\ Count}} & \textbf{\makecell{Average \\ Length}} & \textbf{\makecell{Min \\ Length}} & \textbf{\makecell{Max \\ Length}} \\
\midrule
Dramatron & 10 & 3,814 & 2,841 & 4,809 \\
MoPS & 10 & 3,736 & 3,220 & 5,252 \\
Writingprompts & 10 & 1,873 & 1,524 & 2,233 \\
Agents' Room & 10 & 1,364 & 800 & 2,484 \\
DOC & 10 & 1,260 & 1,061 & 1,860 \\
\midrule
\textbf{Overall} & \textbf{50} & \textbf{2,410} & \textbf{800} & \textbf{5,252} \\
\bottomrule
\end{tabularx}
\end{table}

\end{document}